\documentclass{article}

\usepackage{microtype}
\usepackage{graphicx}
\usepackage{subcaption}
\usepackage{booktabs} 

\usepackage{hyperref}

\usepackage[most]{tcolorbox}

\newtcolorbox{examplebox}[1][]{
    colback=gray!8,
    colframe=gray!40,
    boxrule=0.5pt,
    arc=2pt,
    left=6pt,
    right=6pt,
    top=6pt,
    bottom=6pt,
    fonttitle=\bfseries,
    #1
}

\newtcolorbox{titledbox}[2][]{
    colback=gray!8,
    colframe=gray!40,
    boxrule=0.5pt,
    arc=2pt,
    left=6pt,
    right=6pt,
    breakable,
    top=6pt,
    bottom=6pt,
    fonttitle=\bfseries,
    title=#2,
    #1
}

\usepackage[preprint]{icml2026}

\usepackage{amsmath}
\usepackage{amssymb}
\usepackage{mathtools}
\usepackage{amsthm}

\usepackage[capitalize,noabbrev]{cleveref}

\DeclareMathOperator*{\argmin}{\arg \min}

\theoremstyle{plain}
\newtheorem{theorem}{Theorem}[section]

\theoremstyle{definition}
\newtheorem{definition}[theorem]{Definition}

\theoremstyle{remark}

\renewcommand{\vec}[1]{\boldsymbol{#1}}

\usepackage[textsize=tiny]{todonotes}

\icmltitlerunning{Epistemic uncertainty estimation methods are fundamentally incomplete}

\begin{document}

\twocolumn[
  \icmltitle{Position: Epistemic uncertainty estimation  \\
    methods are fundamentally incomplete}



  \icmlsetsymbol{equal}{*}

  \begin{icmlauthorlist}
    \icmlauthor{Sebastián Jiménez}{equal,yyy}
    \icmlauthor{Mira Jürgens}{equal,yyy}
    \icmlauthor{Willem Waegeman}{yyy}
  \end{icmlauthorlist}

\icmlaffiliation{yyy}{Department of Data Analysis and Mathematical Modeling, Ghent University, Ghent, Belgium}

  \icmlcorrespondingauthor{Sebastián Jiménez}{sebastian.jimenez@ugent.be}
  \icmlcorrespondingauthor{Mira Jürgens}{mira.juergens@ugent.be}

  \icmlkeywords{Uncertainty Estimation, Uncertainty Disentanglement, Heteroscedastic Regression}

  \vskip 0.3in
]



\printAffiliationsAndNotice{\icmlEqualContribution}

\begin{abstract}
Identifying and disentangling sources of predictive uncertainty is essential for trustworthy supervised learning. We argue that widely used second-order methods that disentangle aleatoric and epistemic uncertainty are fundamentally incomplete. First, we show that unaccounted bias contaminates uncertainty estimates by overestimating aleatoric (data-related) uncertainty and underestimating the epistemic (model-related) counterpart, leading to incorrect uncertainty quantification. Second, we demonstrate that existing methods capture only partial contributions to the variance-driven part of epistemic uncertainty; different approaches account for different variance sources, yielding estimates that are incomplete and difficult to interpret. Together, these results highlight that current epistemic uncertainty estimates can only be used in safety-critical and high-stakes decision-making when limitations are fully understood by end users and acknowledged by AI developers.  

\end{abstract}

\section{Introduction}
\label{sec1}

\begin{figure*}[t]
    \centering
    \begin{subfigure}[t]{0.49\textwidth}
        \centering
        \includegraphics[width=\textwidth,trim=0 10 0 0, clip]{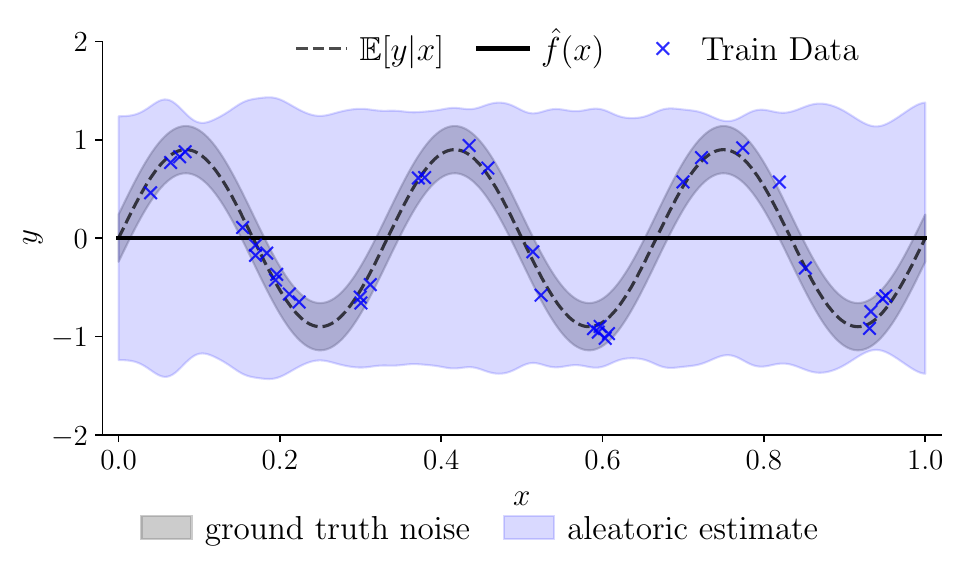}
        \label{fig:motivation_bias}
    \end{subfigure}
    \hfill
    \begin{subfigure}[t]{0.49\textwidth}
        \centering
        \includegraphics[width=\textwidth,trim=0 10 0 0, clip]{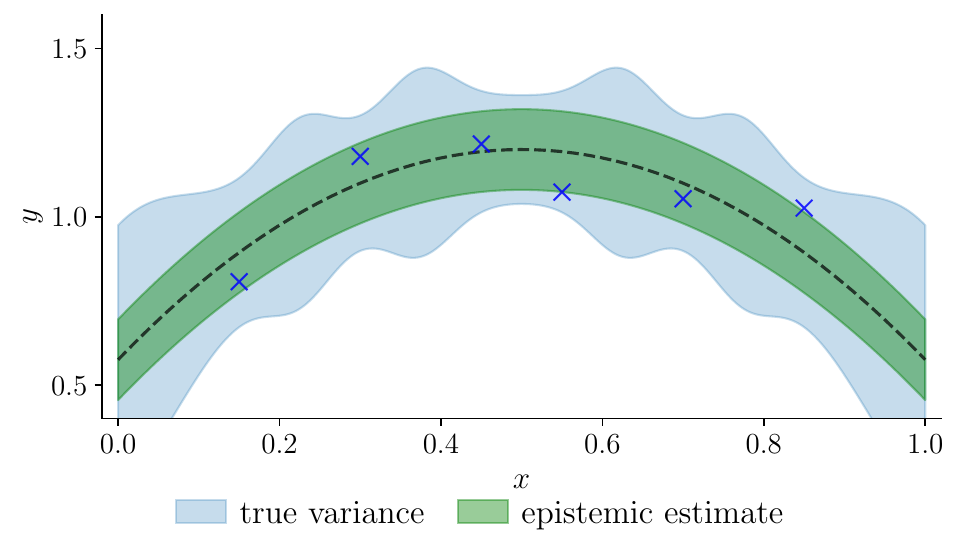}
        \label{fig:motivation_variance}
    \end{subfigure}
    \caption{
    \textbf{Conceptual illustration of our two main arguments for the regression case.}
    Left: When high model bias is present, we show that aleatoric uncertainty 
    \textit{overestimates} the ground truth noise.
    (Sec.~\ref{sec: generalized bias variance}).
   Right: Many second-order methods capture only \textit{procedural} and not data uncertainty (Sec.~\ref{sec: data and procedural uncertainty}), thereby underestimating the variance part of the epistemic uncertainty.}
    \label{fig:motivation}
\end{figure*}

    
    

Uncertainty estimation in machine learning is essential for assessing the reliability of model predictions. In recent supervised learning research, one often distinguishes between two types of uncertainty: 
aleatoric and epistemic uncertainty \cite{kendall2017uncertainties,hullermeier2021aleatoric}. Assuming that the ground
truth data-generating process can be described by a conditional probability distribution $p(y| \vec{x})$, aleatoric uncertainty
reflects the inherent randomness of this distribution, and is hence irreducible. Epistemic uncertainty, on the other hand, appears because the unknown $p(y| \vec{x})$ needs to be estimated using data. As a result, epistemic uncertainty can be reduced by collecting more data, improving model specification, or changing the learning algorithm. While the correct \textit{disentanglement} \cite{depeweg2018decomposition,valdenegro2022deeper} 
 of these two types of uncertainty is crucial for model interpretability, and for downstream tasks such as out-of-distribution detection \cite{hendrycks2017baseline} and active learning \cite{gal2017deep}, it is far from trivial.

One principled way of representing epistemic uncertainty is through a \textit{second-order distribution}, that is, a distribution over possible predictive distributions \cite{bengs2022pitfalls}. In this work, we focus on second-order distributions because many existing uncertainty quantification approaches can be interpreted within this framework, including Bayesian methods, ensemble methods, and deterministic approaches \cite{gawlikowski2023survey, he2025survey}. In Bayesian models, uncertainty is represented by a posterior distribution over model weights, which induces a distribution over a classical first-order distribution (e.g. a distribution over mean and variance of a Gaussian in a regression setting). In posterior predictive sampling, one takes multiple samples of the weights from the posterior distribution and computes the corresponding second-order distribution~\cite{gal2016dropout,denker1990transforming,neal2012bayesian,blundell2015weight}. Ensemble methods similarly induce a distribution over first-order distributions by aggregating outputs from multiple independently trained models \cite{lakshminarayanan2017simple, ovadia2019can}. Finally, some approaches aim to model epistemic uncertainty more directly by learning the parameters of the second-order distribution using specific loss functions~\cite{charpentier2021natural,amini2020deep,sensoy2018evidential,malinin2018predictive}. Viewed through this lens, these approaches provide a common framework for disentangling aleatoric and epistemic uncertainty. 

Under a second-order representation, uncertainty can be disentangled in different ways depending on the learning task. In classification, uncertainty is commonly analyzed using information-theoretic decompositions, which separate total, aleatoric, and epistemic uncertainty \cite{depeweg2018decomposition}. For regression, analogous variance-based decompositions are typically employed \cite{lakshminarayanan2017simple,depeweg2018decomposition,sale2023second}. However, a growing body of critical work argues that such second-order decompositions are not theoretically well-founded and can be difficult to interpret in practice. For evidential second-order risk minimization, recent analyses highlight identifiability and convergence pathologies, showing that the resulting epistemic uncertainty estimates are tightly controlled by regularization, making them only meaningful in a relative sense  \cite{bengs2022pitfalls,meinert2023unreasonable,juergensepistemic}. 
Complementarily, in Bayesian deep learning, \cite{fellaji2024epistemic} report failure modes in which epistemic uncertainty can collapse as model size increases, contrary to what is expected. More broadly, \cite{kirchhof2025position} emphasizes that aleatoric/epistemic definitions and decomposition principles can be mutually inconsistent, and that practical estimators may become strongly correlated, challenging the premise that these components are meaningfully disentangled. In the same sense, recent works show that these decompositions are theoretically dubious \cite{wimmer2023quantifying,smith2024rethinking}, and may produce highly correlated aleatoric and epistemic uncertainty estimates \cite{mucsanyi2024benchmarking}, questioning their ability to disentangle different sources of uncertainty meaningfully. 

{\bf Our position.} The most prominent recent papers that propose uncertainty disentanglement methods have collectively received almost 40,000 citations\footnote{Number obtained from Google Scholar. Check Appendix \ref{app:impact of literature} for details.}, and their usage in safety-critical applications is increasing. This relevance, together with the recent critical work motivate a more fundamental reconsideration of what epistemic uncertainty is intended to represent and quantify in predictive models. Accordingly, \textbf{we argue that recent and highly-cited epistemic uncertainty estimation methods are fundamentally incomplete}. To substantiate this claim, we proceed as follows. In Section \ref{sec: epistemic uncertainty definition}, we review and systematize the different sources of epistemic uncertainty identified in the literature, distinguishing between estimation uncertainty (variance-related), model uncertainty (bias-related), and distributional uncertainty. In Section \ref{sec: generalized bias variance} we adopt the bias–variance decomposition of Bregman divergences \cite{hastie2009elements,gruber2023sources} and demonstrate that the bias component is as central to epistemic uncertainty as the variance component. Within this framework, we show that \textbf{bias induced by the regularization mechanisms commonly used in neural networks can substantially distort uncertainty estimates}, even when the true data-generating function lies within the model’s hypothesis space. In Section \ref{sec: data and procedural uncertainty}, we turn to the variance component and demonstrate that \textit{existing uncertainty quantification methods fail to capture it in its entirety}; instead, different approaches account for different variance contributions, resulting in estimates that are inherently incomplete and difficult to interpret. Finally, in Section \ref{sec: alternative views} we discuss alternative perspectives on epistemic uncertainty proposed in the literature and relate them to our work.


\section{Setup and preliminaries}
\label{sec: epistemic uncertainty definition}
\paragraph{Supervised learning setup.} We consider an instance space $\mathcal{X} \subset \mathbb{R}^D$ 
and label space $\mathcal{Y}$, where $\mathcal{Y} \subseteq \mathbb{R}$ for regression and $\mathcal{Y} = \{1, \ldots, K\}$
 for classification. We assume a dataset $\mathcal{D}_N = \{(\vec{x}_i, y_i)\}_{i=1}^N$ of i.i.d.\ samples from an unknown
joint distribution $P$ on $\mathcal{X} \times \mathcal{Y}$, with ground-truth conditional distribution $p(y|\vec{x})$.
 As a loss function $\ell: \mathcal{P}(\mathcal{Y}) \times \mathcal{Y} \to \mathbb{R}$ we assume a \emph{strictly proper scoring rule}
 \citep{gneiting2007strictly} which uniquely minimizes the expected loss when the prediction equals the ground truth.
Every strictly proper scoring rule induces a convex entropy
 $\mathbb{H}_\ell$ and a Bregman divergence $D_\ell$
  such that the pointwise expected score gap equals the divergence:
\begin{equation*}
     D_\ell(p(\cdot|\vec{x}), \hat{p}(\cdot|\vec{x})) : =\mathbb{E}_{y|\vec{x}}[\ell(\hat{p}(\cdot|\vec{x}), y)]
     - \mathbb{E}_{y|\vec{x}}[\ell(p(\cdot|\vec{x}), y)],
\end{equation*}
with entropy $\mathbb{H}_\ell(p(\cdot|\vec{x})) := \mathbb{E}_{y|\vec{x}}[\ell(p(\cdot|\vec{x}), y)]$ \cite{banerjee2005clustering,gneiting2007strictly}.
For a hypothesis space $\mathcal{H} := \{p_{\vec{\theta}} : \mathcal{X} \to \mathcal{P}(\mathcal{Y}) | \vec{\theta} \in \Theta\}$,
 of possible predictive distributions parametrized by $\boldsymbol{\theta} \in \Theta$,
 the \emph{risk} of an estimator $\hat{p} \in \mathcal{H}$ integrates the pointwise expected loss over $\vec{x}$:
\begin{equation*}
    \mathcal{L}(\hat{p}) :=
     \mathbb{E}_{\vec{x}}\left[\mathbb{E}_{y|\vec{x}}[\ell(\hat{p}(\cdot|\vec{x}), y)]\right] = \mathbb{E}_{(\vec{x}, y)}[\ell(\hat{p}(\cdot|\vec{x}), y)].
\end{equation*}
By strict properness, the unconstrained risk minimizer is $p(y|\vec{x})$ itself.
 The \emph{Bayes-optimal estimator} within $\mathcal{H}$ is 
 \begin{equation}
    \label{eq: bayes estimator}
 \hat{p}^* := \arg\min_{\hat{p} \in \mathcal{H}} \mathcal{L}(\hat{p}).
 \end{equation}

 \textbf{Second-order distributions.} 
 Following common literature \cite{bengs2022pitfalls,hullermeier2021aleatoric,gawlikowski2023survey,li2025out},
  we represent uncertainty via a second-order distribution $q(\vec{\theta}|\vec{x})$ over parameters $\theta \in \Theta$ of the (first-order) distribution.
   The predictive distribution can then obtained by model averaging:
\begin{equation}
    \hat{p}(y|\vec{x}) = \int_{\Theta} p(y|\vec{\theta}, \vec{x}) \, q(\vec{\theta}|\vec{x}) \, d\vec{\theta}.
    \label{eq:model_averaging}
\end{equation}
Existing measures that decompose the uncertainty of  $q(\vec{\theta}|\vec{x})$ into an
aleatoric and epistemic component focus on entropy \cite{depeweg2018decomposition}, distance to reference sets \cite{sale2023second},
or proper scoring rules \cite{hofman2024quantifying,fishkov2025uncertainty}. In regression, the \textit{variance-based decomposition} \cite{depeweg2018decomposition, sale2023second} is most commonly used. In this approach, the variance of the prediction $\hat{y}$, obtained via model averaging as in Eqn.~\ref{eq:model_averaging}, is decomposed into an aleatoric and an epistemic part:
\begin{equation}
\label{eq: aleatoric-epistemic}
    \text{Var}(\hat{y}| \vec{x}) = \underbrace{\mathbb{E}_{\vec{\theta}}\big[\text{Var}(\hat{y}| \vec{x}, \vec{\theta})\big]}_{\text{ aleatoric estimate}} +
     \underbrace{\text{Var}_{\vec{\theta}}\big(\mathbb{E}[\hat{y}|\vec{x}, \vec{\theta}]\big)}_{\text{epistemic estimate}}.
\end{equation}

\paragraph{Sources of epistemic uncertainty.}

\begin{figure}[t]
    \centering
    \includegraphics[width=\linewidth]{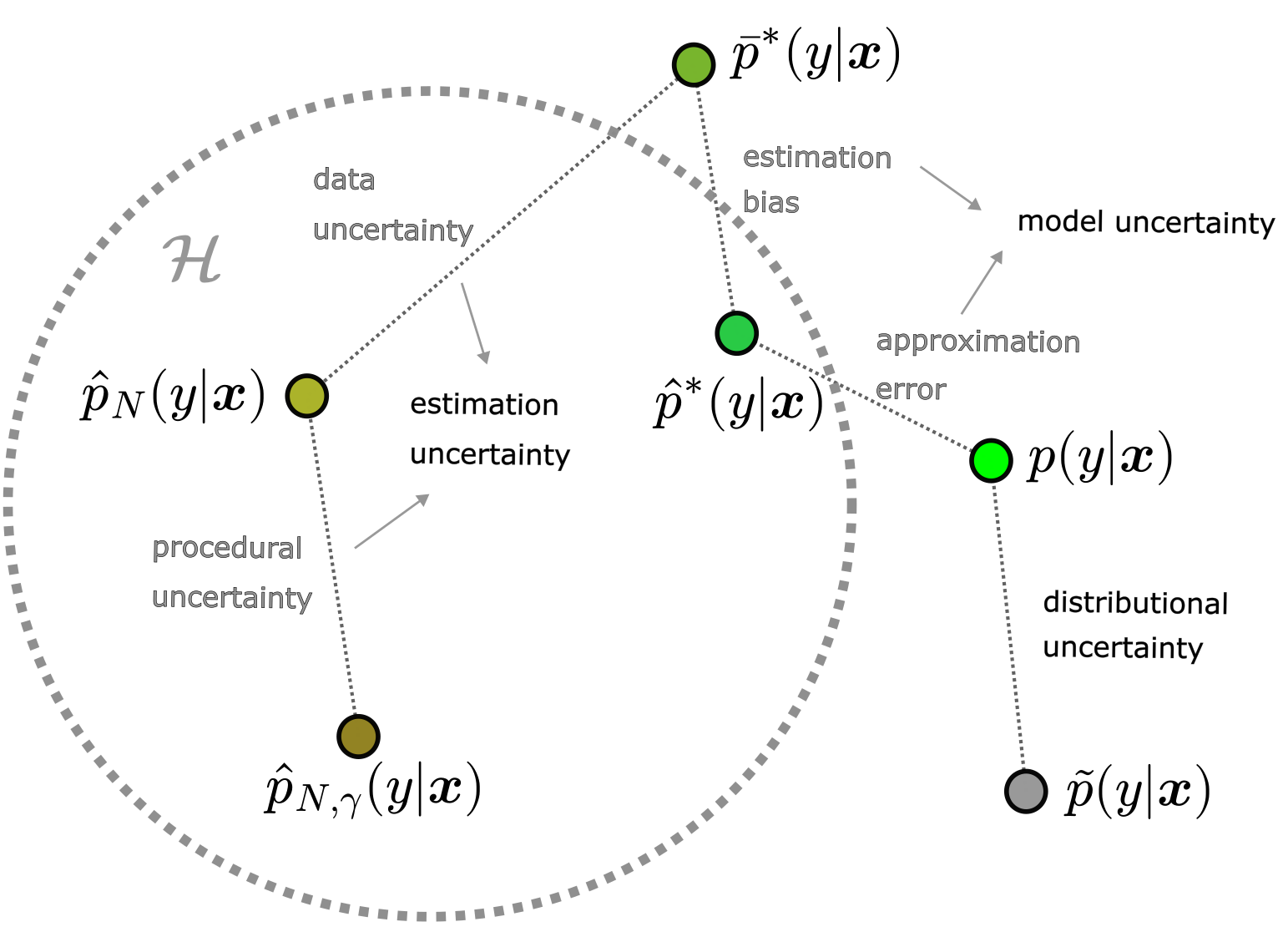}
    \caption{Sources of epistemic uncertainty in machine learning, where $p(y| \vec{x})$ denotes the true data-generating process. Different sources of epistemic uncertainty can lead to the estimate $\hat{p}(y|\vec{x})$ being further from the ground truth. See Sec.~\ref{sec: epistemic uncertainty definition} for further details.  
     Figure inspired from \citet{huang2023efficient}}
    \label{fig: epistemic uncertainties}
\end{figure}
As aleatoric uncertainty corresponds to the intrinsic randomness of the data-generating process,
one can directly associate it with the stochasticity induced by the ground truth conditional distribution, using measures such as variance or entropy. 
Epistemic uncertainty, on the other hand, measures the uncertainty about the ground truth, i.e., the expected deviation from it. It is reducible
 by collecting more knowledge, hence decreasing the uncertainty about the parameters $\vec{\theta}$,
  but also the chosen hypothesis space $\mathcal{H}$.
   This uncertainty can stem from different sources, which several authors discuss \cite{hullermeier2021aleatoric,huang2023efficient,malinin2018predictive}.
Unifying those perspectives, we will distinguish in total five distinct sources of epistemic uncertainty (see Figure~\ref{fig: epistemic uncertainties}): approximation error, estimation bias, data uncertainty, procedural uncertainty and distributional uncertainty.

Following  \cite{huang2023efficient}, we  define $\mathcal{D}_N \sim P^N$ as the random variable reflecting the \emph{data uncertainty},
 where $P^N$ is the distribution over all datasets of size $N$.
Similarly, let $\gamma \sim P^{\gamma}$ be the random variable reflecting all \emph{procedural uncertainty}, which arises from the training process from a single model (due to e.g.\ random weight initialization or data ordering) and is present even with infinite data.
 $P^{\gamma}$ denotes the distribution over which the procedural parameters are sampled. The estimator $\hat{p}_{N, \gamma}(y|\vec{x})$ depends on both
 types of uncertainties.
We define the \emph{dual mean} as the Bregman centroid \cite{banerjee2005clustering,pfau2013bregman}: 
\begin{equation}
    \bar{p}^*(y|\vec{x}) := \argmin_{\tilde{p} \in \mathcal{P}(\mathcal{Y})} 
    \mathbb{E}_{\mathcal{D}_N, \gamma}\left[D_\ell(\tilde{p}, \hat{p}_{N,\gamma}(\cdot|\vec{x}))\right].
    \label{eq:dual_mean}
\end{equation}
Note that $\bar{p}^*$ does not have to lie in $\mathcal{H}$, as depicted in Figure \ref{fig: epistemic uncertainties} For example, when $\mathcal{H}$ only consists of decision stumps, the dual mean might be a more complex decision tree.
 One can now divide the epistemic uncertainty into:

\textbf{Model uncertainty:} discrepancy between $\bar{p}^*(y|\vec{x})$ and $p(y|\vec
    {x})$. It can be further decomposed into \textit{estimation bias}, the systematic deviation of $\bar{p}^*(y|\vec{x})$ 
    from the Bayes estimator $\hat{p}^*(y|\vec{x})$ (Eq.~\ref{eq: bayes estimator}) 
    and \textit{approximation error}, which is the difference between $\hat{p}^*(y|\vec{x})$ and $p(y|\vec{x})$ \cite{brown2024bias}:
\begin{align}
\label{eq: approximation and estimation bias}
            &\mathbb{E}_{(\vec{x}, y)}[D_{\ell}(p(\cdot|\vec{x}), \bar{p}^*(\cdot|\vec{x}))]\\&
            \nonumber
            = \underbrace{\mathcal{L}(\hat{p}^*(y|\vec{x})) - \mathcal{L}(p(y|\vec{x}))}_{\text{approximation error}} + \underbrace{\mathcal{L}(\bar{p}^*(y|\vec{x})) - \mathcal{L}(\hat{p}^*(y|\vec{x}))}_{\text{estimation bias}}.
    \end{align}
        
\textbf{Estimation uncertainty:} variation of $\hat{p}_{N,\gamma}(y |\vec{x})$ around the dual mean $\bar{p}^*(y|\vec{x})$.
     It further decomposes into data and procedural uncertainty. For a fixed dataset $\mathcal{D}_N=\{(\vec{x}_i, y_i)\}_{i=1}^N$,
data uncertainty can be defined as the difference between $\bar{p}^*(y | \vec{x})$ and
 \begin{equation*} 
\hat{p}_N(y|\vec{x}) = \argmin_{\hat{p}_{N,\gamma} \in \mathcal{H}} \mathbb{E}_{\gamma}\big[\frac{1}{N} \sum_{i=1}^N \ell(\hat{p}_{N,\gamma}(\cdot|\vec{x}_i), y_i)\big],
 \end{equation*} 
 where the expected empirical loss is taken over the procedural parameters $\gamma$.
Similarly, for fixed procedural parameters $\gamma$, procedural uncertainty can be
 defined as the difference between $\bar{p}^*(y | \vec{x})$ and 
 \begin{equation*} 
\hat{p}_{\gamma}(y|\vec{x}) = \argmin_{\hat{p}_{N,\gamma} \in \mathcal{H}} \mathbb{E}_{\mathcal{D}_N}\big[ \frac{1}{N} \sum_{i=1}^N \ell(\hat{p}_{N,\gamma}(\cdot|\vec{x}_i), y_i)\big].
\end{equation*}
Using this notation, estimation uncertainty is the difference between $\bar{p}^*(y |\vec{x})$ and $\hat{p}_{N,\gamma}(y |\vec{x})$, as visualized in Figure~\ref{fig: epistemic uncertainties}.  

\textbf{Distributional uncertainty:} distribution shift between training and test data \cite{malinin2018predictive}, respectively, and their associated conditional distributions $p(y|\vec{x})$ and $\tilde{p}(y|\vec{x})$.  

\section{Epistemic uncertainty estimation methods are fundamentally incomplete}
In this section we argue that current approaches to estimating aleatoric and epistemic uncertainty are fundamentally incomplete. To our opinion, this incompleteness has two main reasons:
\begin{itemize}
\item \textbf{First, estimation bias and approximation error both distort epistemic and aleatoric uncertainty estimates}. 
Since second-order representations do not have access to the ground truth conditional class distribution, it is not possible to incorporate model uncertainty into epistemic uncertainty estimates. Consequently, a complete representation of epistemic uncertainty is fundamentally unattainable (Section~\ref{sec: generalized bias variance}).
\item \textbf{Second, existing estimators are also incomplete because they do not represent data uncertainty and procedural uncertainty at the same time}. From a bias-variance perspective, these two types of uncertainty are closely related to the variance of predictions (see Section~\ref{sec: data and procedural uncertainty}). 
\end{itemize}

To illustrate our arguments, we will focus on the estimation of uncertainty in a regression setting. 
Much of the 
existing critical literature on uncertainty disentanglement has concentrated on the classification setting, but the regression setting allows us to present the core ideas more transparently, while remaining complementary to prior work. We consider a data-generating process such that $y_i = f(\vec{x}_i) + \epsilon_i$ and $\epsilon_i \sim \mathcal{N}(0, \sigma^2(\vec{x}_i))$. Here, $\epsilon_i$ is data-dependent (\textit{heteroscedastic}) noise, and $f$ is a deterministic function,
such that $p(\cdot|\vec{x})$ forms the density of a Gaussian distribution with mean $f(\vec{x})$ and variance
$\sigma^2(\vec{x})$. 

For illustrations we make use of the synthetic regression setting of \citet{rossellini2024integrating}. A dataset $\mathcal{D}_N=\{(x_i, y_i)\}_{i=1}^N$ is generated where $x_i\sim Beta(1.2, 0.5)$ and
\begin{align}
\label{eq: sine data problem}
    y_i=\sin\left(\frac{1}{5(x_i+0.16)^3}\right)+\epsilon_i, \quad \epsilon_i\sim\mathcal{N}(0, \sigma^2(x_i)),
\end{align}
with $\sigma^2(x) = x^4$. This data-generating process is chosen because it represents a problem with two main regions. The left side of the data domain ($x<0.4$) has low true aleatoric uncertainty and a difficult-to-learn function, which is expected to induce a high epistemic uncertainty. The right side of the data domain ($x\geq0.4$) has high true aleatoric uncertainty and a simple function (which is expected to induce a low epistemic uncertainty). This one-dimensional setting is ideal for visually illustrating our arguments regarding uncertainty disentanglement (Figures~\ref{fig:deep-ensemble-unc} and \ref{fig:evaluation_de_reference}). As an example, we will illustrate the behavior of Deep Ensembles, considering a simple multi-layer perceptron that outputs $\hat{f}(x)$ and $\hat{\sigma}^2(x)$ as the mean and variance of the Gaussian distribution at $x$. Predictions coming from different parametrizations of the neural network are then aggregated to obtain a second-order distribution. In Appendix~\ref{app: experimental details} we provide similar illustrations for a real-world regression dataset and for other second-order models.


\subsection{Bias distorts uncertainty estimations.}
\label{sec: generalized bias variance}
From a statistician's viewpoint, the classical bias-variance decomposition provides further insight into the interplay between estimation uncertainty and aleatoric uncertainty \cite{hastie2009elements,gruber2023sources}. While it is originally defined for the mean squared error as a loss function, it can also be generalized to any proper scoring rule \cite{gruber2023uncertainty, pfau2013bregman}. As a small further extension, we incorporate data and procedural components into the decomposition:
\begin{align}
\label{eq: bias-variance}
\centering
\nonumber
\mathbb{E}_{\mathcal{D}_N,\gamma}[\mathcal{L}(\hat{p})]&=
    \underbrace{\mathbb{H}_{\ell}(p(y|x))}_{\text{aleatoric uncertainty}} + \underbrace{D_{\ell}(p(y|x),\bar{p}^*(y|x))}_{\text{generalized bias}} \\& +\underbrace{\mathbb{E}_{\mathcal{D}_N,\gamma}[D_{\ell}(\bar{p}^*(y|x), \hat{p}_{N, \gamma}(y|\vec{x}))]}_{\text{generalized variance}}.
\end{align}

 The first term in Eqn.~\ref{eq: bias-variance} represents the intrinsic noise of the data-generating process, and therefore corresponds to aleatoric uncertainty, which is irreducible with more training data. 
 The second term captures the generalised bias, defined as the divergence between the mean predicted distribution $\bar{p}^*(y|\vec{x})$ averaged over data and procedural uncertainty, and the true conditional distribution $p(y|\vec{x})$. 
Finally, the third term reflects the generalized variance of the predictive distribution under data and procedural uncertainty. In Section~\ref{sec: data and procedural uncertainty}, we will further decompose it into data and procedural components, by means of the conditional bias-variance decomposition for Bregman divergences (see  \cite{gneiting2007strictly,adlam2022understanding}). 

  Connecting the bias component to Eqn.~\ref{eq: approximation and estimation bias}, one can decompose it into approximation error and estimation bias. Mainly the approximation error has seen some discussion in previous research on epistemic uncertainty, because the approximation error is nonzero when the ground-truth model is not in the hypothesis space, which might happen for certain hypothesis spaces, such as linear models.  However, for universal function approximators, such as many types of deep neural networks, the approximation error will vanish \citep{Hornik1989,cybenko1989,Allenby2009}. 
 Therefore, we believe that estimation bias deserves more discussion in modern deep learning research. It arises from the use of finite training datasets and from regularization procedures employed during training, which is inevitable to make learning possible with a finite dataset \cite{hastie2009elements}. \textbf{Approximation error and estimation bias are both not included in existing epistemic uncertainty estimators, such as the one in Eqn.~\ref{eq: aleatoric-epistemic}, supporting our position that epistemic uncertainty estimation is fundamentally incomplete}.
 
In the classical bias-variance decomposition for regression, the first term of Eqn.~\ref{eq: bias-variance} is commonly estimated using Eqn.~\ref{eq: aleatoric-epistemic} as $\mathbb{E}[\hat{\sigma}^2]$, where $\hat{\sigma}^2$ denotes the predictive variance learned from the model. When minimizing squared loss, this quantity does not recover the true aleatoric noise (the details can be found in Appendix~\ref{app:biased aleatoric uncertainty}). Instead, it satisfies $\mathbb{E}[\hat{\sigma}^2]=\mathbb{E}[(y-\hat{f}(x))^2]+\sigma^2 $, as shown in \cite{zhang2024one}. Consequently, the learned aleatoric term absorbs observation noise and systematic prediction error: whenever $\hat{f}(x)$ deviates from the conditional mean, that deviation contributes to the learned $\hat{\sigma}^2$, inflating the aleatoric uncertainty estimation. Bias in the estimation of the mean will result in an overestimation of aleatoric noise. 

\begin{figure}[t]
  \includegraphics[width=\linewidth]{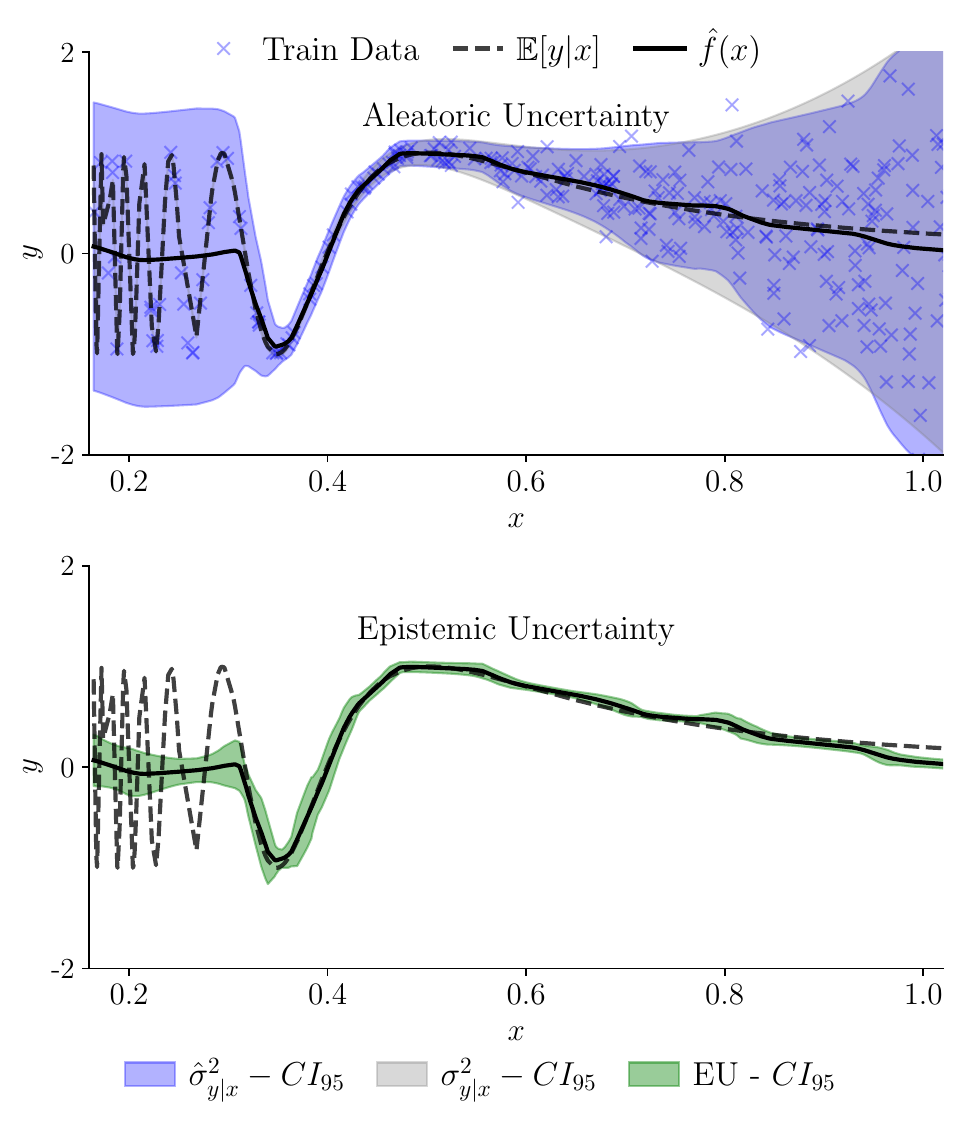}
  \caption{Deep Ensemble uncertainty estimates. The top row shows the training data points (blue crosses), the true conditional mean (dashed line), the model predictions (complete line), the $95\%$ confidence interval for aleatoric uncertainty (purple), as well as the true $95\%$ confidence interval of the aleatoric uncertainty (gray). The bottom row shows the estimate of epistemic uncertainty (green).}
  \label{fig:deep-ensemble-unc}
\end{figure}

\begin{titledbox}{Running Example - Deep Ensembles}
 Figure~\ref{fig:deep-ensemble-unc} presents the uncertainty estimates produced by a Deep Ensemble on the synthetic dataset that we described earlier. The figure illustrates two key failure modes. First, in the top panel, the left side of the input domain exhibits a systematic prediction error: due to regularization, the model has not yet recovered the true function in this region. Despite the observation noise being close to zero, the estimated aleatoric uncertainty is inflated, indicating that bias in the mean predictor is being absorbed into the aleatoric term. Second, in the bottom panel, the epistemic uncertainty remains low in the same region, even though the predictor is clearly misspecified with respect to the true function. \textbf{Taken together, these results show that, in the presence of systematic bias, deep ensembles can simultaneously overestimate aleatoric uncertainty and underestimate epistemic uncertainty, yielding an epistemic estimate that is inherently incomplete}.
\end{titledbox}
 
Despite a complex function on the left side of the domain, and a high aleatoric uncertainty on the right side, the synthetic example that we consider is with a one-dimensional feature space very simple. However, if uncertainty estimation is already unreliable in such a simple setting, there is very few hope that things get better for modern high-dimensional datasets. In Appendix~\ref{app: real data experiments} we observe exactly the same issues for a real-world regression dataset. Moreover, for other second-order methods, estimation bias also leads to an overestimation of aleatoric uncertainty --- see Appendix~\ref{app: synthetic data experiments} for additional experimental results that support this claim. In Bayesian methods, for instance, estimation bias is enforced by placing  priors on the weights of the neural network. For simple hypotheis spaces, this allows for a flexible way of incorporating domain knowledge, but the choice of priors becomes quite arbitrary in large Bayesian neural networks \cite{wilson2020bayesian, fortuin2022priors}. Other methods, such as Monte Carlo Dropout and Bootstrap ensembles, also contain regularization techniques by design, which also lead to estimation bias. 

\subsection{Data and procedural components of epistemic uncertainty}
\label{sec: data and procedural uncertainty}
In supervised learning, a trained predictor is a function of both the observed dataset 
and the learning procedure. A learning algorithm based on a neural network returns a predictive distribution $\hat{p}_{D_N,\gamma}(y|\vec{x})$,
which is dependent on both the dataset $D_N$ as well as the procedural parameters $\gamma$. As a result, variance  can be decomposed into a 
 procedural component, formed by the expected variance induced by $\gamma$, while the data is fixed, and a
data component, consisting of the variance induced by resampling datasets $D_N$. Using the law of total variance,
for $\hat{y} \sim \hat{p}_{\mathcal{D}_N, \gamma}(y|\vec{x})$ we have
\begin{equation}
\label{eq: variance decomposition mse}
\begin{split}
    \text{Var}_{\mathcal{D}_N, \gamma}(\hat{y}|\vec{x}) &= \mathbb{E}_{\mathcal{D}_N}\big[\text{Var}_{\gamma}(\hat{y}|\vec{x}) |\mathcal{D}_N\big] \\
    &\quad + \text{Var}_{\mathcal{D}_N}(\mathbb{E}_{\gamma}[\hat{y}|\vec{x}] |\mathcal{D}_N).
\end{split}
\end{equation}
So far we claimed that second-order models ignore the bias component when estimating epistemic uncertainty. 
\textbf{Using Deep Ensembles as a running example, we also argue that common second-order methods only track parts of the epistemic uncertainty that originates from variance.} 
 
\begin{titledbox}{Running Example - Deep Ensembles}
\paragraph{Deep Ensembles capture procedural uncertainty only.}
\label{ex: deep ensembles}
Let $\gamma \sim P^\gamma$ denote the random seed (weight initialization, 
data ordering, etc.) that differentiates each ensemble member's training, 
where all members are trained on the \emph{same} dataset $\mathcal{D}_N$. 
For the regression setting and a Deep Ensemble of size $d$, each member produces predictive parameters 
$\vec{\theta}{\gamma_i}(\vec{x}) = (f_{\gamma_i}(\vec{x}), \sigma^2_{\gamma_i}(\vec{x}))$, for $i=1, \dots, d$. 
The empirical second-order distribution is a mixture of Dirac measures:
\begin{equation*}
    q(\vec{\theta}|x) = \frac{1}{d} \sum_{i=1}^{d} \delta_{\theta_{\gamma_i}(x)}(\theta).
\end{equation*}
Applying the variance-based decomposition (Eqn.\ \ref{eq: aleatoric-epistemic}) and taking $d \to \infty$, the epistemic estimate is:
\begin{equation*}
    \text{Var}_\gamma\big(\mathbb{E}_{y|\vec{x}}[\hat{y}|\vec{x},\gamma]\big)
    = \mathbb{E}_\gamma\left[(f_\gamma(\vec{x}) - \bar{f}(\vec{x}))^2\right],
\end{equation*}
where $\bar{f}(\vec{x}) = \mathbb{E}_\gamma[f_\gamma(\vec{x})]$. This term is linked 
to the first term in Eqn.~\ref{eq: variance decomposition mse}, reflecting the procedural uncertainty. 
The data uncertainty component 
$\text{Var}_{D_N}(\mathbb{E}_\gamma[\hat{y}|\vec{x}]| \mathcal{D}_N)$, which captures how predictions 
would vary under different training datasets from the same distribution, is 
entirely absent from the Deep Ensemble's epistemic estimate.
\end{titledbox}

 While Deep Ensembles might explore multiple (local) optima of the training loss by using different random seeds \cite{wilson2020bayesian},
 Bayesian approaches attempt to explicitly approximate the posterior over parameters,
 which can yield more principled estimates of epistemic uncertainty.
  However, the practical methods used to approximate the posterior distribution also face restrictions:
  Approximate inference methods like the Laplace approximation or variational inference often capture only
   a \emph{single} local mode \cite{wilson2020bayesian}, employ simplifying assumptions \cite{psaros2023uncertainty},
  or can be computationally costly in high-dimensional weight spaces. Other issues arise with methods that try to estimate the second-order distribution directly via loss minimisation, such as Deep Evidential Regression \cite{amini2020deep}. 

The theoretical analysis above raises an immediate question: Given a second-order uncertainty estimation method, how can one assess 
which components of epistemic uncertainty it represents? 
In the absence of observable ground-truth uncertainty, epistemic uncertainty is typically evaluated
through downstream tasks \cite{gawlikowski2023survey}: \textbf{out-of-distribution detection} \cite{hendrycks2017baseline,ovadia2019can,KopetzkiCZGG21},
\textbf{robustness to adversarial attacks} \cite{MalininG19,SmithG18}, \textbf{selective prediction} \cite{geifman2017selective} or
\textbf{active learning} \cite{NguyenSH22}. These evaluation paradigms have important limitations. First, they conflate multiple uncertainty sources: out-of-distribution detection performance depends on
both epistemic uncertainty quality and the specific distributional shift \cite{ovadia2019can},
while selective prediction benefits from total uncertainty rather than epistemic uncertainty alone \cite{hofman2025uncertainty}.
Second, task performance does not isolate whether the decomposition is faithful, since a method
may achieve good performance through mechanisms unrelated to true epistemic uncertainty.
More fundamentally, \citet{li2025out} argue that out-of-distribution detection methods may be generally ill posed:
uncertainty-based methods conflate high uncertainty with being far from training data, while feature-based methods
conflate feature-space distance with it.
Finally, \textbf{none of these evaluation methods assess whether epistemic uncertainty captures
the components we identify as missing}.

 We argue that separately quantifying procedural and data uncertainty is only possible through statistical simulation, 
where repeated sampling from the data-generating process can be performed or approximated. Following \citet{juergensepistemic}, we formalize 
this through the notion of a \emph{reference distribution}.
\begin{definition}
\label{def: reference distribution}
Consider a hypothesis space of models with $\phi \in \Phi$ model parameters and
 $\gamma \in \Gamma$ procedural parameters. We define the \textit{reference distribution} w.r.t. $P^N$ and $P^{\gamma}$ as
\begin{align*}
    q_{N, \gamma}(\vec{\theta}|\vec{x}, \phi) &:= \mathbb{P}(\vec{\hat{\theta}}_{N,\gamma}(x) = \vec{\theta}) \\&= \int_{\Gamma}\int_{\mathcal{X} \times \mathcal{Y}} \mathbb{I} \Big[(\vec{\hat{\theta}}_{N, \gamma}(x)
   = \vec{\theta})\Big] d P^N dP^\gamma,
\end{align*}
where $\vec{\hat{\theta}}_{N, \gamma}$ is the empirical estimate of $\vec{\theta}$,
obtained via empirical risk minimization.
\end{definition}
Intuitively, the reference distribution captures the full estimation uncertainty by marginalizing over both data uncertainty (variation across 
datasets $\mathcal{D}_N \sim P^N$) and procedural uncertainty (variation 
across $\gamma \sim P^\gamma$). Its variance decomposes exactly according
to Eqn.~\ref{eq: variance decomposition mse}, enabling direct comparison with the 
epistemic estimates produced by any second-order method.
In practice, the reference distribution is estimated by training models 
on $n_d$ independent datasets with $n_\gamma$ random initializations each, 
yielding $n_d \times n_\gamma$ models whose empirical variance approximates 
the theoretical quantity. For real-world datasets where sampling from the data-generating distribution is infeasible, disjoint partitioning provides a practical approximation (see Appendix~\ref{app: reference distribution details}).

\begin{figure}[t]
    \centering
    \includegraphics[width=\linewidth]{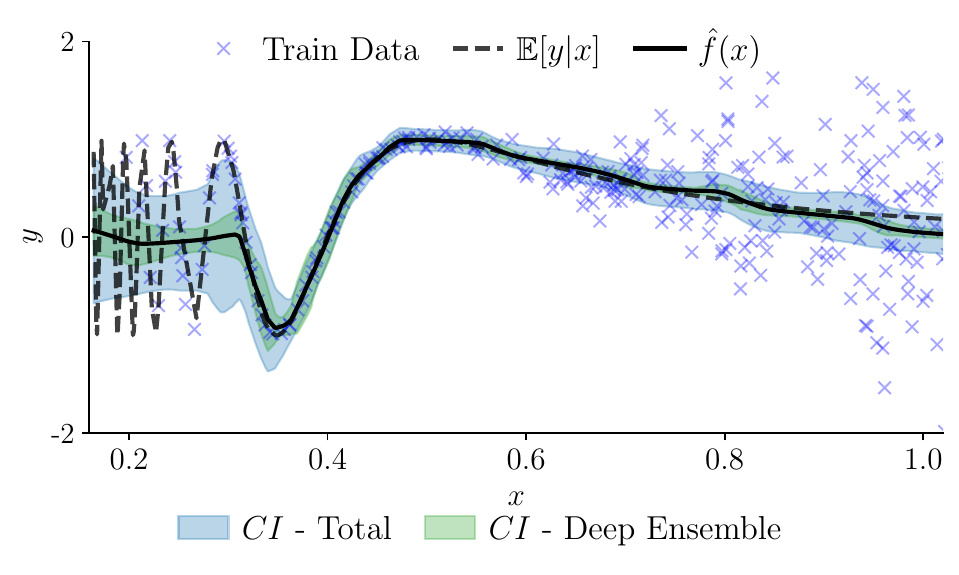}
    \caption{Comparison of reference distribution and Deep ensemble estimates. The confidence intervals of procedural and data variance as estimated by the reference distribution are shown in blue, while 
    the Deep Ensemble estimate of epistemic uncertainty are shown in green.
    }
    \label{fig:evaluation_de_reference}
\end{figure}

\begin{titledbox}{Running Example - Deep Ensembles}


We evaluate a Deep Ensemble of $10$ members on the synthetic data example described in Eqn.~\ref{eq: sine data problem}. The reference distribution is estimated using $n_{d}=20$ resampling steps and $n_{\gamma}=10$ separate models are trained. The full experimental details are provided in Appendix~\ref{app: reference distribution details}. Figure~\ref{fig:evaluation_de_reference} illustrates that Deep Ensembles capture only procedural uncertainty, underestimating the variance from the reference distribution.
\end{titledbox}

Further results on synthetic and real-world data for additional second-order models (Appendix \ref{app: synthetic data experiments}-\ref{app: real data experiments}) reinforce our arguments: Figure \ref{fig:sine_uncertainty_relations_grid} demonstrates that across all evaluated methods, epistemic uncertainty estimates fall systematically below the estimate of the reference distribution. Notably, Bayesian methods with richer posteriors produce higher epistemic estimates in low data regimes. However, this rather reflects their failure to learn the underlying data-generating distribution than the correct variance.

As bootstrapped ensembles are trained on resampled versions of $\mathcal{D}_N$, one might expect
them to better capture the data uncertainty $\text{Var}_{\mathcal{D}_N}(\mathbb{E}_{\gamma}[\hat{y}|\vec{x}] | \mathcal{D}_N)$, since under regularity conditions, the bootstrap variance consistently approximates the true sampling variance \cite{efron1992bootstrap}. Recent work \cite{jain2025frequentist} decomposes bootstrap-based epistemic uncertainty estimates into data and procedural components. However, bootstrap-based estimates are inherently limited by the empirical support of the dataset and the estimation is only approximate in the finite sample regime. In our experiments (Appendix~\ref{app: synthetic data experiments}-\ref{app: real data experiments}) we find that the total epistemic uncertainty as given by the reference distribution is still underestimated.

\section{Alternative views}
\label{sec: alternative views}
Our position intersects with several
 lines of critical and constructive work. We review the dominant 
 paradigm of second-order uncertainty representations and its documented limitations. We also examine popular alternative frameworks. 

\subsection{Second-order uncertainty quantification methods}
\label{sec: methods for uq}
The prevailing approach represents epistemic uncertainty through second-order distributions over predictive distributions,
from which aleatoric and epistemic components are extracted via decomposition formulas \citep{hullermeier2021aleatoric, gawlikowski2023survey}.
 Bayesian neural networks \cite{neal2012bayesian}
form a principled approach to quantify epistemic uncertainty by modelling a prior and posterior distribution over the neural network's weights.
Due to the intractability of the posterior distribution induced by this hypothesis space \cite{kendall2017uncertainties},
one typically samples from it, using different approximation techniques. The second-order distribution 
$q(\vec{\theta}|\vec{x})$ is then approximated via Monte Carlo integration. \textbf{Variational inference} \cite{jordan1999introduction,blundell2015weight,blei2017variational}
considers a variational (simpler) distribution
and optimizes the lower bound of the marginal likelihood. However, posterior quality depends critically on the variational approximation and prior specification.
\textbf{Monte Carlo Dropout} \cite{gal2016dropout} can also be understood as a Bayesian approximation, where one trains a
single neural network with dropout rate $\rho$ and performs multiple stochastic forward passes at inference time with the same dropout rate.
\textbf{Spectral-Normalized Gaussian Processes} \cite{liu2020simple} represent the second-order distribution
by approximating a Gaussian distribution over the classifier output. \textbf{Stochastic Weight Averaging Gaussian} \cite{maddox2019simple}
fits a Gaussian distribution over the model weights by using the stochastic gradient descent iterates to derive estimates for moment and covariance.
Similarly, the \textbf{Laplace approximation} \cite{mackay1992practical,daxberger2021laplace} yields a Gaussian approximation by building a Taylor
expansion around the maximum-a-posteriori estimate.
\textbf{Deep Ensembles} \cite{lakshminarayanan2017simple} train a number of different models on the same dataset,
but with different weight initializations. The resulting predictions can be seen as samples
from an approximate posterior, thereby forming a Bayesian approximation.
In \textbf{Evidential Deep Learning} \cite{ulmerprior,sensoy2018evidential,malinin2018predictive},  the parameters of the second-order distribution are directly learned via loss minimization.
 \citet{sensoy2018evidential} place Dirichlet distributions
 over class probabilities for classification, while \textbf{Deep Evidential Regression} \cite{amini2020deep} extends
  this to regression using the Normal-Inverse-Gamma as conjugate prior distribution.

\subsection{Alternative uncertainty representations}
Given the documented limitations of second-order distributions,
 several alternative frameworks have emerged.
  We focus on deterministic methods based on distances or direct loss prediction, which have gained traction for computational efficiency and competitive
   empirical performance, then briefly discuss credal sets and conformal prediction.

\textbf{Deterministic methods.}
Deterministic  uncertainty quantification methods \cite{postels2022practicality} quantify
epistemic uncertainty using either the distribution of latent representations learned by the model,
or by directly predicting the loss of the network's predictions. \textbf{Deterministic Uncertainty Quantification} \cite{van2020uncertainty}
replaces the softmax layer with an RBF-like output layer around class centroids,
thereby measuring distance to the centroids as a proxy for uncertainty. The \textbf{Mahalanobis method} 
\cite{lee2018simple} captures epistemic uncertainty post-hoc by measuring the distance from the learned class-conditional distributions
 in feature space. \textbf{Direct Epistemic Uncertainty Prediction} \cite{lahloudeup} 
quantifies epistemic uncertainty as the excess risk of the predictor, i.e. the gap between pointwise expected 
loss and the loss of the Bayes predictor. It is learned via a secondary model trained on out-of-sample errors.
Distance-based methods implicitly capture a form of distributional uncertainty:
 distance from training data correlates with
 regions where predictions are unreliable. However, \textbf{they do not explicitly decompose epistemic uncertainty into data and procedural components, nor do they account for systematic bias}.
 On the other hand, methods based on predicting the loss of the model are robust to model bias, but 
 require training and maintaining an additional error-prediction model.

 \textbf{Credal sets and imprecise probabilities.}
Rather than placing distributions over distributions, credal sets represent epistemic uncertainty as convex sets
 of probability distributions \citep{walley1991statistical}. 
Recent work has operationalized credal sets in deep learning \cite{caprio2024credal,capriocredal}. Theoretically, credal sets offer a more expressive uncertainty 
language, however, they are not yet fully established in mainstream machine learning practice.
 Computational challenges arise from working with sets rather than point estimates,
  and the connection to standard training procedures remains less direct than for probabilistic methods.

\textbf{Conformal prediction.}
Conformal prediction \citep{vovk2005algorithmic,angelopoulos2023conformal}
 produces prediction sets with finite-sample coverage guarantees, which hold for any sample size and any underlying model.
However, it addresses a different problem than epistemic uncertainty quantification, since it
does not decompose uncertainty into aleatoric and epistemic components. Conformal methods are thus complementary to, rather than a replacement for, principled epistemic uncertainty estimation.

\section{Conclusion and call to action}
Epistemic uncertainty is generally not well defined and is still a heuristic concept for machine learning models,
 leading to interpretability issues for uncertainty estimates. Our analysis reveals that the prominent second-order uncertainty estimation methods 
 systematically fail to capture what they claim to measure. Two mechanisms drive this failure: First, the predictive bias
 contaminates the aleatoric-epistemic uncertainty decomposition and should be evaluated as an epistemic component.
Second, epistemic uncertainty estimates typically only reflect parts of the true variance, often 
capturing procedural rather than data uncertainty.

\textbf{Call to action.}
We advocate for a more fine-grained taxonomy that explicitly distinguishes between
 model uncertainty, comprising approximation error plus estimation bias, and
 estimation uncertainty, comprising data and procedural components. New methods should 
 be evaluated not only on downstream tasks but also against reference distributions that decompose
  variance into different sources. Claims of epistemic uncertainty estimation should specify
   which components are captured and acknowledge which are not.
In safety-critical applications, we recommend: (1) assessing model bias separately through held-out validation or domain expertise, (2) treating epistemic uncertainty estimates as lower bounds on reducible uncertainty rather than complete characterizations, and (3) using multiple complementary uncertainty methods, as different methods capture different uncertainty components.

\newpage

\bibliography{example_paper}

\bibliographystyle{icml2026}

\newpage
\appendix
\onecolumn
\section{Impact of recent literature}
\label{app:impact of literature}

In recent years, uncertainty disentanglement has gained importance in the machine learning literature. Considering the following collection of highly relevant papers, they have received almost 40,000 citations according to Google Scholar:
\cite{der2009aleatory,kendall2017uncertainties,hullermeier2021aleatoric,mackay1992practical,depeweg2018decomposition,malinin2018priornet,MalininG19,amini2020deep,lakshminarayanan2017simple,wilson2020bayesian,gal2016uncertainty,charpentier2020posterior,charpentier2021natural,stadler2021graph,lee2018simple,liu2020energy,van2020uncertainty,sun2022out,deng2023uncertainty}

\section{Incomplete uncertainty estimators}
\subsection{Bias aleatoric uncertainty estimators}
\label{app:biased aleatoric uncertainty}
For a data-generating process such that $y_i = f(\vec{x}_i) + \epsilon_i$, where $\epsilon_i \sim \mathcal{N}(0, \sigma^2(\vec{x}_i))$ is data-dependent (\textit{heteroscedastic}) noise, and $f$ is a deterministic function,
such that $p(\cdot|\vec{x})$ forms the density of a Gaussian distribution with mean $f(\vec{x})$ and variance $\sigma^2(\vec{x})$. There are multiple ways to model aleatoric uncertainty. One popular way in the literature to do so \cite{zhang2024one} is using the variance attenuation \cite{lakshminarayanan2017simple,amini2020deep,valdenegro2022deeper}, which means to train a neural network with two heads (one that outputs the predictive mean and the other that outputs the predictive variance) using a loss function that reflects the likelihood of the observed data:
\begin{equation}
\label{eq: aleatoric unc va method}
    \ell_{va}=\frac{1}{M}\sum^N_{i=1}\frac{(y_i-\hat{f}(x_i))^2}{2\hat{\sigma}^2(x_i)}+\frac{\log(\hat{\sigma}^2(x_i))}{2}.
\end{equation}
We rely on the derivation presented by \cite{zhang2024one} to show how Eqn.~\ref{eq: aleatoric unc va method} does not produce an unbiased estimator of the aleatoric uncertainty $\sigma^2(\vec{x})$, and refer the reader to its paper for all the details. In this setting, the expected value of the predictive variance becomes
\begin{equation}
    \mathbb{E}[\hat{\sigma}^2(x)] = (f(x)-\hat{f}(x))^2+\sigma^2(x).
\end{equation}
Where one can see that the expected aleatoric noise differs from the true aleatoric noise by a square difference of the predictive mean and the true label.

While the original contribution of \cite{nix1994estimating} suggest to perform a warm-up period, one can see that this modification to the training process would improve the mean prediction, by fitting the model in a simpler loss landscape, and therefore the accuracy of $\hat{f}$, leaving the expected value of the predictive variance unchanged. Other methods have been proposed to produce a better estimate of aleatoric uncertainty. \cite{seitzer2022pitfalls} proposed adding an extra term $\lfloor\hat{\sigma}^{2\beta}(x)\rfloor$ to the second term of $\ell_{va}$, where $\lfloor\cdot\rfloor$ denotes the stop gradient operation. However, this modification does not change the expected value of the estimator, it just restricts the learning path of the predictive mean. \cite{sluijterman2025confident} proposed a method that divides the training process into two parts, by creating some modified datapoints for the training of the second part. This process improves mean predictions and the potential to avoid overfitting, while the bias of the expected aleatoric uncertainty estimation is still there when the predictive mean deviates from the true value.

\subsection{Regularization effects on predictive variance}
\label{app:shrinked epistemic uncertainty}
Neural networks are trained with multiple sources of regularization due to the limited sample size (e.g. early stopping, weight decay, dropout). A body of works in the literature, from textbooks to recent machine learning papers \cite{larsen1994generalization,nusrat2018comparison,ninness2004effect,girosi1995regularization}, show how regularization reduces predictive variance by constraining how strongly the learned model can respond to finite-sample fluctuations in the training data, which happens through parameter shrinkage, effective dimensionality reduction, and smoothing of the learned function.

By reducing the variance, the third term of Eqn.~\ref{eq: bias-variance} shrinks and the second term of the same equation increases, as bias increases from regularization. Our argument is key in this aspect: if we aim to capture epistemic uncertainty, looking only at the variance, which has been affected by the training process, gives an incomplete picture of the whole epistemic uncertainty.

\section{Experimental details}
\label{app: experimental details}

\subsection{Reference distribution approximation}
\label{app: reference distribution details}
\paragraph{Synthetic dataset.} To obtain an empirical approximation of the reference distribution (see Definition ~\ref{def: reference distribution}), we sample $20$ different datasets from the known data-generating process to approximate the data uncertainty component of the epistemic uncertainty. For each dataset, we train $10$ different MLP models with different random weight initializations, which allows us to obtain a procedural uncertainty estimate. Each MLP uses the whole dataset. The MLP models have $4$ hidden layers with $100$ neurons each. We use a learning rate of 0.009 and train the models for $500$ epochs. We repeated the process for different training dataset sizes: $50, 100, 500$ with batch sizes $32, 32, 64$ respectively. Seeds for the datasets are $\{7, 42, 2024, 123, 999, 50, 100, 150, 200, 250, 300, 350, 400, 450, 500, 550, 600, 650, 700, 750\}$.

\begin{figure*}[t]                  
\includegraphics[width=\textwidth]{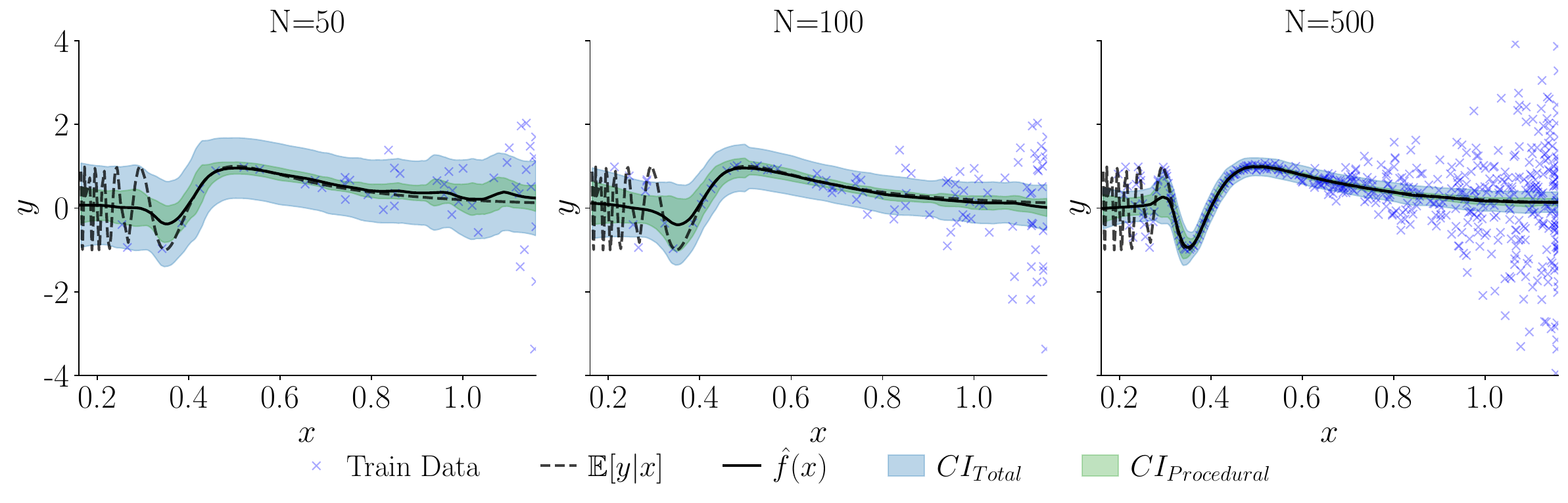}
    \caption{Estimated procedural and data uncertainty of the reference distribution for increasing sample sizes. The reference distribution is estimated by resampling training data and procedural parameters, using $n_{\gamma}=10$ and $n_d =20$. Procedural and data uncertainty are calculated using the decomposition formula in Eqn.~\ref{eq: bias-variance}. Even when the model predictions (complete line) have an important deviation (bias) from the ground-truth expected value (dashed line), the epistemic uncertainty estimate decreases with more data.}
    \label{fig:procedural-var-ref}
\end{figure*}

 Figure~\ref{fig:procedural-var-ref} shows the obtained procedural and data uncertainty of the reference distribution for different training dataset sizes. First, we see that the total variance of the reference distribution decreases when sample size increases, supporting the claim that the reference distribution properly represents data uncertainty. Second, we see that the estimate of the procedural uncertainty also decreases with an increasing sample size, which is not unexpected, because initial procedural configurations become less important when more data is observed. Third, in the region with high epistemic uncertainty, the reference distribution has a very large bias, due to the underlying complexity of the data-generating function. The bias is in fact, so high that aleatoric uncertainty cannot be correctly estimated anymore.
This last finding confirms our central position: Any claim to properly represent epistemic uncertainty must examine not only how different sources of uncertainty are modeled, but also whether bias is systematically distorting the partitioning of uncertainty. When the bias is high, estimates of both epistemic and aleatoric uncertainty can be completely misleading.

\paragraph{Real dataset.} To estimate the reference distribution as described in Section \ref{sec: data and procedural uncertainty}, we partition the training dataset into $100$ disjoint subsets of $12,067$ observations each (i.e., $n_{d}=100$). This partition allows us to approximate the data uncertainty component from the epistemic uncertainty. We argue that in big datasets, where enough disjoint subsamples can be taken, it is possible to approximate the variance coming from different datasets at training time. For each subset, we train $n_{\gamma}=5$ multilayer perceptrons (MLPs) with different random weight initializations, this part of the process is the same as in the synthetic data setting and results in an approximation of the procedural component of the epistemic uncertainty. This whole process yields a total of $500$ independently trained models. Each model produces predictive means $\hat{y}_{N,\gamma}(\vec{x})$, and by computing the empirical variance across these 500 predictions, we obtain an estimate of $\text{Var}_{\mathcal{D}_N, \gamma}(\hat{y}|\vec{x})$ as defined in Eqn.~\ref{eq: variance decomposition mse}. This procedure operationalizes the reference distribution introduced in Definition \ref{def: reference distribution}: the pseudo-resampling across datasets ($\mathcal{D}_N$) captures data uncertainty while the random initializations account for procedural uncertainty.
Given the large sample size of the dataset, the partitioning approach provides a computationally feasible approximation of repeated sampling from the underlying data-generating process, effectively \textit{emulating} access to its stochastic variability. This design allows us to decompose the epistemic uncertainty component into its data and procedural parts in a controlled manner, while being in a real-world data setup.

\subsection{Synthetic data experiments}
\label{app: synthetic data experiments}
To illustrate the limitations of existing uncertainty quantification methods in producing reliable disentangled estimates, we showcase the behavior of eight representative methods of the three second-order quantification approaches we discuss in the main paper. This section shows the experimental details and results from the uncertainty quantification methods applied to the problem described by Eqn.~\ref{eq: sine data problem}.

\subsubsection{Experimental illustrations}

\begin{figure}
\includegraphics[width=\textwidth]{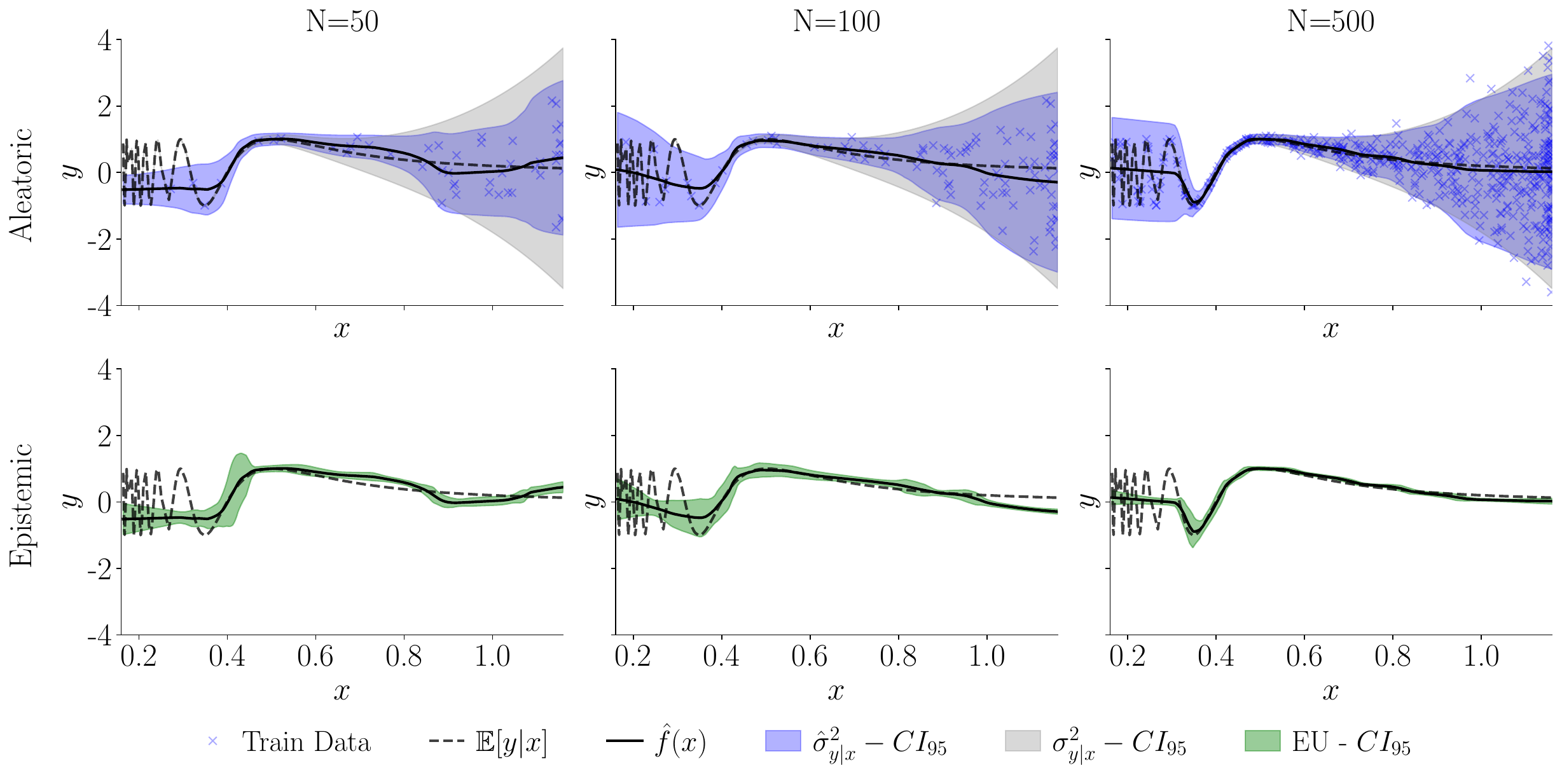}
    \caption{Deep Ensemble estimates, obtained with different sample sizes $N\in \{50,100,500\}$. The top row shows the training data points (blue crosses), the true conditional mean (dashed line), the model predictions (complete line), the $95\%$ confidence interval for aleatoric uncertainty (purple), as well as the true $95\%$ confidence interval of the aleatoric uncertainty (gray). The bottom row shows the estimate of epistemic uncertainty (green).}
    \label{fig:deep-ensemble}
\end{figure}

\begin{figure}[H]    
\centering
\includegraphics[width=0.7\textwidth]{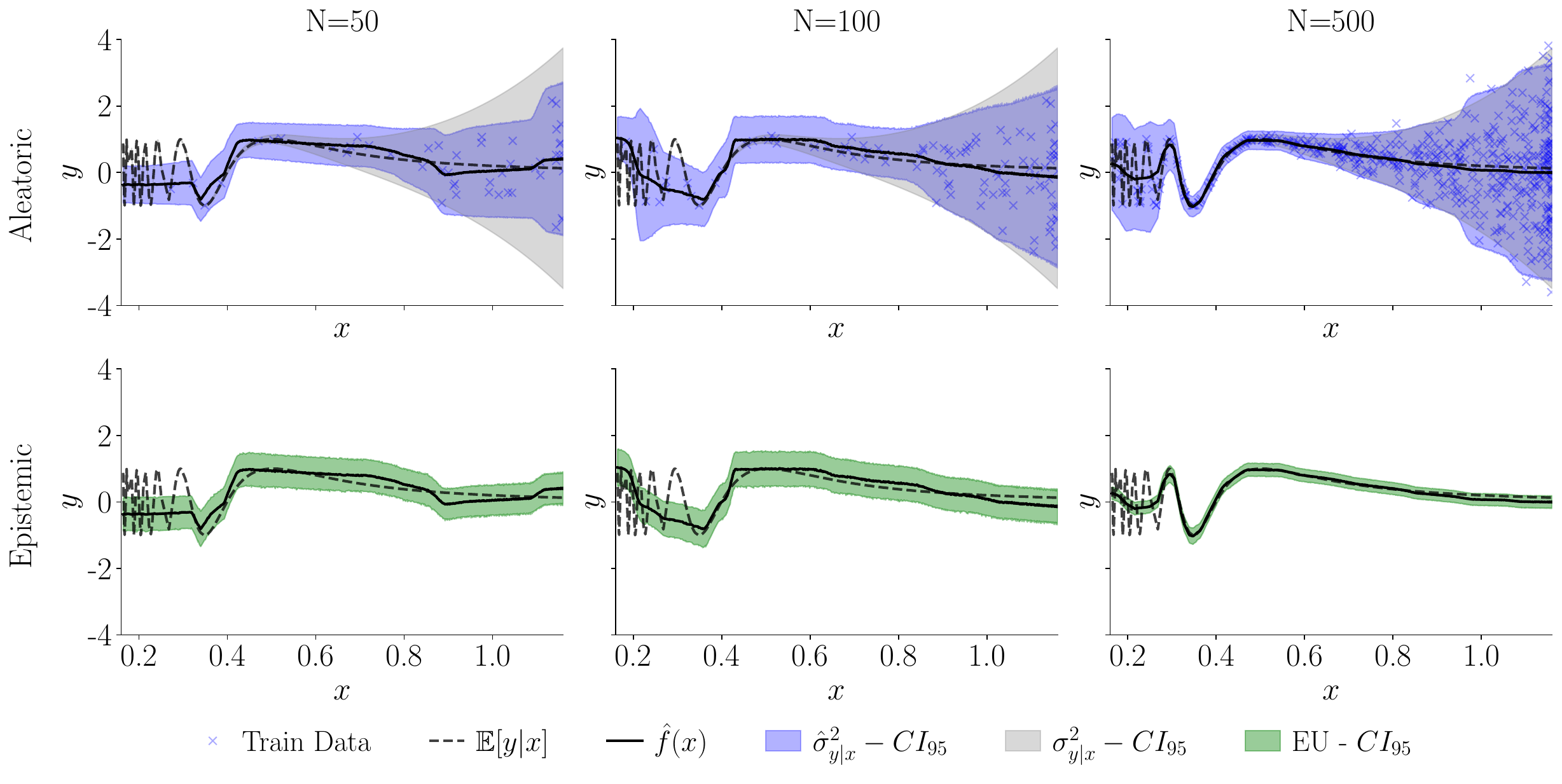}
    \caption{VI estimates, obtained with different sample sizes as in Figure \ref{fig:deep-ensemble}.}
    \label{fig:vi}
\end{figure}

\begin{figure}[H]
\centering
\label{fig:gp}
    \includegraphics[width=0.7\textwidth]{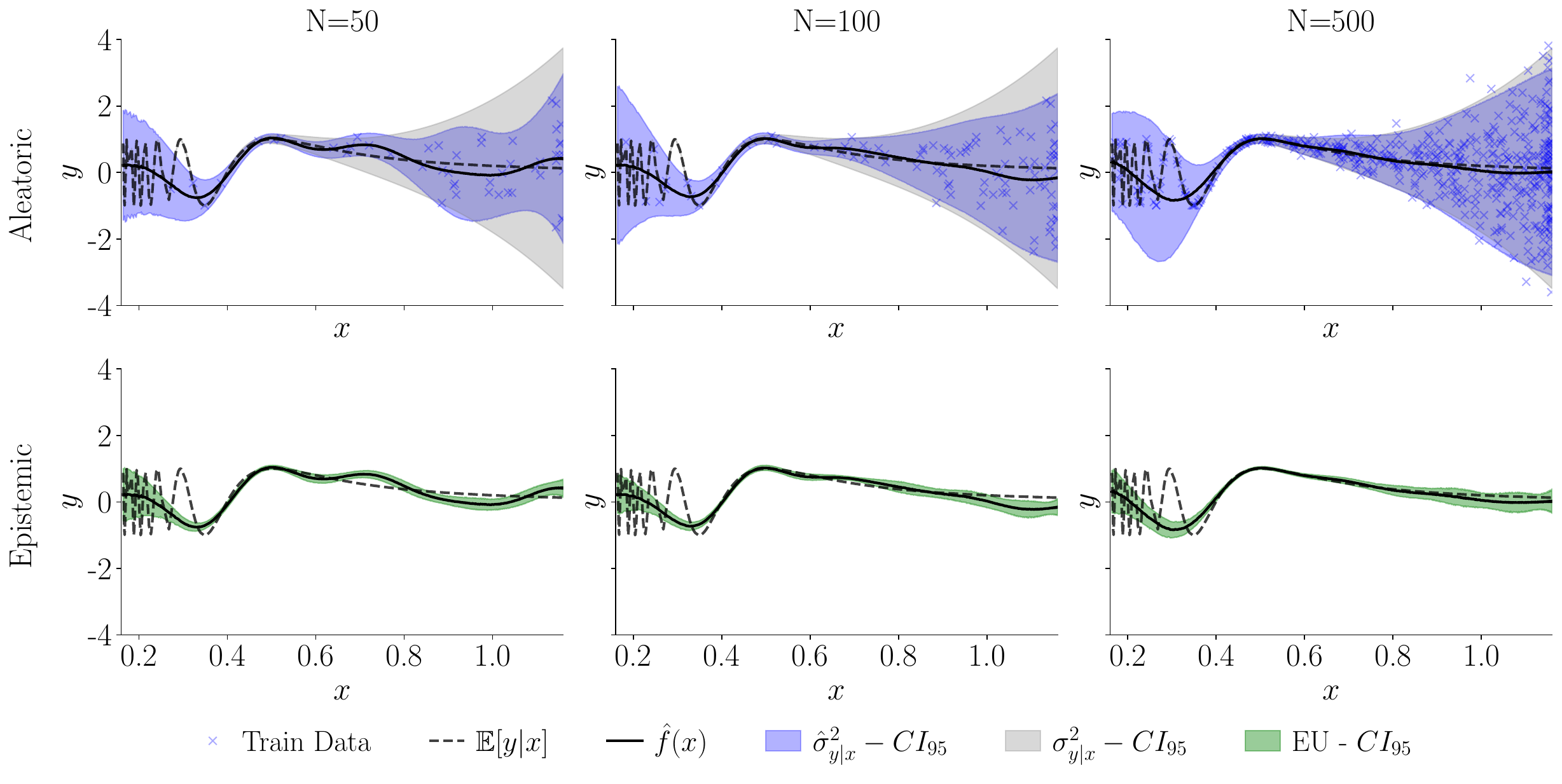}
    \caption{GP estimates, obtained with different sample sizes as in Figure \ref{fig:deep-ensemble}.}
 \end{figure}

\begin{figure}[H]
\centering
\label{fig:bootstrap}
    \includegraphics[width=0.7\textwidth]{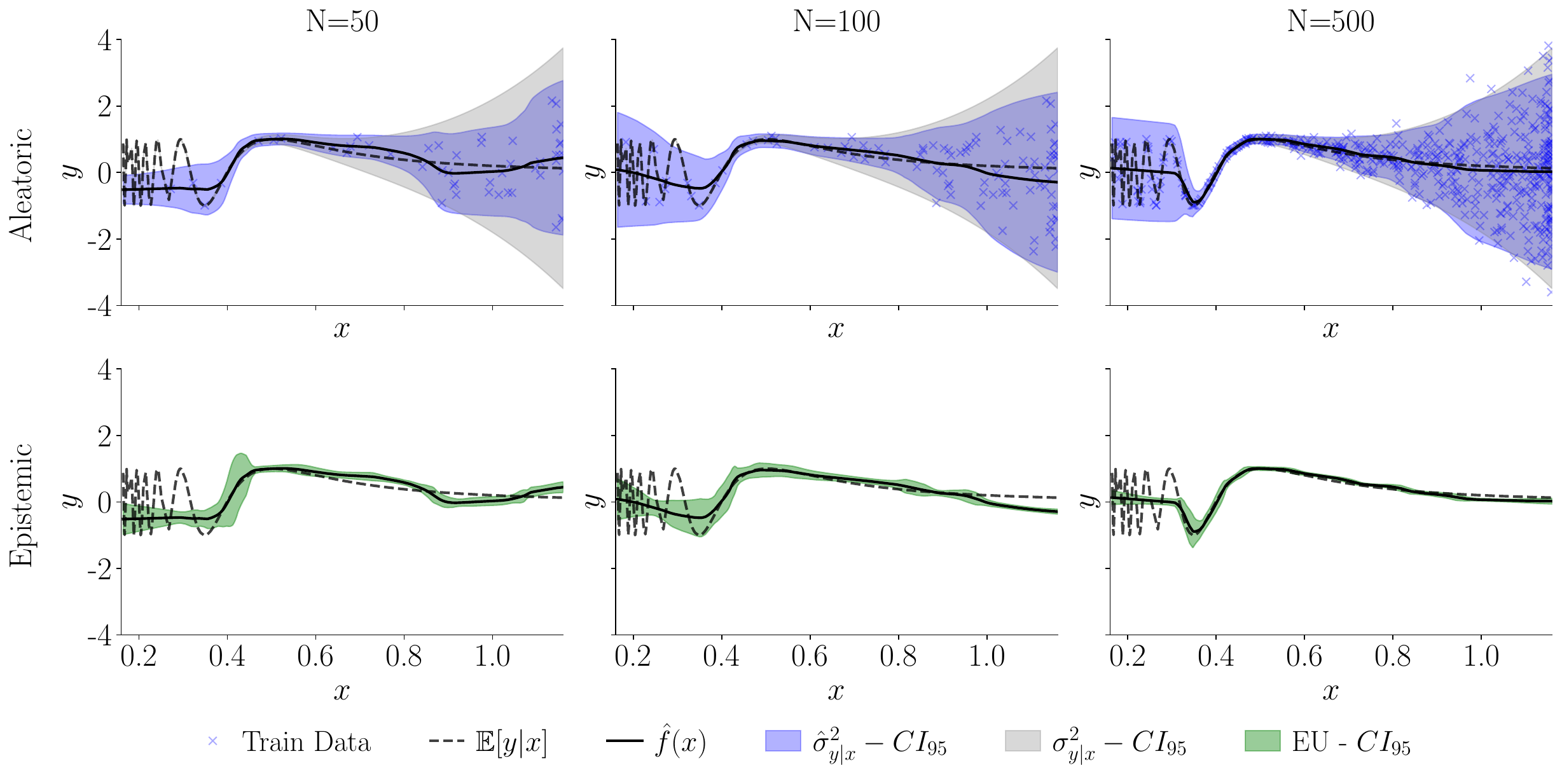}
    \caption{Bootstraped ensemble estimates, obtained with different sample sizes as in Figure \ref{fig:deep-ensemble}.}
\end{figure}

\begin{figure}[H]
\centering
    \includegraphics[width=0.7\textwidth]{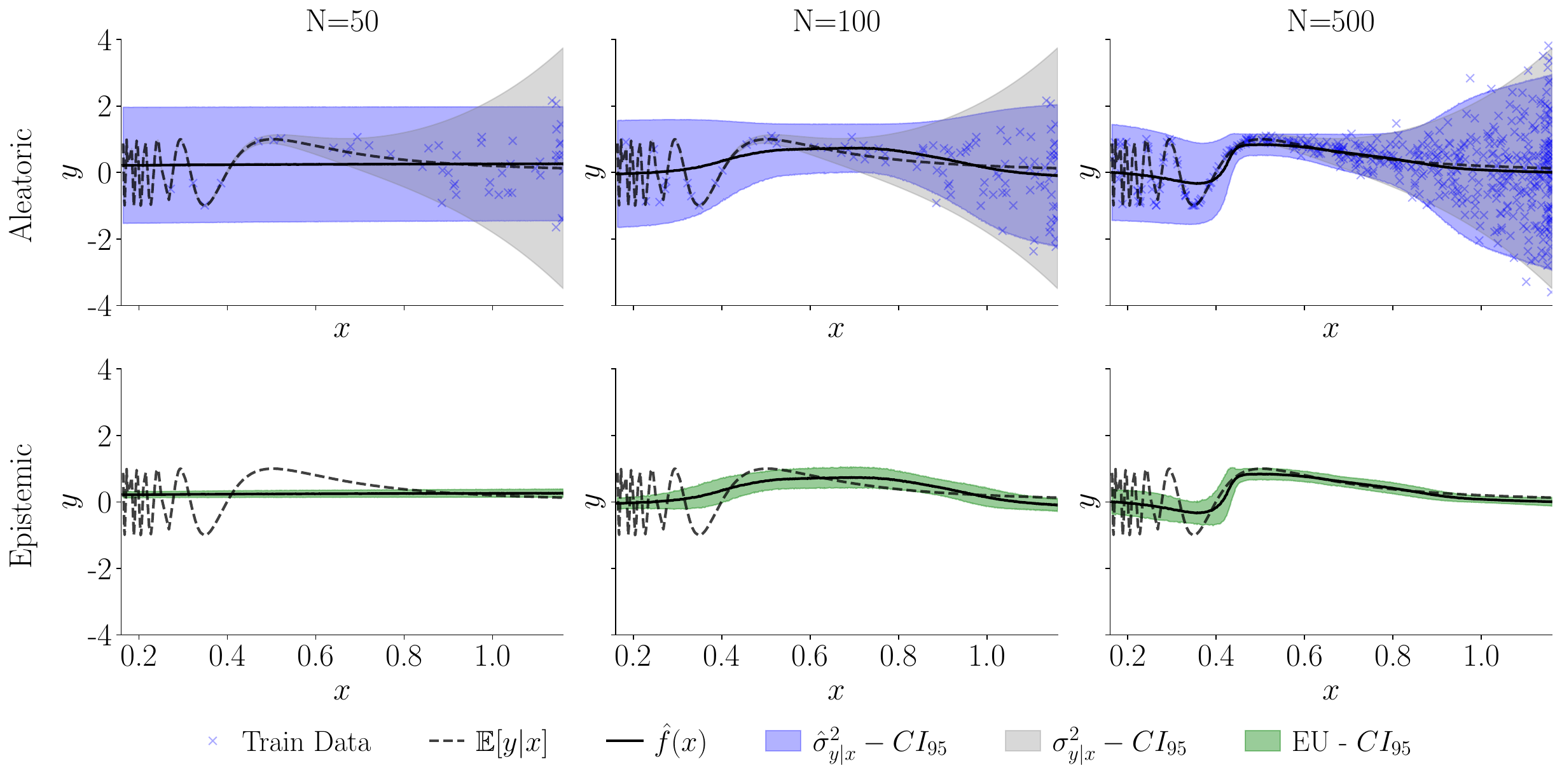}
    \caption{MC Dropout estimates, obtained with different sample sizes as in Figure \ref{fig:deep-ensemble}.}
    \label{fig:mc}
\end{figure}

\begin{figure}[H]
\centering
\label{fig:hmc}
    \includegraphics[width=0.7\textwidth]{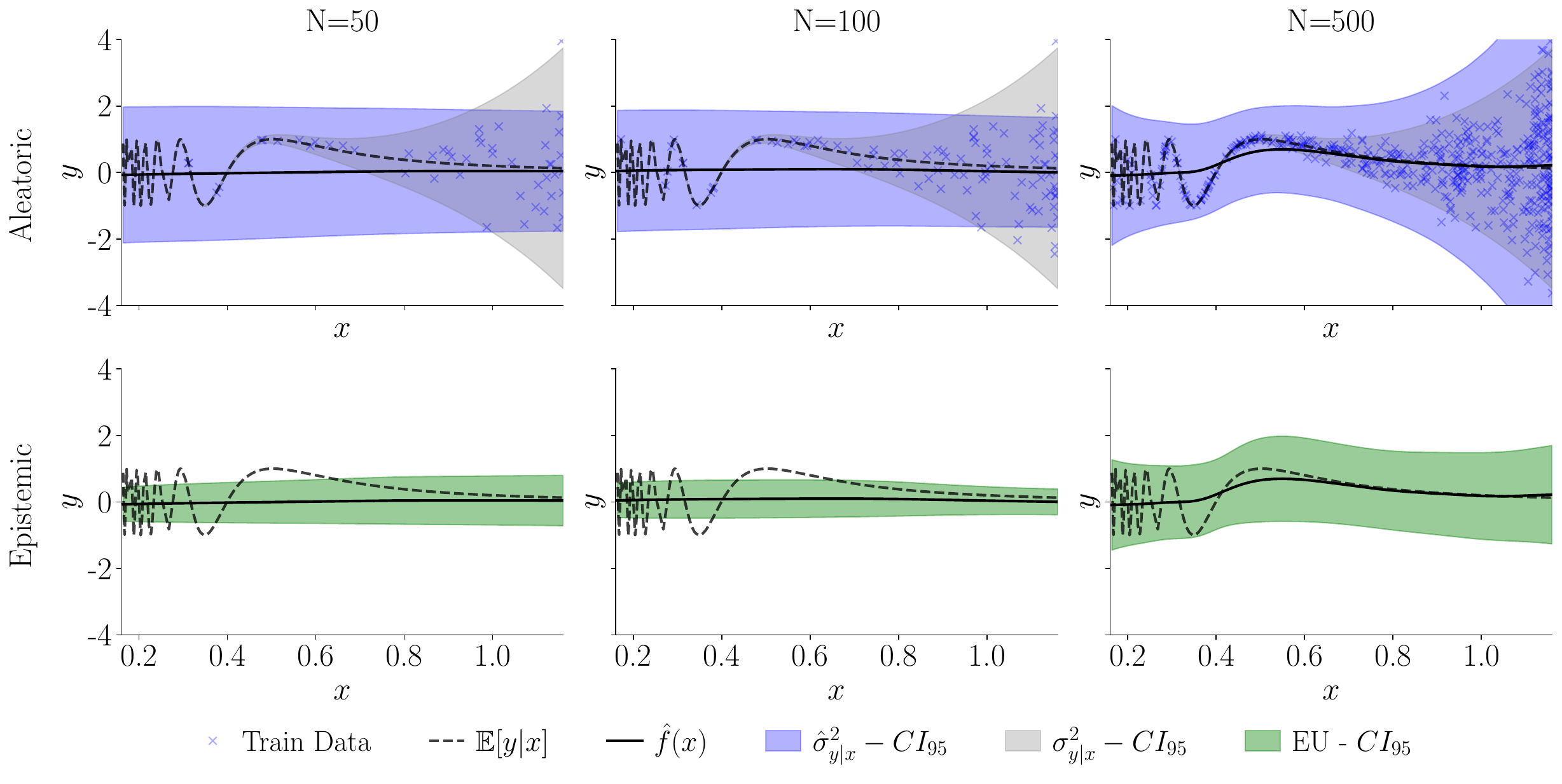}
    \caption{HMC estimates, obtained with different sample sizes as in Figure \ref{fig:deep-ensemble}.}
\end{figure}

\begin{figure}[H]
\centering
\includegraphics[width=0.7\textwidth]{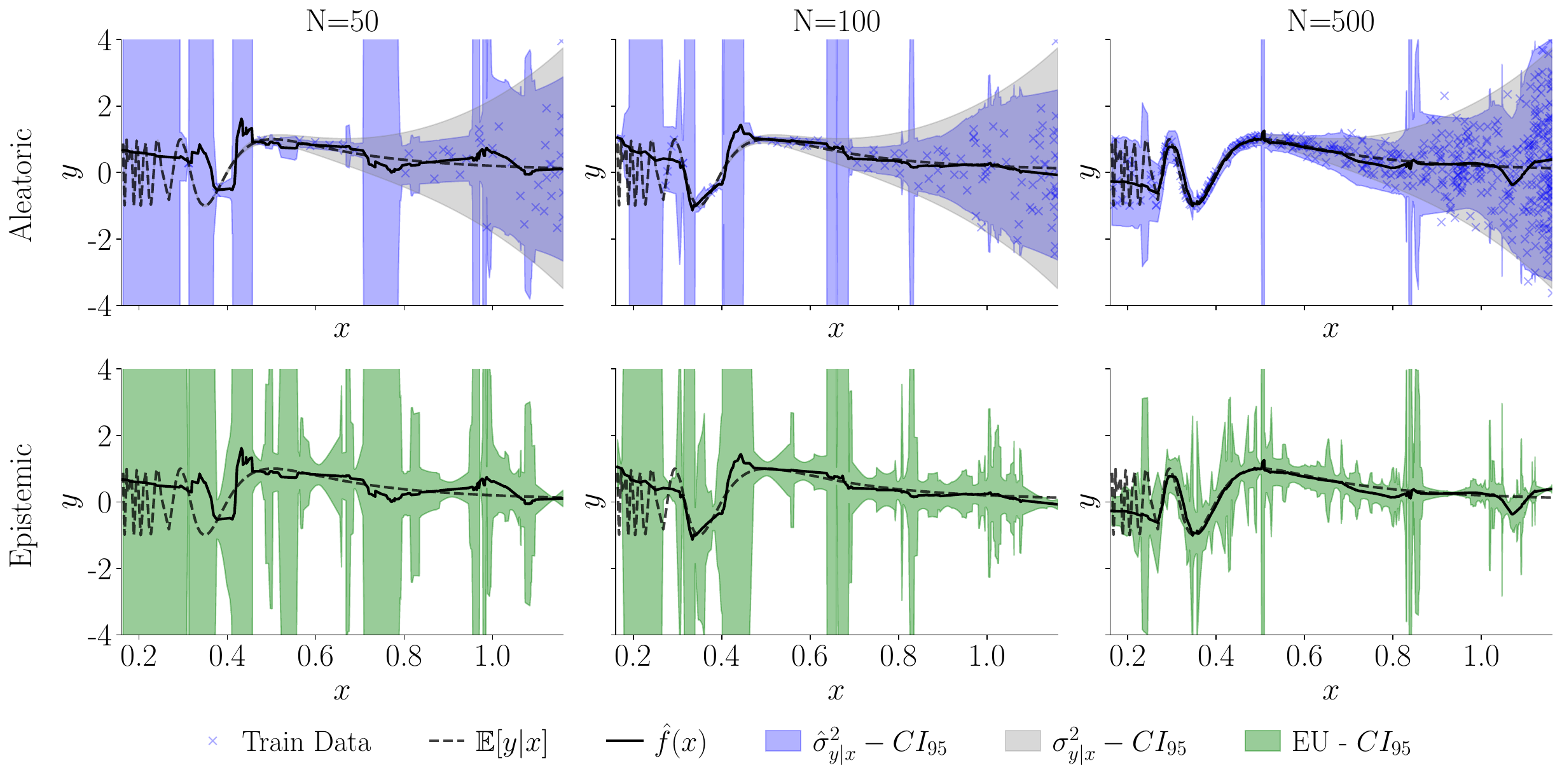}
    \caption{Laplace estimates, obtained with different sample sizes as in Figure \ref{fig:deep-ensemble}.}
    \label{fig:laplace}
\end{figure}

\begin{figure}[H]
\centering
    \includegraphics[width=0.7\textwidth]{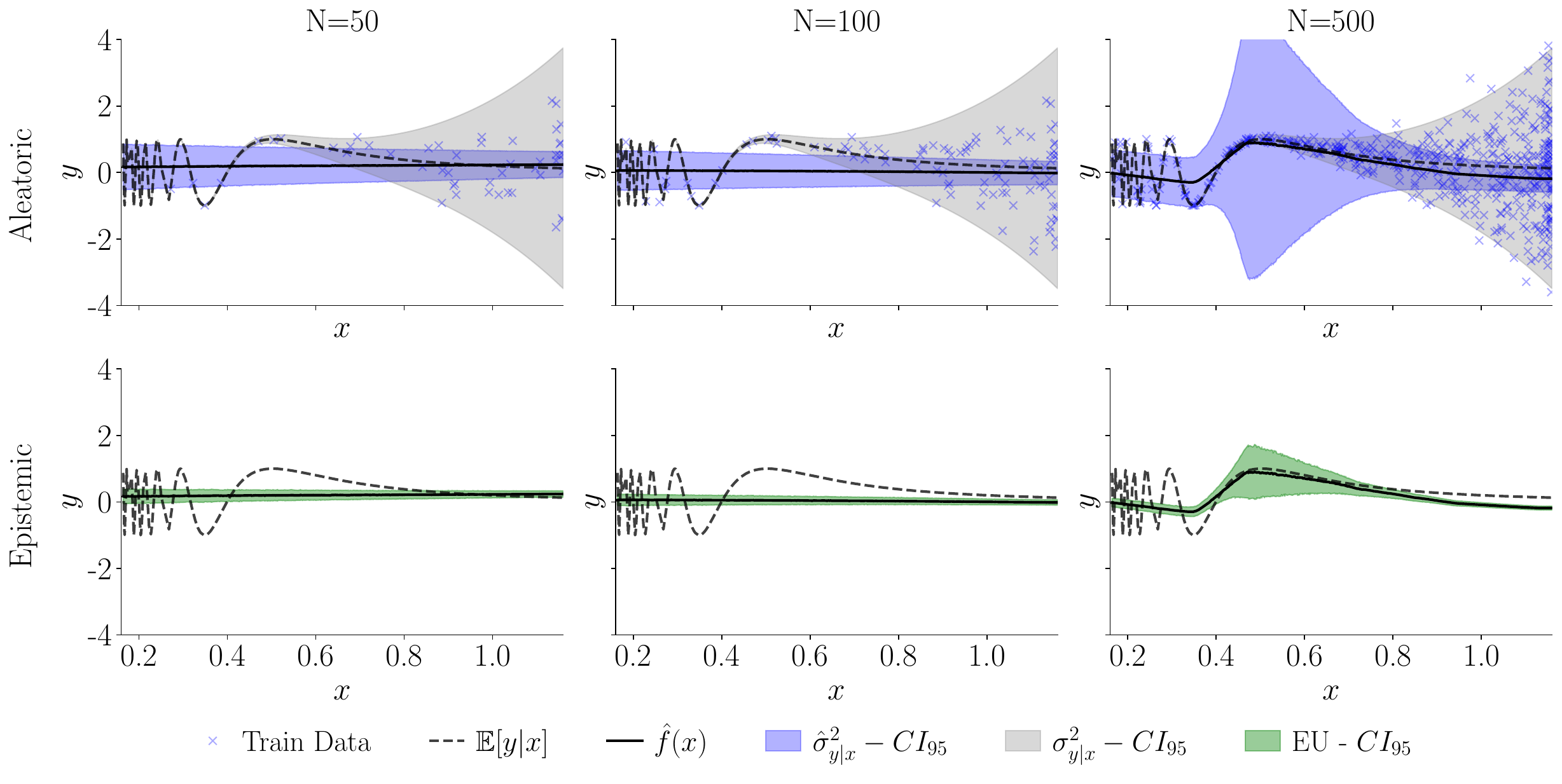}
    \caption{DER estimates, obtained with different sample sizes as in Figure \ref{fig:deep-ensemble}.}
    \label{fig:edl}
\end{figure}

We use the variance-based split of the predictive variance as in Eqn. \ref{eq: variance decomposition mse} to evaluate the methods in the heteroscedastic regression setting.  Figure \ref{fig:deep-ensemble} illustrates that epistemic uncertainty estimates decrease monotonically with sample size, consistent with the behavior observed for the reference distribution. This trend confirms that, as the dataset grows, models become increasingly confident, yet not necessarily more accurate, particularly in difficult-to-learn regions. In these settings, even with abundant data, the model can remain overconfident in systematically biased predictions. Such behavior reveals a key limitation of current epistemic uncertainty measures and disentanglement frameworks: they fail to account for systematic bias, leading to an underestimation of epistemic uncertainty. This pattern is not unique to Deep Ensembles, but is consistently observed across other methods (see Figures \ref{fig:vi}–\ref{fig:edl}), indicating that this limitation is intrinsic to the way epistemic uncertainty is commonly represented rather than a method-specific artifact. Bias is approximated as specified in Eqn.~\ref{eq: bias-variance}: datasets are repeatedly resampled from the true data‐generating process, models are trained on each sample, and the mean deviation of the predictions from the true conditional mean is computed. Table \ref{tab:uncertainties_combined} shows that bias is frequently larger than --and largely uncorrelated with-- the corresponding epistemic uncertainty estimates. These findings underscore our central position: any attempt to disentangle aleatoric from epistemic uncertainty must assess not only how each component is modeled, but also the extent to which systematic bias distorts their partitioning.

\begin{table}[t]
\medskip
\centering
\noindent\textbf{Training sample size: $N = 100$}
\medskip
\begin{tabular*}{\textwidth}{@{\extracolsep{\fill}}lcccc}
\toprule
Model & $\hat{\sigma}(x)$ & $\text{Var}\big[\hat{y}|\vec{x}\big]$ & $\mathbb{E}_{\mathcal{D}_N}[\hat{y}(x)-f(x_i)]$ & $\hat{\sigma}(x)-\sigma(x)$ \\
\midrule
Reference     & 0.83 (0.01)   & 0.041 (0.005)  & 0.084 (0.002) & 0.248 (0.004) \\
Bootstrap     & 0.87 (0.167)  & 0.006 (0.008)  & 0.125 (0.022) & 0.239 (0.027) \\
DeepEnsemble  & 0.87 (0.148)  & 0.007 (0.011)  & 0.147 (0.018) & 0.235 (0.021) \\
EDL           & 0.21 (0.002)  & 0.003 (0.0)    & 0.269 (0.01)  & 0.751 (0.005) \\
ViELBONet     & 0.95 (0.064)  & 0.079 (0.001)  & 0.127 (0.021) & 0.234 (0.022) \\
MCDropoutNet  & 0.90 (0.097)  & 0.009 (0.002)  & 0.190 (0.026) & 0.334 (0.033) \\
HMCNet        & 9.08 (13.734) & 2.131 (0.224)  & 0.179 (0.015) & 7.870 (0.853) \\
LaplaceNet    & 2.43 (5.559)  & 4.211 (3.047)  & 0.133 (0.024) & 0.800 (0.540) \\
GP            & 0.86 (0.048)  & 0.008 (0.001)  & 0.194 (0.013) & 0.308 (0.020) \\
\bottomrule
\end{tabular*}

\medskip
\centering
\noindent\textbf{Training sample size: $N = 500$}

\medskip
\begin{tabular*}{\textwidth}{@{\extracolsep{\fill}}lcccc}
\toprule
Model & $\hat{\sigma}(x)$ & $\text{Var}\big[\hat{y}|\vec{x}\big]$ & $\mathbb{E}_{\mathcal{D}_N}[\hat{y}(x)-f(x_i)]$ & $\hat{\sigma}(x)-\sigma(x)$ \\
\midrule
Reference     & 1.04 (0.016)  & 0.013 (0.001)  & 0.044 (0.001) & 0.076 (0.003) \\
Bootstrap     & 0.98 (0.100)  & 0.003 (0.004)  & 0.085 (0.001) & 0.131 (0.010) \\
DeepEnsemble  & 0.99 (0.136)  & 0.004 (0.005)  & 0.100 (0.002) & 0.122 (0.010) \\
EDL           & 0.67 (0.247)  & 0.013 (0.007)  & 0.168 (0.016) & 0.967 (0.031) \\
ViELBONet     & 1.06 (0.115)  & 0.009 (0.000)  & 0.097 (0.005) & 0.122 (0.012) \\
MCDropoutNet  & 1.06 (0.079)  & 0.008 (0.003)  & 0.119 (0.006) & 0.163 (0.008) \\
HMCNet        & 4.17 (20.356) & 0.378 (0.017)  & 0.129 (0.003) & 1.936 (1.495) \\
LaplaceNet    & 1.29 (3.777)  & 0.487 (3.802)  & 0.115 (0.016) & 0.187 (0.345) \\
GP            & 1.04 (0.069)  & 0.008 (0.003)  & 0.172 (0.005) & 0.152 (0.014) \\
\bottomrule
\end{tabular*}
\caption{Aleatoric and epistemic uncertainty estimates, bias, and distance from the aleatoric uncertainty to ground truth noise across models. "Referece" refers to the variance and predictive mean as estimated by the reference distribution. Values are means over the test set from five runs; standard deviations are in parentheses.}
\label{tab:uncertainties_combined}
\end{table}

\begin{figure}
    \centering
    \begin{subfigure}[b]{0.32\textwidth}
        \includegraphics[width=\textwidth]{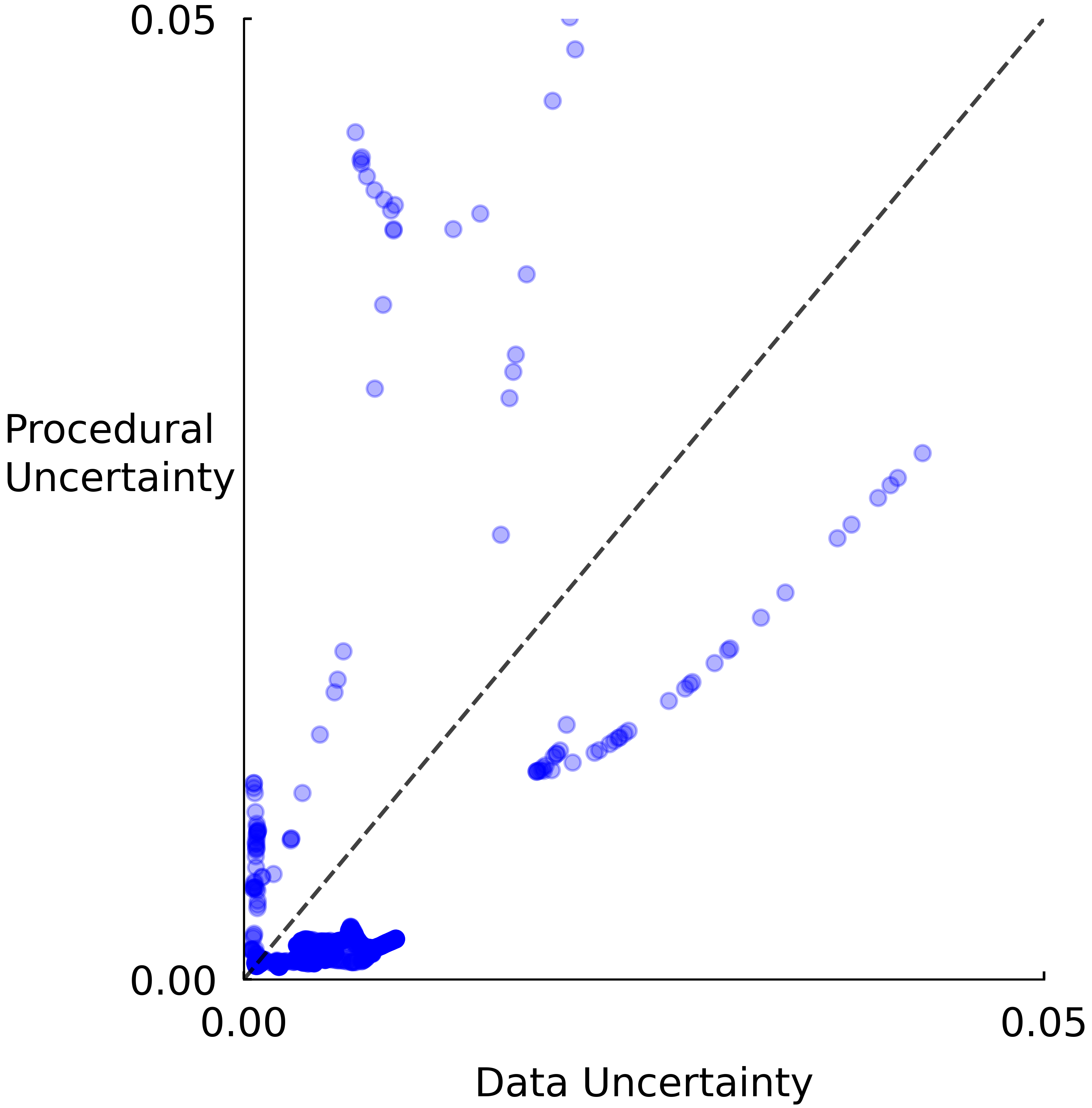}
        \caption{}
        \label{fig:procedural_data_sine}
    \end{subfigure}
    \begin{subfigure}[b]{0.32\textwidth}
        \includegraphics[width=\textwidth]{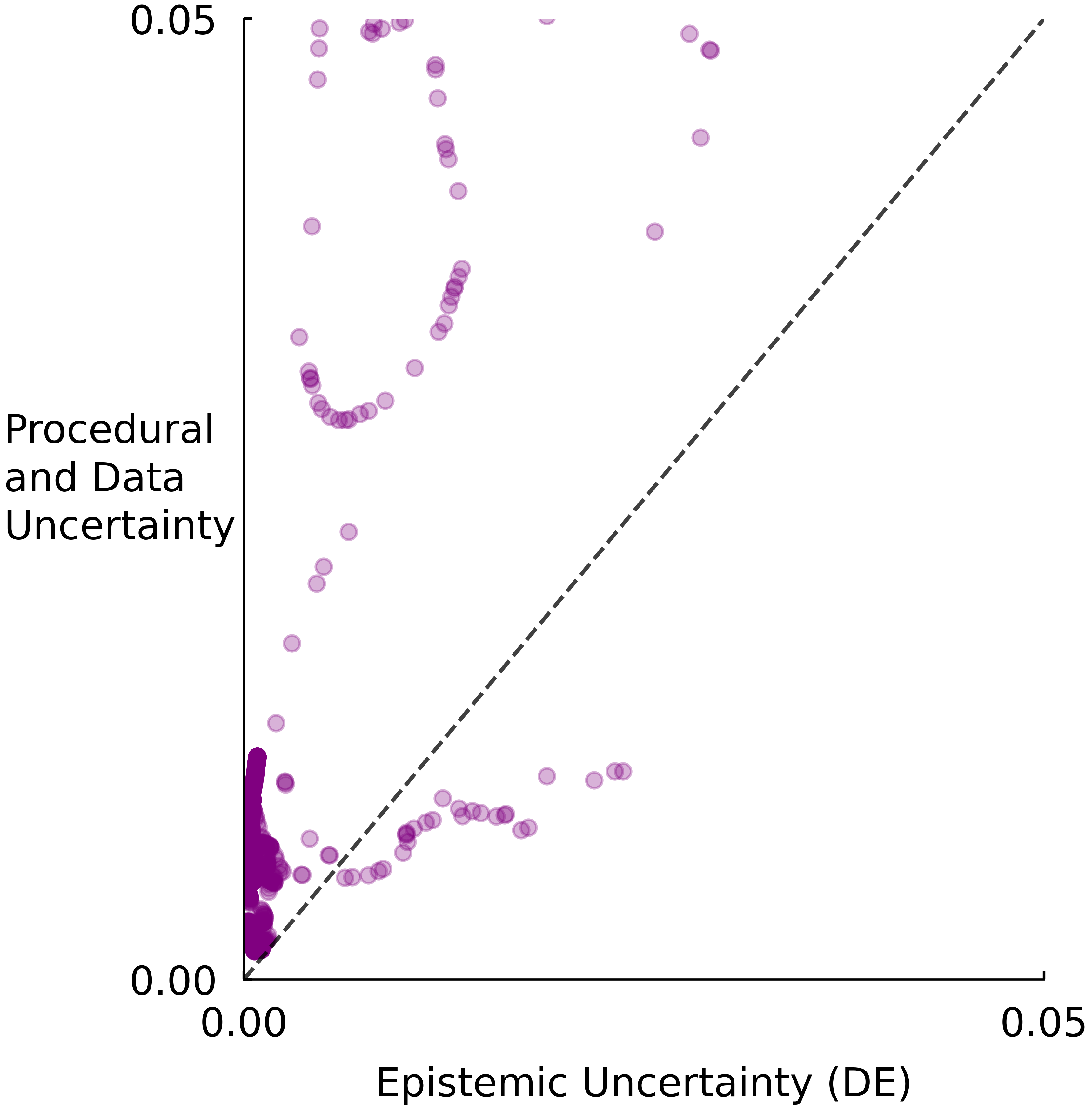}
        \caption{}
        \label{fig:epistemic_sine_comparison_de}
    \end{subfigure}
    \begin{subfigure}[b]{0.32\textwidth}
        \includegraphics[width=\textwidth]{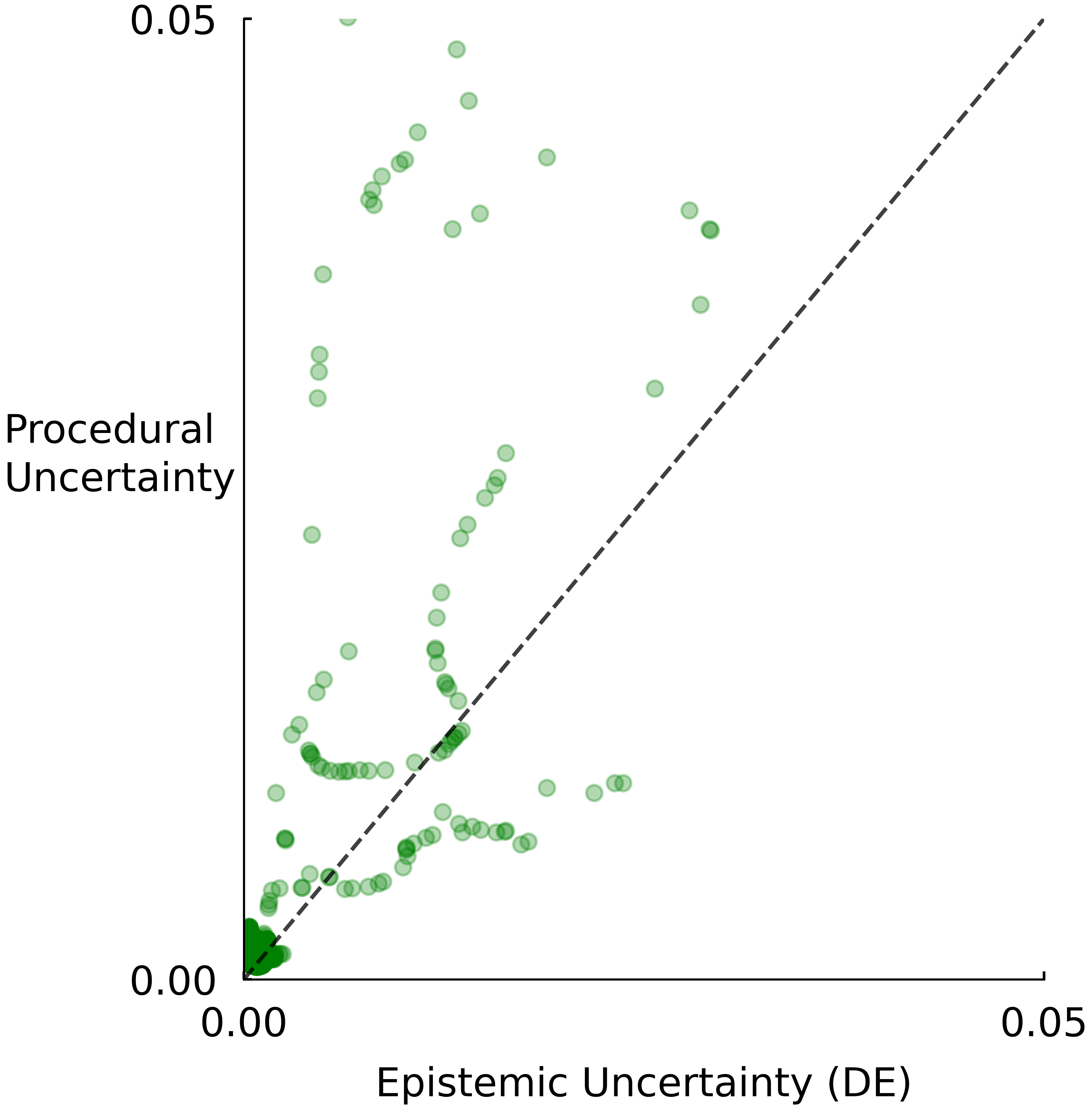}
        \caption{}
        \label{fig:epistemic_vs_procedural_sine_comparison_de}
    \end{subfigure}
    \caption{Epistemic uncertainty estimates for the synthetic dataset. (a)Decomposition of the epistemic uncertainty estimated with the reference distribution into procedural and data uncertainties. There exists a strong correlation between both, which is expected, and procedural uncertainty outweighs across almost all the data points. (b)Comparison between the epistemic uncertainty estimated with the reference distribution (both data and procedural components) and the epistemic uncertainty estimated with the Deep Ensemble. Here, the epistemic part of the reference distribution is larger, showing that the Deep Ensemble underestimates this uncertainty. (c)When the data uncertainty component is removed from the estimate of the reference distribution, both estimates are much closer, suggesting that the Deep Ensemble fails to capture the data uncertainty component of the epistemic uncertainty.}
    \label{fig:procedural_and_data_sine_deep_ensemle}
\end{figure}

\begin{figure}[t]
    \centering
    \begin{subfigure}[b]{0.4\textwidth}
        \includegraphics[width=\textwidth]{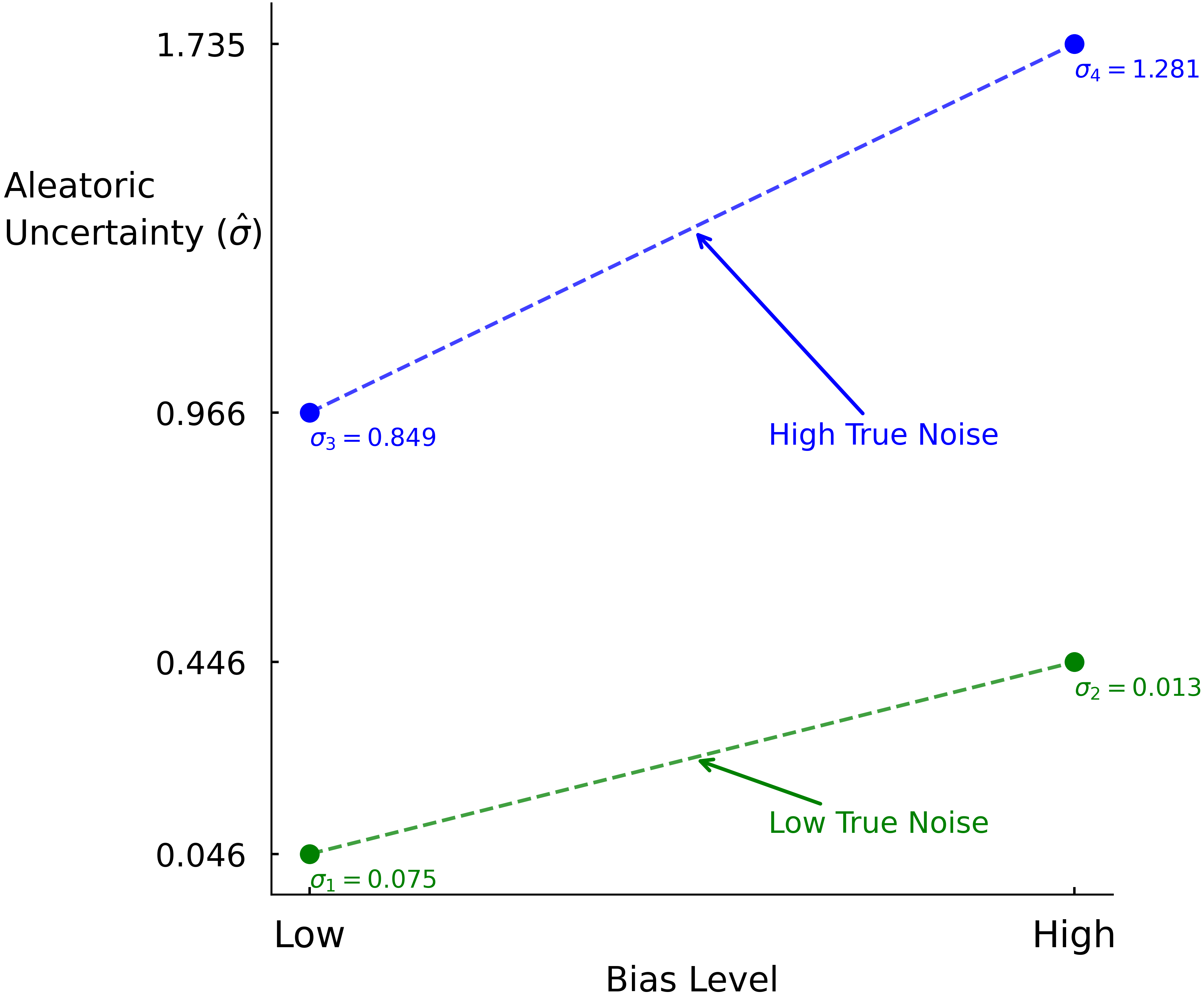}
        \caption{Synthetic data (Equation~\ref{eq: sine data problem})}
        \label{fig:aleatoric_vs_bias_sine}
    \end{subfigure}
    \begin{subfigure}[b]{0.4\textwidth}
        \includegraphics[width=\textwidth]{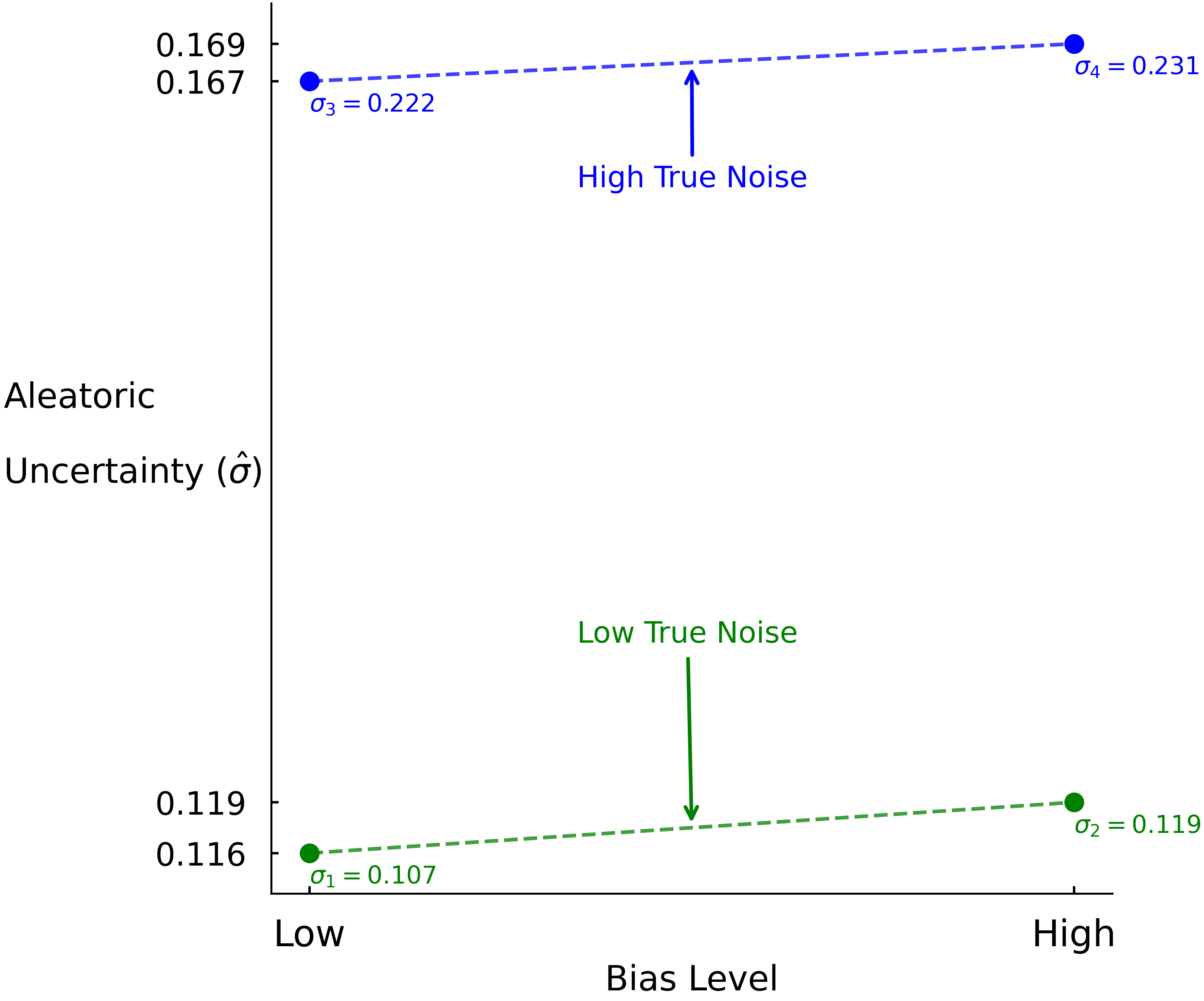}
        \caption{Taxi dataset.}
        \label{fig:aleatoric_vs_bias_taxi}
    \end{subfigure}
\label{fig:aleatoric_vs_bias}
\caption{Comparison of the estimated aleatoric uncertainty with the true (simulated) aleatoric noise across regions characterized by different levels of model bias and inherent data noise. The notation $\sigma_i$ for $i \in {1, 2, 3, 4}$ denotes the true aleatoric uncertainty for each of the four regions: 1. low bias and low true noise, 2. high bias and low true noise, 3. low bias and high true noise, and 4. high bias and high true noise. For the synthetic dataset, the division between high- and low-noise regions follows Eqn.~\ref{eq: sine data problem}: points with $x < 0.6$ are considered low-noise, while those with $x \geq 0.6$ are considered high-noise. Within each noise region, samples are ranked according to their model bias, and the lower half of points (by bias magnitude) defines the low-bias region, while the upper half defines the high-bias region. For the taxi dataset, the high- and low-noise regions are determined using the $90$-th percentile of the simulated true aleatoric uncertainty, as described in Section~\ref{app: real data experiments}. Within each noise region, data points are further divided into low- and high-bias subsets based on the 70th percentile of the estimated model bias. This construction allows for a direct comparison between the estimated and true aleatoric uncertainty across systematically defined regions, highlighting how model bias affects the reliability of aleatoric uncertainty estimates in both synthetic and real-world scenarios.}
\end{figure}

\begin{figure}[t]
    \centering
    \begin{subfigure}{0.23\textwidth}
        \includegraphics[width=\linewidth]{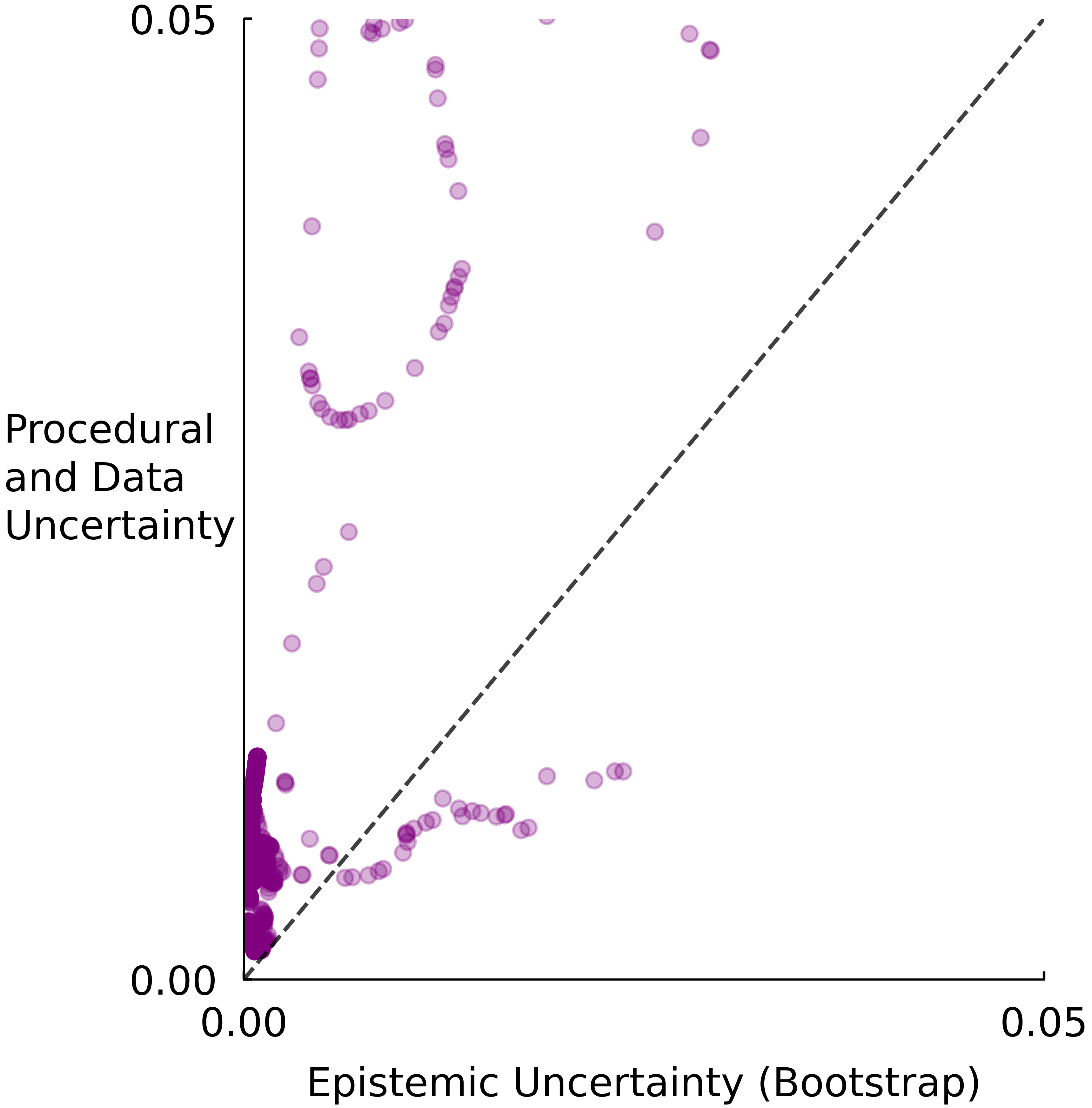}
        \caption{Bootstrap}
    \end{subfigure}
    \begin{subfigure}{0.23\textwidth}
        \includegraphics[width=\linewidth]{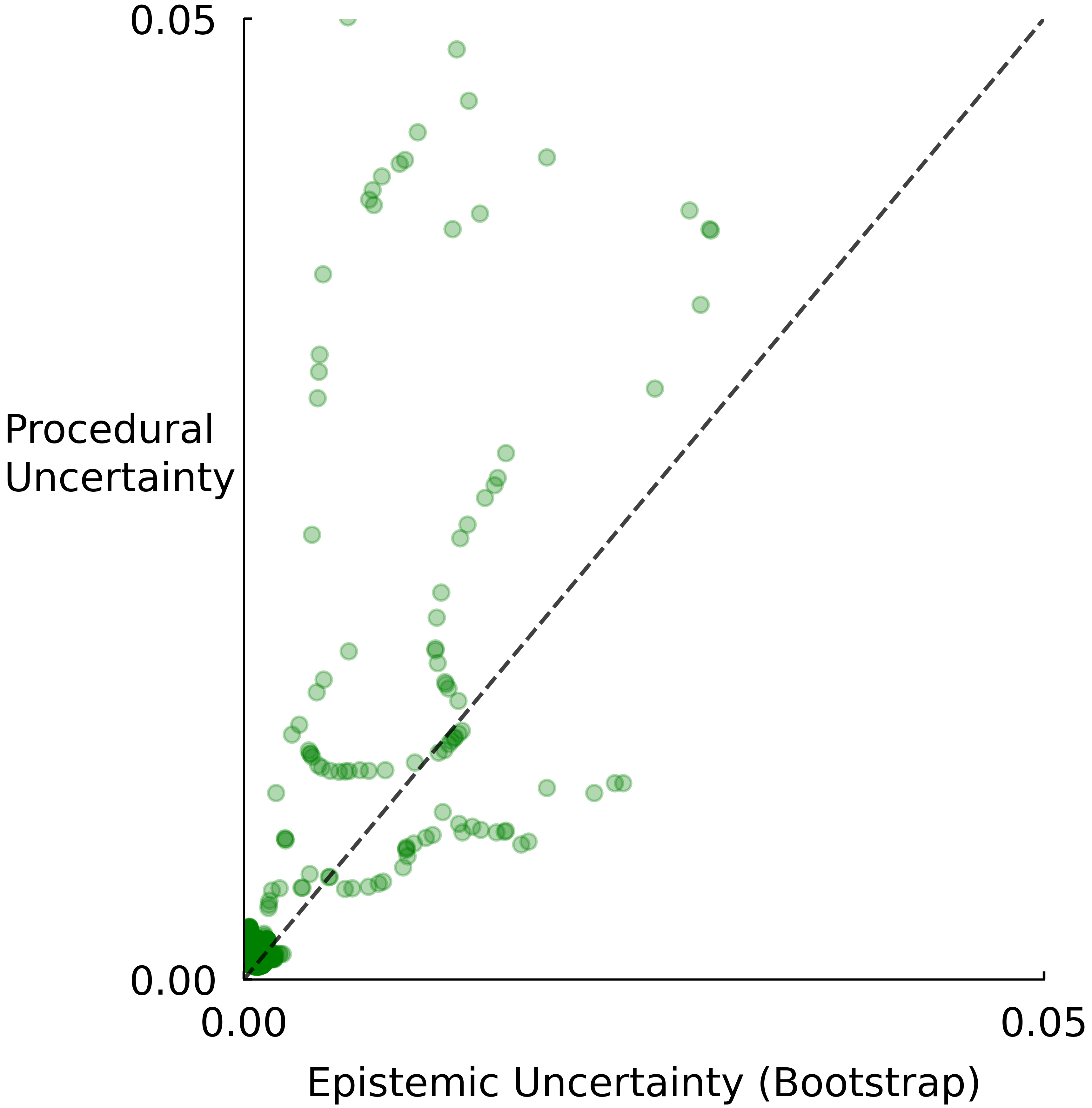}
        \caption{Bootstrap}
    \end{subfigure}
    \begin{subfigure}{0.23\textwidth}
        \includegraphics[width=\linewidth]{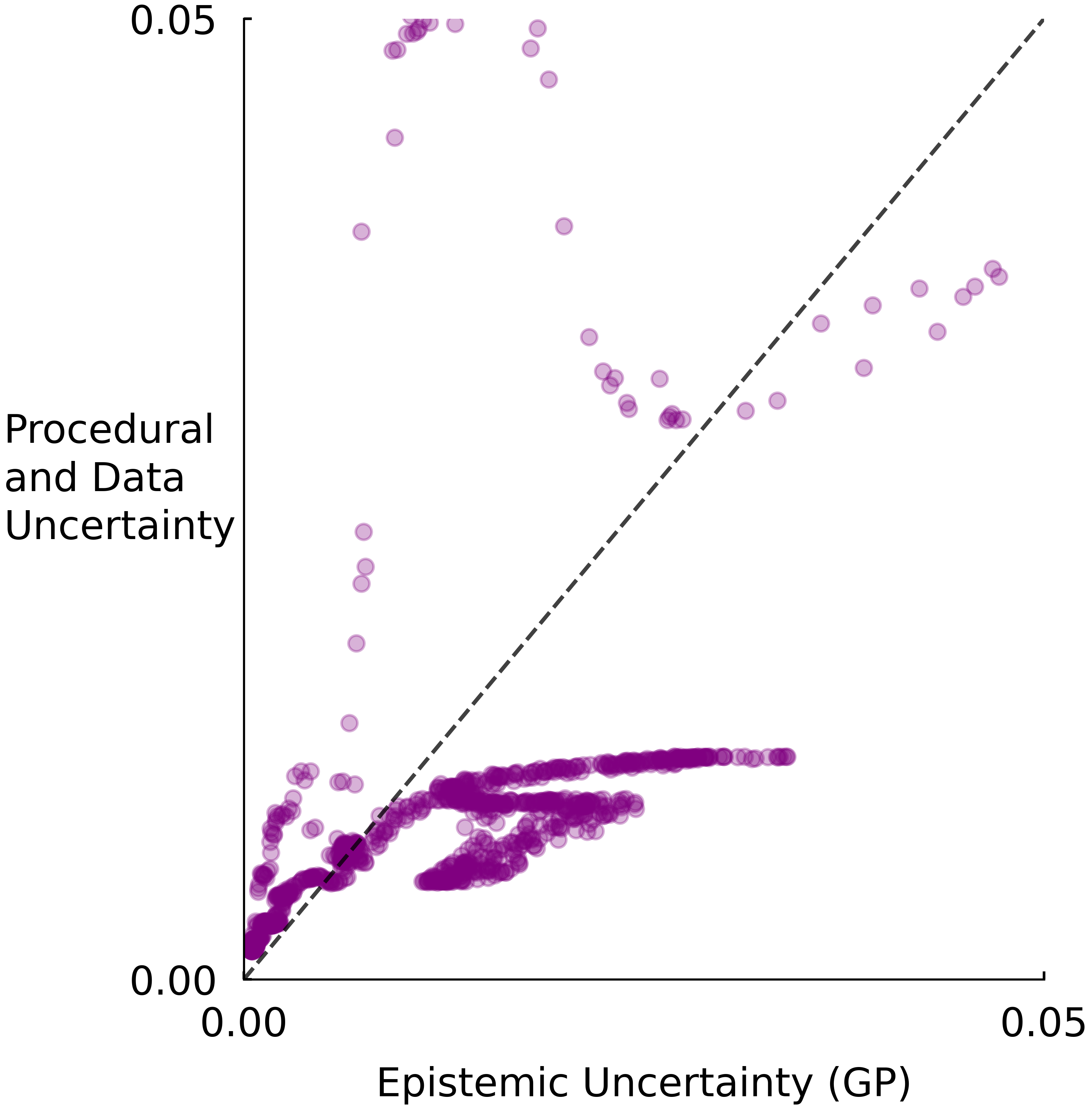}
        \caption{GP}
    \end{subfigure}
    \begin{subfigure}{0.23\textwidth}
        \includegraphics[width=\linewidth]{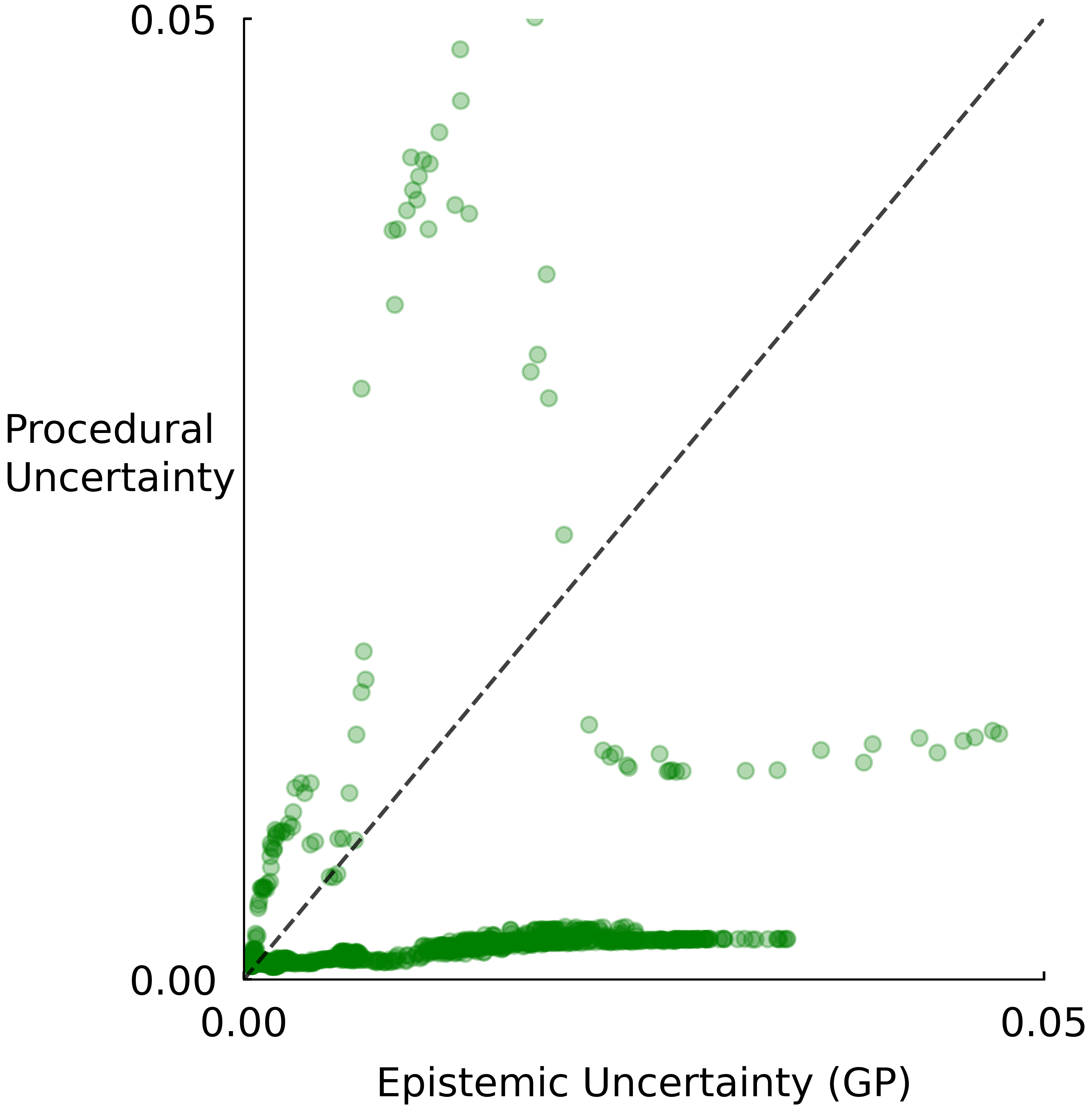}
        \caption{GP}
    \end{subfigure}

    \vspace{1em}

    \begin{subfigure}{0.23\textwidth}
        \includegraphics[width=\linewidth]{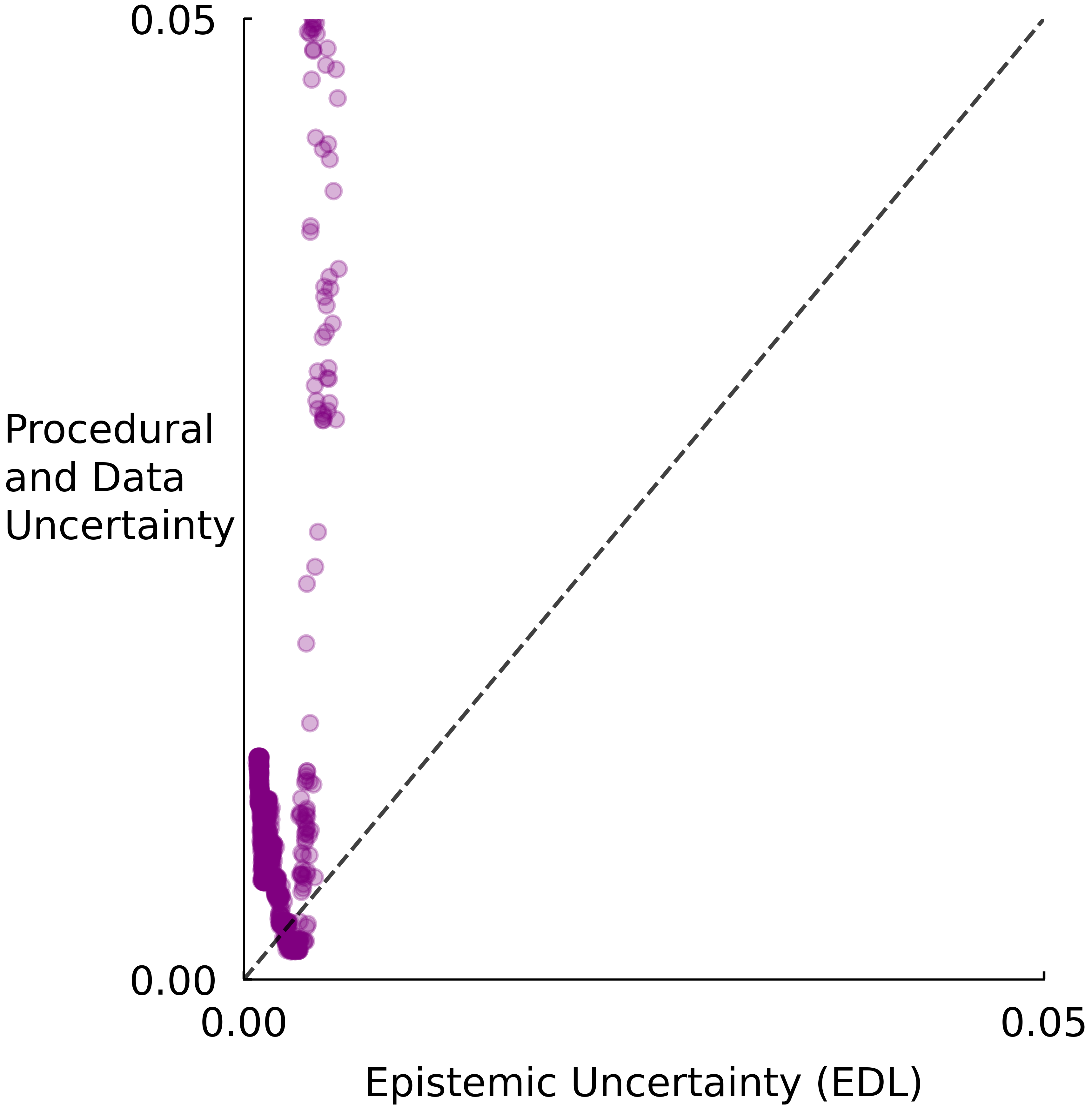}
        \caption{EDL}
    \end{subfigure}
    \begin{subfigure}{0.23\textwidth}
        \includegraphics[width=\linewidth]{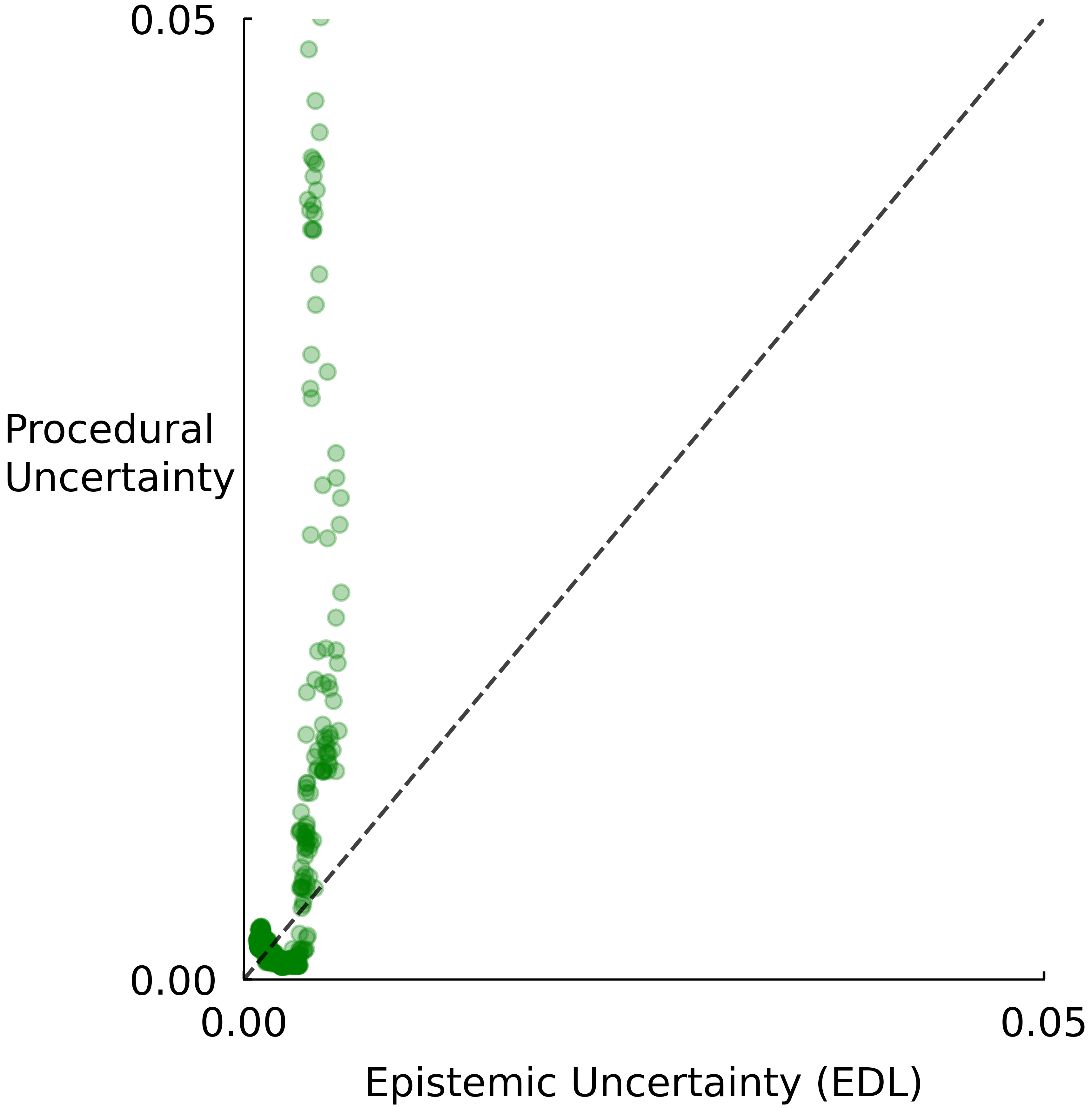}
        \caption{EDL}
    \end{subfigure}
    \begin{subfigure}{0.23\textwidth}
        \includegraphics[width=\linewidth]{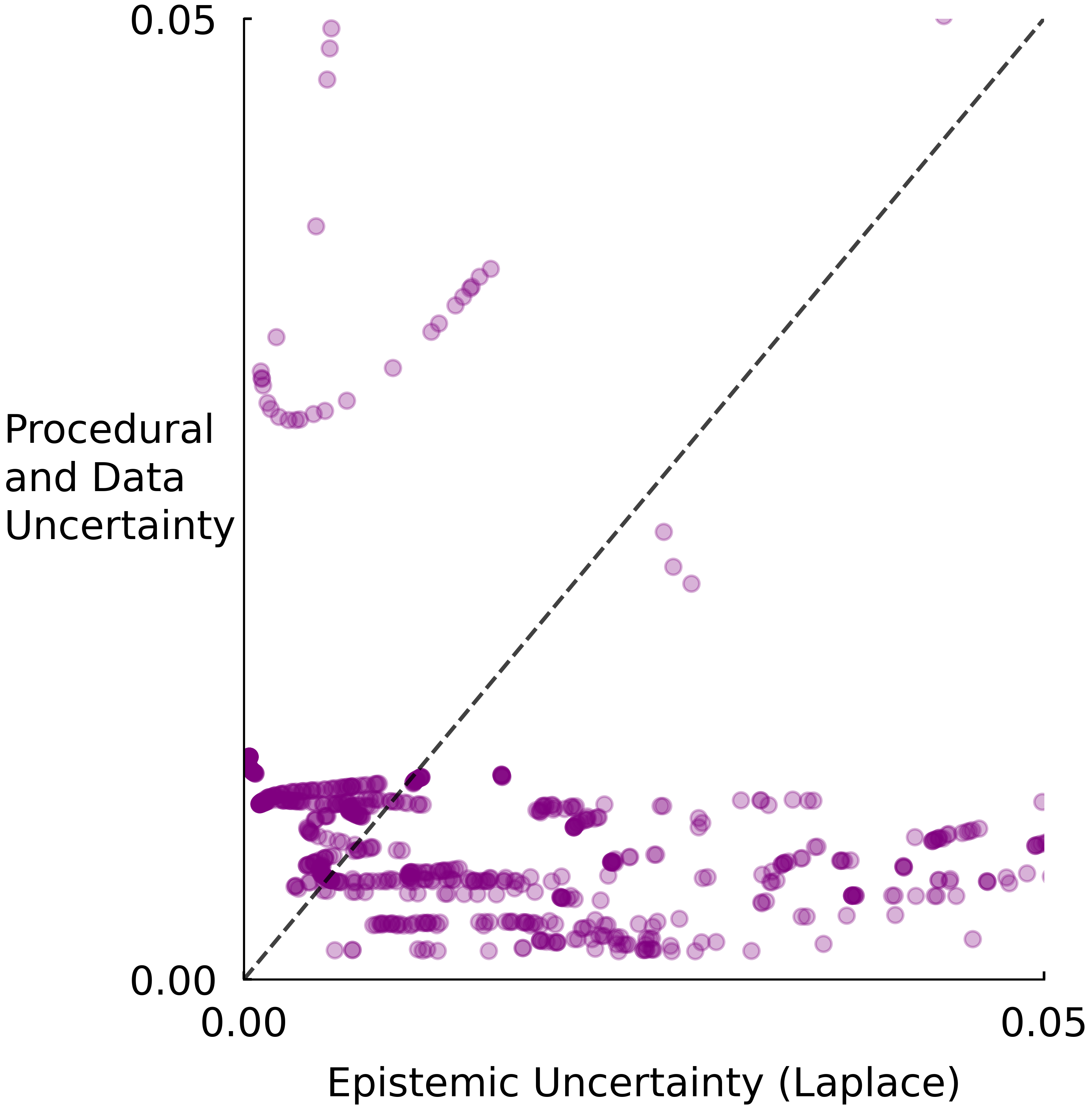}
        \caption{Laplace}
    \end{subfigure}
    \begin{subfigure}{0.23\textwidth}
        \includegraphics[width=\linewidth]{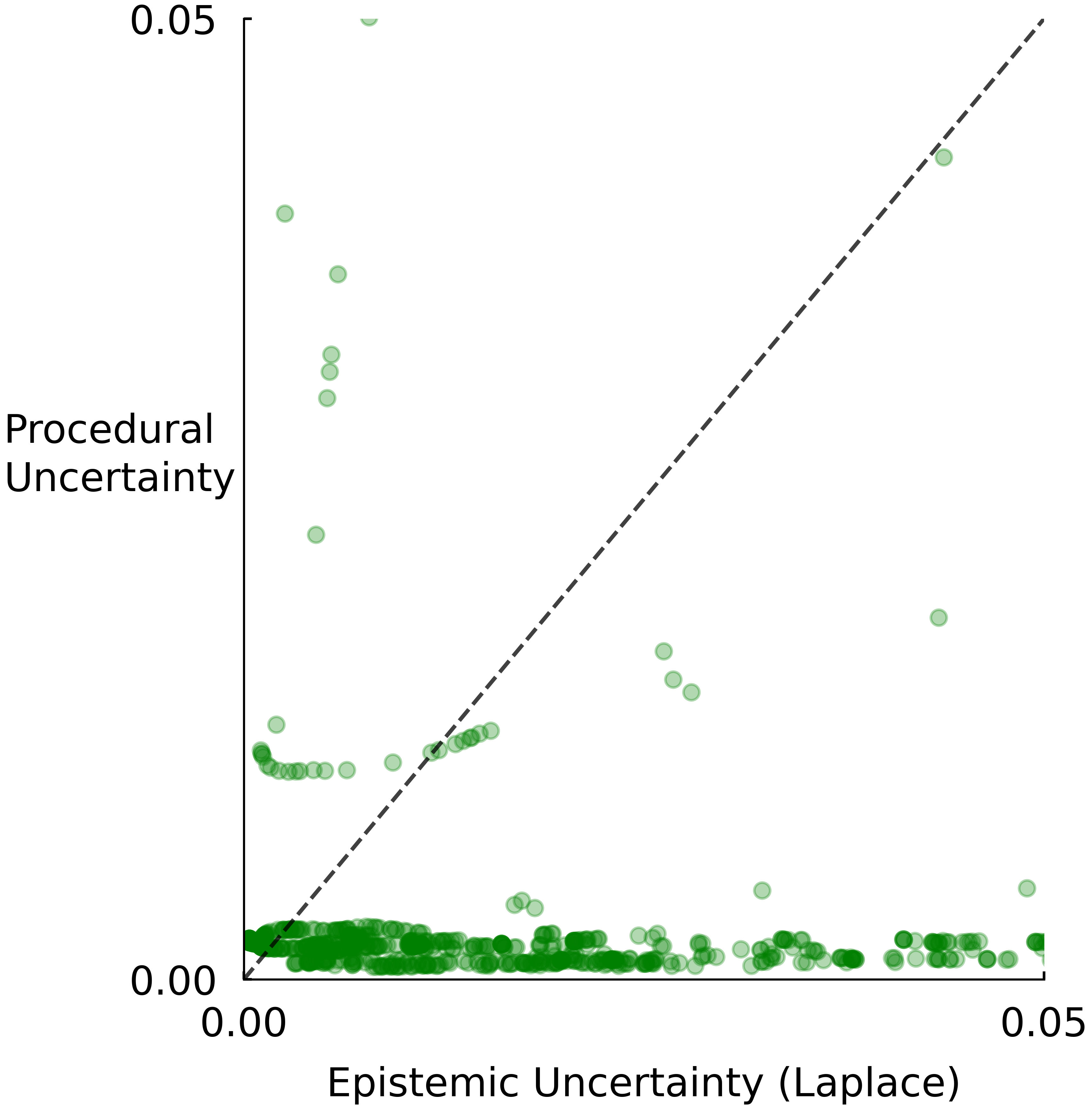}
        \caption{Laplace}
    \end{subfigure}

    \vspace{1em}

    \begin{subfigure}{0.23\textwidth}
        \includegraphics[width=\linewidth]{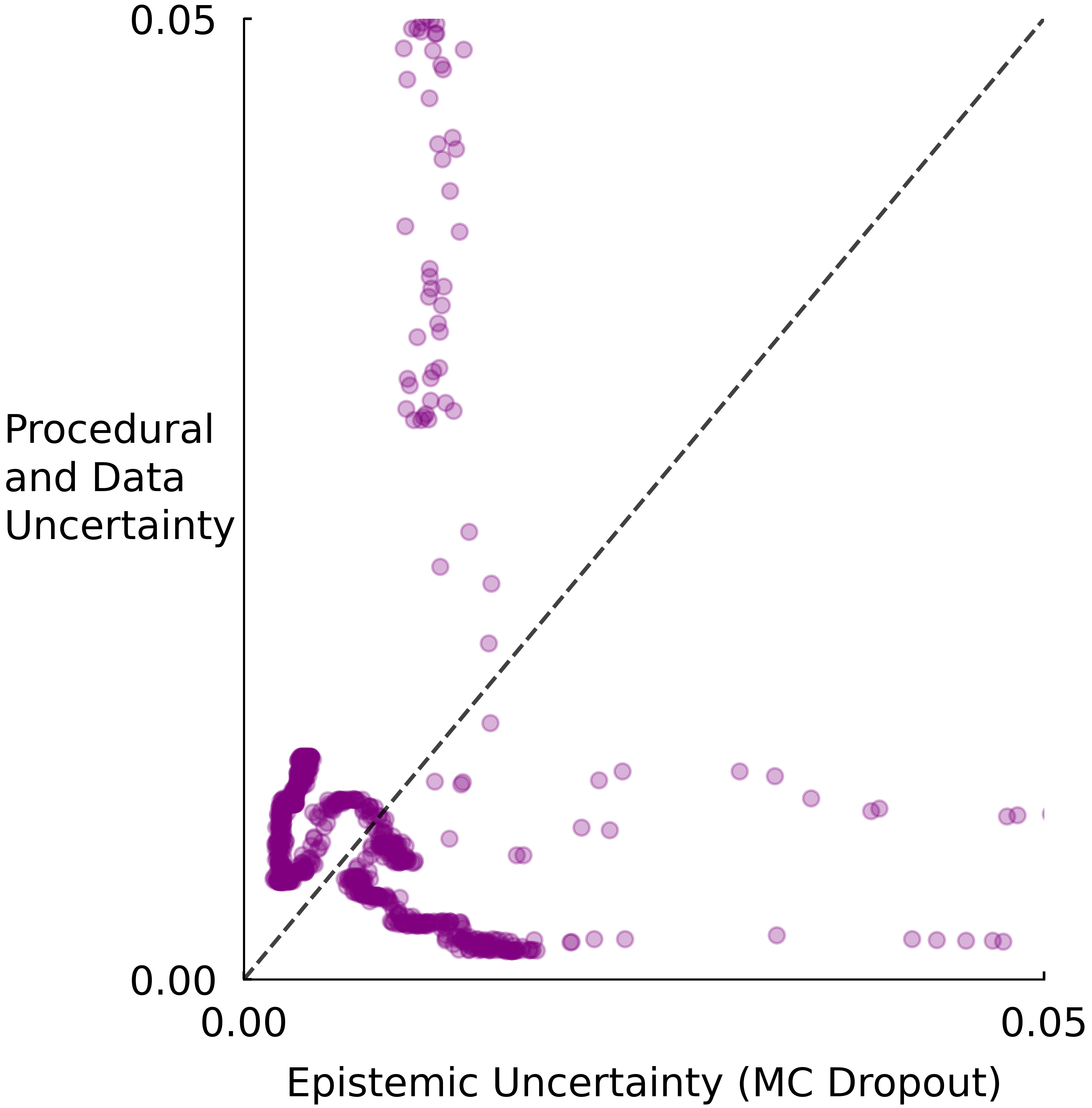}
        \caption{MC Dropout}
    \end{subfigure}
    \begin{subfigure}{0.23\textwidth}
        \includegraphics[width=\linewidth]{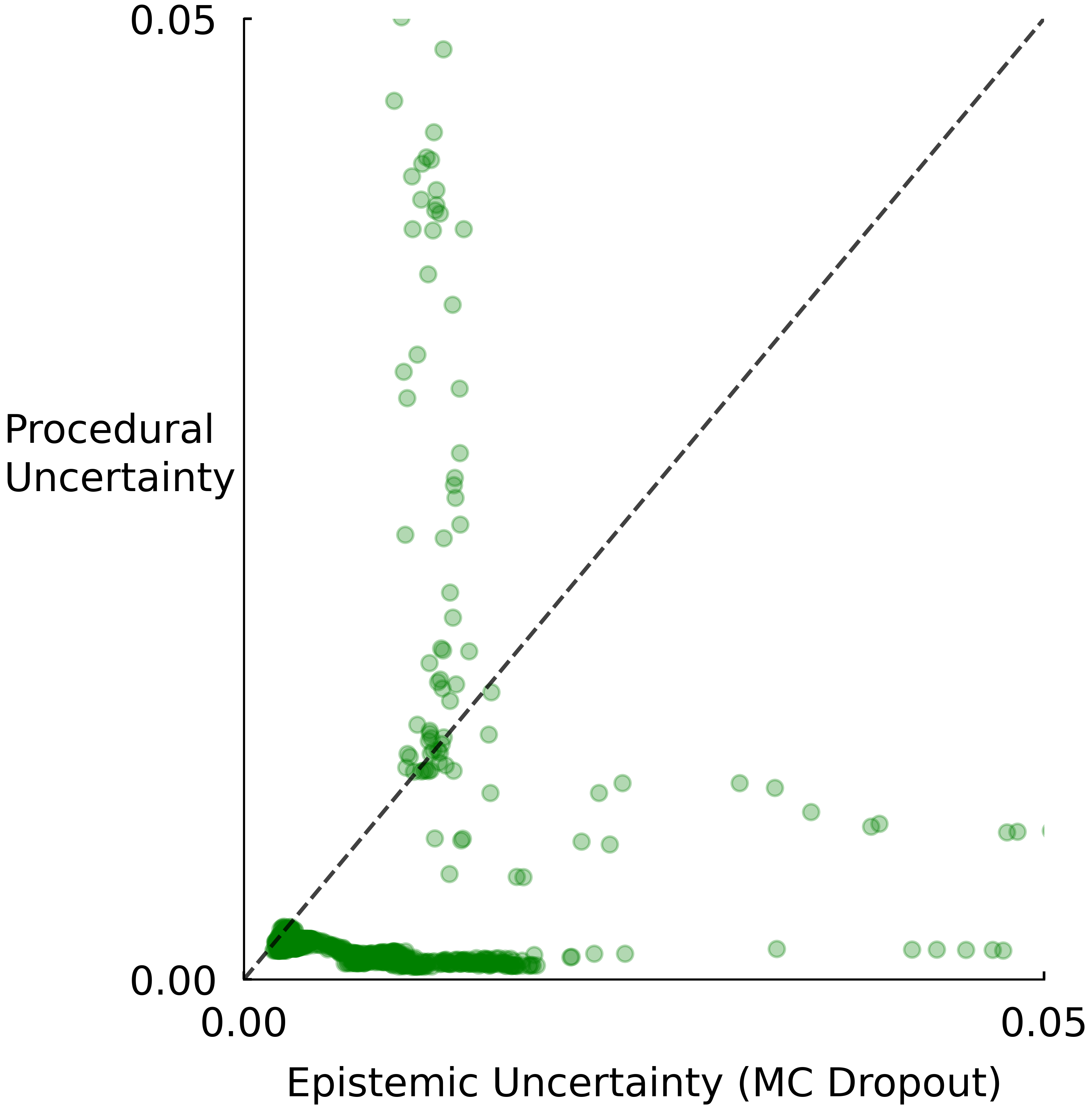}
        \caption{MC Dropout}
    \end{subfigure}
    \begin{subfigure}{0.23\textwidth}
        \includegraphics[width=\linewidth]{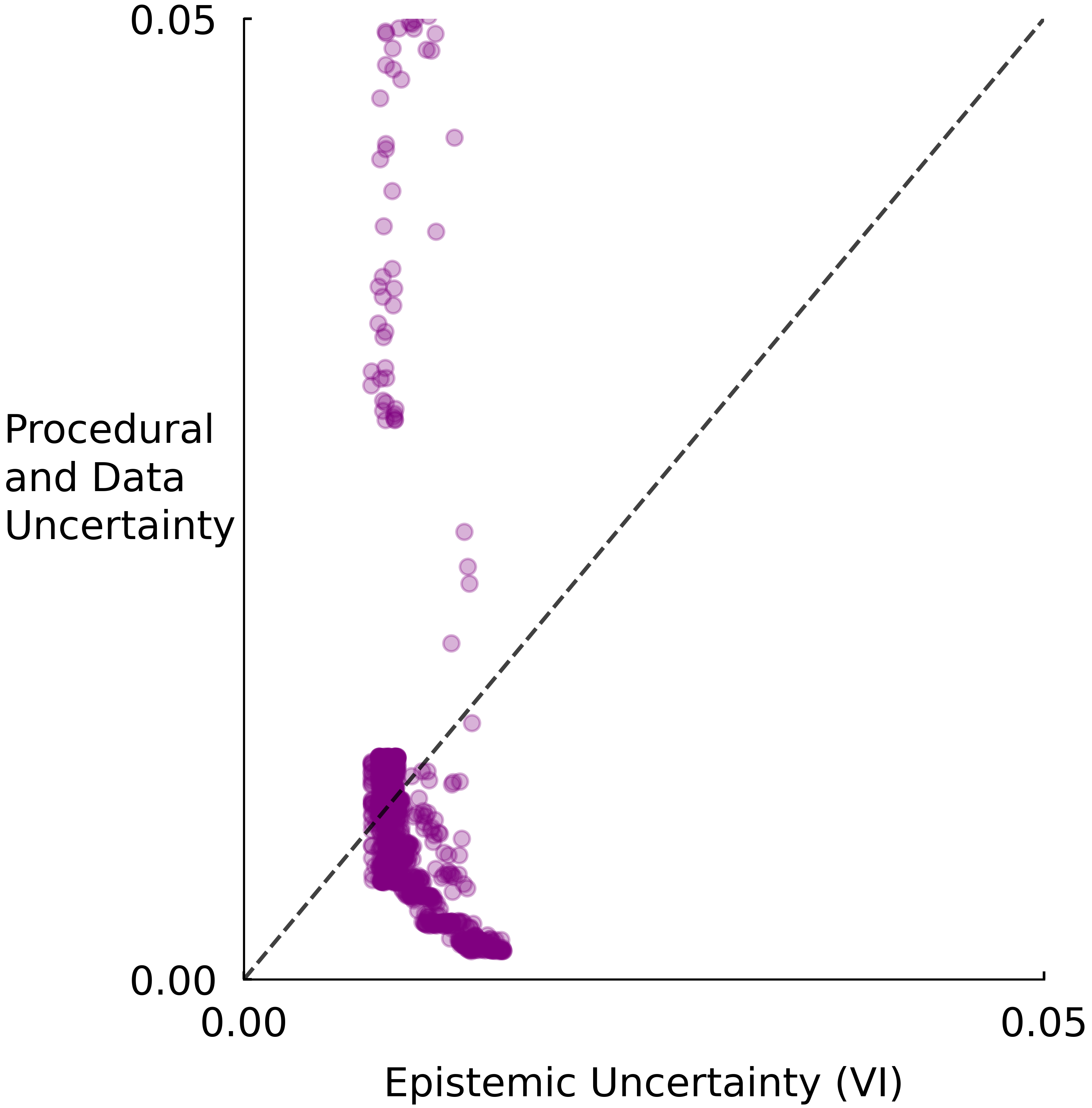}
        \caption{VI}
    \end{subfigure}
    \begin{subfigure}{0.23\textwidth}
        \includegraphics[width=\linewidth]{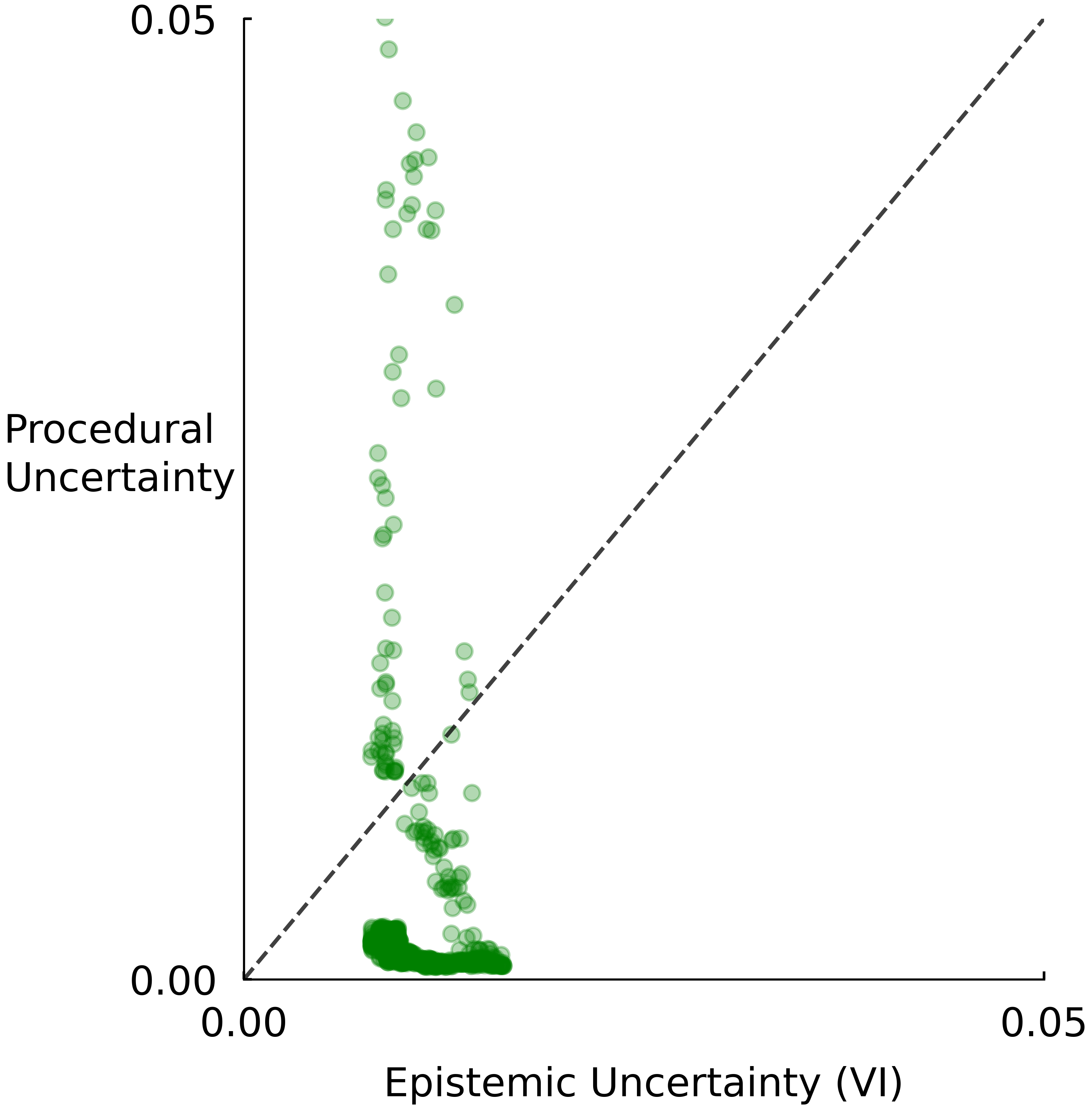}
        \caption{VI}
    \end{subfigure}

    \caption{Comparison of procedural and data uncertainties vs epistemic uncertainty across methods on the synthetic regression dataset (Eqn.~\ref{eq: sine data problem}).}
    \label{fig:sine_uncertainty_relations_grid}
\end{figure}

Figures \ref{fig:epistemic_vs_procedural_sine_comparison_de} and \ref{fig:sine_uncertainty_relations_grid} further illustrate the relationship between the epistemic uncertainty estimates and the procedural and data uncertainty components obtained from the reference distribution. When comparing these components, the relation shifts upward once data uncertainty is included alongside procedural uncertainty, indicating that most methods primarily capture the procedural component of epistemic uncertainty. This observation highlights the necessity of adopting a more fine-grained taxonomy to identify which parts of epistemic uncertainty are being represented, as well as the importance of the reference distribution in empirically validating these estimates. On the other hand, the aleatoric uncertainty estimates on the right‐hand side of the data support improve as the sample size increases, the behaviour on the left‐hand side remains problematic. Specifically, the model persistently overestimates aleatoric uncertainty. A plausible explanation is that the approximate symmetry of the observations around the learned conditional mean induces the model to misinterpret systematic deviation as inherent data noise. Consequently, this misestimation is not corrected with additional data; instead, the model becomes increasingly confident in its erroneous predictions and continues to interpret the discrepancy from the ground-truth conditional mean as aleatoric uncertainty. This behavior is illustrated in Figure \ref{fig:aleatoric_vs_bias_sine}, which compares the average estimated aleatoric uncertainty with the true aleatoric noise in regions of low and high model bias. The figure shows that in areas of high bias, the estimated aleatoric uncertainty diverges substantially from the true value, demonstrating that biased predictions systematically distort the disentanglement of aleatoric and epistemic uncertainty.

The results for the other methods can be seen in Figures \ref{fig:vi}-\ref{fig:edl}. We highlight two additional key observations: First, we observe that Bayesian approaches --specifically Laplace Approximation (LA) and Hamiltonian Monte Carlo (HMC)-- require a larger amount of data to produce accurate estimates of aleatoric uncertainty in the right-hand region of the input space. This behavior aligns with theoretical expectations of Bayesian inference: The increased data requirement reflects the greater difficulty these models encounter in learning the underlying data-generating function, which in turn leads to higher epistemic uncertainty--consistent with models that make inaccurate predictions. Second, the Deep Evidential Regression (DER) method consistently produces poor estimates of aleatoric uncertainty across all sample sizes and throughout the entire input domain. These inaccuracies are accompanied by overconfident predictions, as evidenced by the systematic underestimation of epistemic uncertainty. This flawed characterization of uncertainty is consistent with findings reported in recent literature \cite{juergensepistemic}.
We see that every method with a low epistemic uncertainty estimate overestimates the aleatoric uncertainty in the left-hand side of the figure due to their inability to acknowledge the bias as part of their epistemic uncertainty representations. Also, for more data-intensive methods, such as the Bayesian approaches, a more careful and realistic estimate of the epistemic uncertainty is obtained.

\subsubsection{Implementation details by method}
\label{sec: Experimental details by method}
All methods are evaluated for different training dataset sizes: $\{50, 100, 500\}$ with batch sizes
 $\{32, 32, 64\}$ respectively. The experiments are repeated for $5$ different seeds: $7, 42, 123, 999$ and $2024$.
  To conduct the experiments, a hyperparameter tuning over relevant parameters was performed.\\

\textbf{Deep Ensemble.} Trained with an MLP of $4$ hidden layers with $100$ neurons each, learning rate of 0.009.
 Training epochs: $50, 250, 500$. Every member of the deep ensemble uses the whole dataset.

\textbf{Bootstrap Ensemble.} Trained as the Deep Ensemble, but each ensemble member uses a boostrap sample
 with $60\%$ of the training data.

\textbf{Monte Carlo Dropout.} A single MLP of $4$ hidden layers with $100$ neurons each is trained.
 A dropout rate of $0.2$ is used at training and inference time. A learning rate of 0.002 is used.

\textbf{Variational Inference.} This model is trained based on \citet{blundell2015weight}.
 An MLP of $4$ hidden layers with $100$ neurons each converted to a Bayesian neural network is used.
  The model is trained with a learning rate of $0.005$, $200$ burn-in epochs, a beta value of $5$, $10$ Monte Carlo samples at training time, and $500$ at test. The model is evaluated for $250$ and $500$ training epochs.

\textbf{Laplace Approximation.} An MLP of $4$ hidden layers with $100$ neurons each is used to find
 the MAP estimate, together with a learning rate of $0.005$. For the Laplace step, a prior precision of $1$
  and a noise value of $2$ are used. The model is trained for $200, 400$, and $800$ epochs, and uses $1000$
   samples from the posterior distribution. When using a training sample size of $500$, a batch size of 128 is used.

\textbf{Hamiltonian Monte Carlo.} An MLP of $4$ hidden layers with $100$ neurons each is used as
 the underlying model. The sampling step consists of $200$ samples with a sampling step of $0.00015$.
  The first $50$ samples are always burnt. A $\tau$ value of $1$ is used as the indicator for a \textit{bendy}
   function. At inference time, we sample $1000$ times from the posterior distribution and drop the first $50$.
    The model is trained for $1000, 2000$, and $3000$ epochs. When using a sample size of 500, a batch size of 128 is used.

\textbf{Deep Evidential Regression.} Leaning on the setup of \citet{amini2020deep}, an MLP of $4$ hidden layers
 with $100$ neurons each is used as the underlying model. The model is trained with a learning rate of $0.0003$
  over $5000$ epochs, and a regularization strength of $\lambda=0.01$.

\textbf{Heteroscedastic Gaussian Process.} Leaning on the variational-sparse recipe of \citet{pmlr-v5-titsias09a},
 we use the GPytorch ApproximateGP module to train a two-output GP, trained with an RBF kernel. 
 The number of inducing points (see \cite{hastie2009elements} for an explanation) is set to $256$,
  the learning rate to 0.005, epochs to $2000$.

\subsection{Real data experiments}
\label{app: real data experiments}
In this section we extend our analysis to the New York taxi trip duration dataset published in this \href{https://www.kaggle.com/competitions/nyc-taxi-trip-duration/data}{Kaggle} competition. The dataset consists of taxi trip records from New York City, where the target variable corresponds to the trip duration and the feature space comprises 14 input variables. The training set contains $1,206,737$ observations, and the test set includes $230,406$ observations. This dataset is particularly suitable for evaluating uncertainty estimation methods for several reasons. First, its large sample size allows us to analyze how epistemic uncertainty behaves when the data-driven component becomes relatively small compared to the model and procedural components, allowing us to estimate data and procedural uncertainty via the reference distribution as we explained in Appendix \ref{app: reference distribution details}. Second, since the task is formulated as a heteroscedastic regression problem, where trip durations exhibit high variability depending on factors such as traffic conditions or route choice, it provides a natural setting for disentangling aleatoric from epistemic uncertainty.
Finally, the dataset’s scale and heterogeneity make it a realistic benchmark for assessing whether uncertainty quantification methods can remain well-calibrated and informative in complex, high-dimensional environments. In the same sense, a simulated estimate of ground truth aleatoric uncertainty can be obtained as we explained in Appendix \ref{app: ground truth aleatoric for taxi data}. As in the synthetic data experiments, we assume that the target variable follows a normal distribution, and the model outputs both the mean and variance of this distribution.

\subsubsection{Ground truth aleatoric uncertainty approximation}
\label{app: ground truth aleatoric for taxi data}

\textbf{Aleatoric uncertainty}. To approximate the ground-truth aleatoric uncertainty, we train an ensemble of $n_{\gamma}=5$ multilayer perceptrons (MLPs) on the full training dataset of $1,206,737$ observations. Since aleatoric uncertainty represents the irreducible noise inherent to the data-generating process, training on the complete dataset minimizes the influence of data-driven and procedural components, yielding the most reliable estimate of this term attainable in practice. We take the mean predicted variance across the ensemble members as the reference estimate of aleatoric uncertainty.
To assess the effect of limited data and model bias on uncertainty disentanglement, we then compare this reference estimate with those obtained when each model is trained using only $1\%$ of the available data. This comparison highlights how insufficient data coverage and higher epistemic uncertainty can distort the estimated aleatoric component, revealing the extent to which current methods conflate epistemic and aleatoric sources of uncertainty.

\subsubsection{Experimental illustrations}

\begin{figure*}[t]
    \centering
    \begin{subfigure}[b]{0.32\textwidth}
        \centering
        \includegraphics[width=\textwidth]{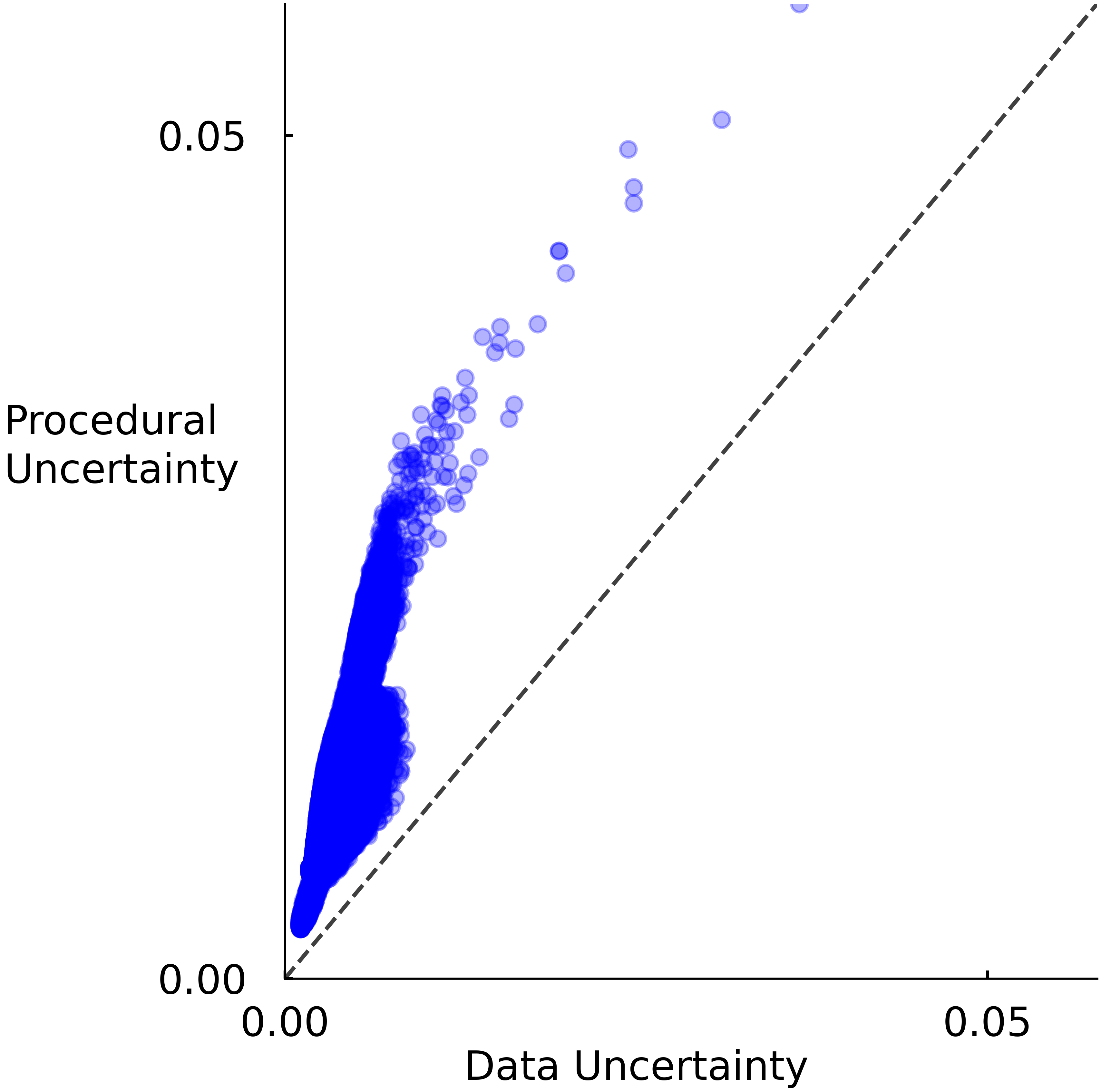}
        \caption{}
        \label{fig:procedural_vs_data_taxi}
    \end{subfigure}
    \begin{subfigure}[b]{0.32\textwidth}
        \centering
        \includegraphics[width=\textwidth]{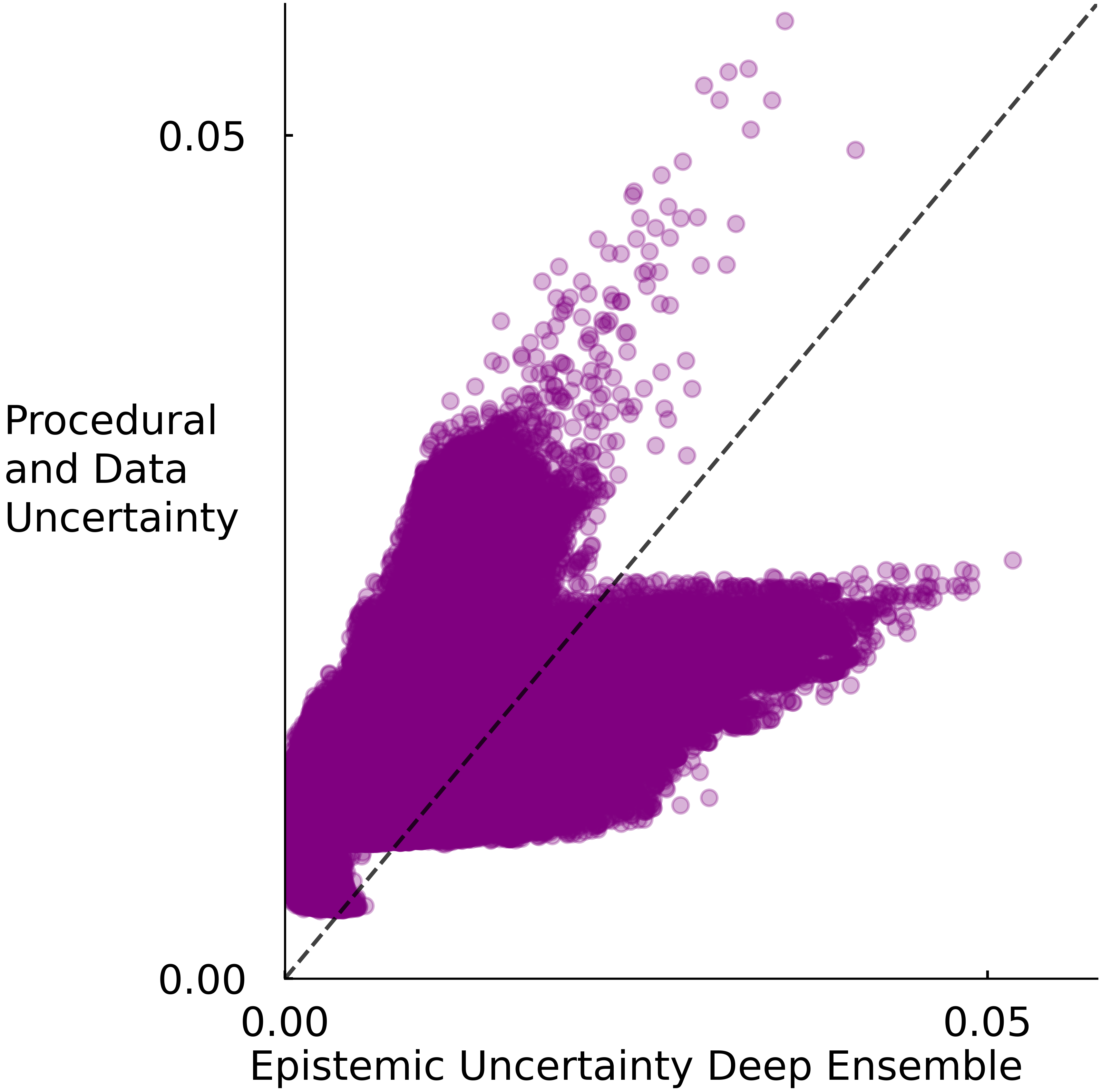}
        \caption{}
        \label{fig:epistemic_taxi_comparison}
    \end{subfigure}
    \begin{subfigure}[b]{0.32\textwidth}
        \centering
        \includegraphics[width=\textwidth]{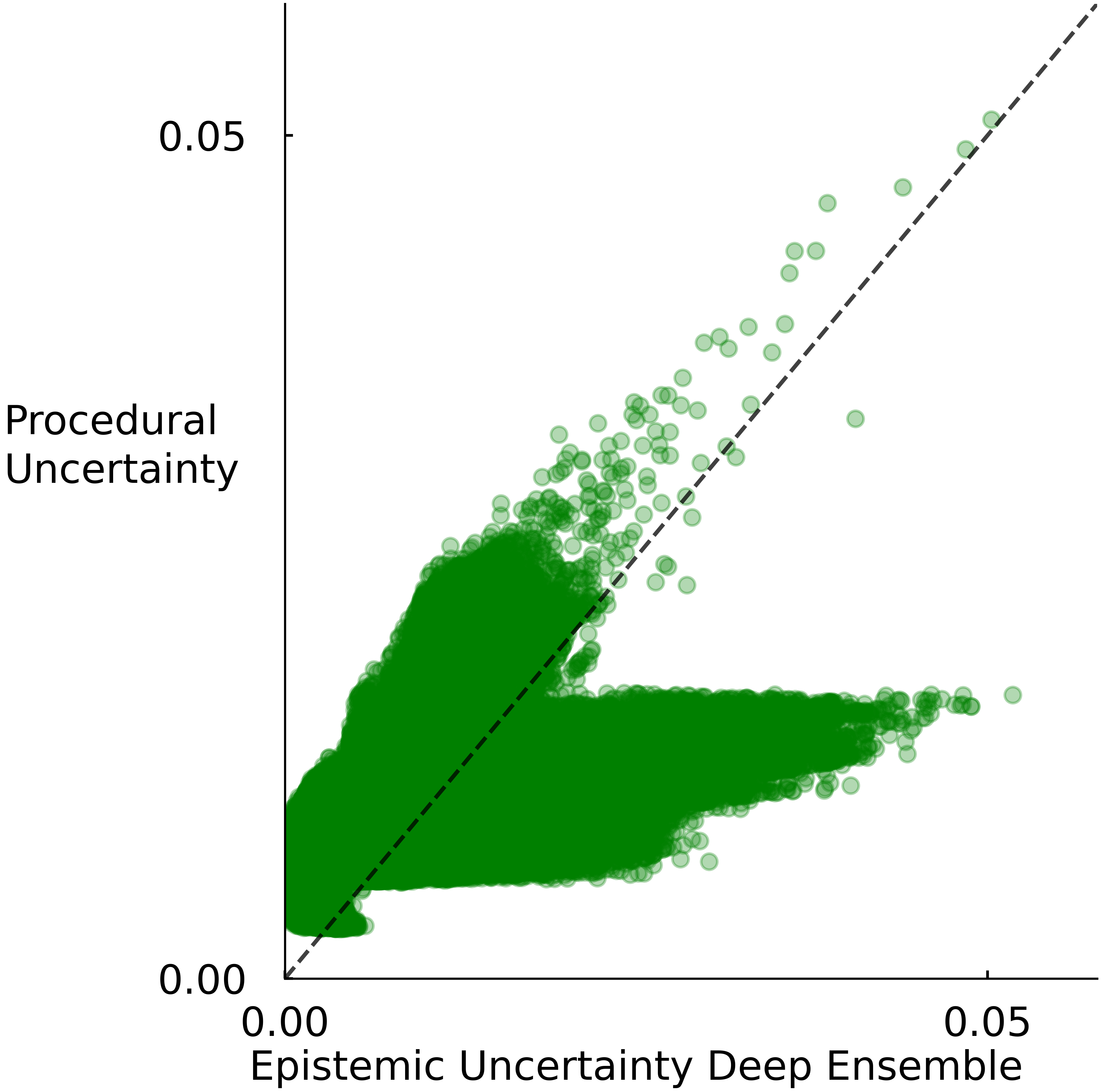}
        \caption{}
        \label{fig:epistemic_vs_procedural_taxi}
    \end{subfigure}
    \caption{Epistemic uncertainty evaluation for the Deep Ensemble on the taxi dataset. 
    (a) Decomposition of the variance of the reference distribution into data and procedural uncertainty. 
    (b) Deep Ensemble epistemic estimates vs.\ the total epistemic 
    uncertainty given by the reference distribution. $30\%$ of points fall below the diagonal. 
    (c) Deep Ensemble epistemic estimates vs.\ the procedural component of the reference distribution only.}
    \label{fig:procedural_data_comparison}
\end{figure*}

Figure~\ref{fig:epistemic_vs_procedural_taxi} shows the resulting decomposition of total epistemic variance into procedural and data components. We observe that procedural uncertainty dominates across most of the input domain. This outcome reflects the high capacity and non-convex optimization landscape of neural networks: even with abundant data, different random initializations can converge to distinct local minima, leading to substantial procedural variation. By contrast, data uncertainty is smaller, indicating that resampling subsets from such a large dataset induces only limited diversity in the fitted models. It is important to note that the relative magnitudes of these two components depend on the experimental configuration. Increasing the subset size $n_{d}$ would typically reduce data uncertainty, while enlarging $n_{\gamma}$ would shrink the procedural uncertainty. As discussed in Example \ref{ex: deep ensembles} and in the synthetic data experiments, Deep Ensembles predominantly capture the procedural component of epistemic uncertainty. This limitation arises because ensemble diversity originates solely from random initialization and stochastic training dynamics, while all models are trained on the same dataset. Consequently, the variance across ensemble members reflects the sensitivity of the optimization process rather than uncertainty about the data distribution itself.
This behavior is observed in Figure \ref{fig:epistemic_taxi_comparison}, where the total variance of the reference distribution, represented on the  $y$-axis as the sum of data and procedural uncertainty, is systematically higher than the epistemic uncertainty predicted by the Deep Ensemble. Quantitatively, the epistemic uncertainty estimate underestimates the reference distribution variance in approximately $80\%$ of the predictive samples. This systematic underestimation indicates that the ensemble fails to account for the data-driven component of epistemic uncertainty, which captures how model predictions would vary under different training datasets drawn from the same data-generating process.
To further examine this relationship, Figure \ref{fig:procedural_vs_data_taxi} decomposes the reference distribution into its procedural and data uncertainty components. When comparing only the procedural component of the reference distribution with the epistemic uncertainty predicted by the Deep Ensemble, the discrepancy disappears. This alignment confirms that Deep Ensembles primarily quantify procedural uncertainty, that is, the uncertainty arising from stochasticity in optimization and model initialization, while largely neglecting the variability induced by finite data sampling.
In practical terms, this result reinforces the theoretical argument presented in Section \ref{sec: data and procedural uncertainty}: the epistemic uncertainty estimated by Deep Ensembles does not reflect the full reducible uncertainty in the model but only its procedural subcomponent. Consequently, downstream tasks that rely on a correct epistemic uncertainty estimation, such as active learning, model selection, or out-of-distribution detection, may be misled by an incomplete uncertainty representation. In the same sense, Figure \ref{fig:aleatoric_vs_bias_taxi} shows that in the region where the model is more incorrect, the difference between the estimated aleatoric uncertainty and the simulated ground truth aleatoric uncertainty is larger than in the region where the model is more accurate, showing how a high bias can distort the disentanglement of aleatoric and epistemic uncertainty. The findings from this real-world regression task, therefore, underscore the importance of distinguishing between procedural and data uncertainty when evaluating or designing second-order uncertainty quantification methods.

\begin{table}[h]
\centering
\small    
\setlength{\tabcolsep}{4pt}

\begin{tabular}{l r r r r r}
\textbf{Model}
& $\hat{\sigma}^2$
& $\hat{\sigma}^2(x)-\sigma_*^2(x)$
& $\mathrm{Var}\!\big[\hat{y}|\vec{x}\big]$
& \shortstack[r]{$\mathrm{Var}_{\mathcal{D}_N,\gamma}\!\big(\hat{y}| \vec{x}\big)-$\\$\mathrm{Var}\!\big[\hat{y}| \vec{x}\big]$}
& \shortstack[r]{$\mathbb{E}_{\mathcal{D}_N}\!\big[\mathrm{Var}_\gamma\!\big(\hat{y}| \vec{x}\big)|\mathcal{D}_N\big]-$\\$\mathrm{Var}\!\big[\hat{y}| \vec{x}\big]$} \\
\midrule
DE & 0.1222 & 0.0004 & 0.0083 & 0.0046 & 0.0013 \\
Bootstrap & 0.1127 & 0.0091 & 0.0082 & 0.0047 & 0.0014 \\
MC Dropout & 0.1284 & 0.0066 & 0.0106 & 0.0023 & 0.001 \\
VI & 1.8859 & 1.7641 & 0.0055 & 0.0074 & 0.0041 \\
Laplace & 0.1754 & 0.0536 & 0.0137 & 0.0008 & 0.0041 \\
HMC & 0.2319 & 0.1101 & 0.1148 & 0.1019 & 0.1052 \\
GP & 1.0334 & 0.9116 & 0.0636 & 0.0507 & 0.054 \\
\bottomrule
\end{tabular}
\caption{Taxi dataset uncertainties. $\text{Var}\big[\hat{y}|\vec{x}\big]$ represents the epistemic uncertainty estimated by each model. $\text{Var}_{\mathcal{D}_N, \gamma}(\hat{y}|\vec{x})$ corresponds to the epistemic uncertainty obtained with the reference distribution, while $\mathbb{E}_{\mathcal{D}_N}\big[\text{Var}_{\gamma}(\hat{y}|\vec{x}) | 
    \mathcal{D}_N\big]$ is the procedural uncertainty obtained by the reference distribution. The values are the average over all test data points.}
\label{tab:uncertainties_biases}
\end{table}

\subsubsection{Data preprocessing}
The raw dataset used for this task can be found in this \href{https://www.kaggle.com/competitions/nyc-taxi-trip-duration/data}{Kaggle} competition. The goal is to predict the duration of a taxi trip in New York. The dataset contains $1,458,644$ observations from 2016. The code to preprocess the data is in the GitHub repository of this paper. The target variable is transformed according to $y'=\log(y+1)$.

\textbf{Data splitting}. We consider the data before June as the training dataset and the data after June as the test dataset. In this way, we get $1,206,737$ training points and $230,406$ test points. However, only $1\%$ of the dataset is used for training the models.

\subsubsection{Implementation details by method}
\textbf{Deep Ensemble.} Trained with an MLP of $10$ hidden layers with $300$ neurons each, learning rate of $0.0009$. Training epochs: $50$. Every member of the deep ensemble uses the whole dataset (containing $1\%$ of the total training dataset).

\textbf{Bootstrap Ensemble.} Trained as the Deep Ensemble, but each ensemble member uses a boostrap sample with $60\%$ of the training data (out of the $1\%$).

\textbf{Monte Carlo Dropout.} A single MLP of $10$ hidden layers with $300$ neurons each is trained. A dropout rate of $0.2$ is used at training and inference time. A learning rate of $0.0009$ is used.

\textbf{Variational Inference.} This model is trained based on \citet{blundell2015weight}. An MLP of $10$ hidden layers with $300$ neurons each converted to a Bayesian neural network is used. The model is trained with a learning rate of $0.0009$, $20$ burn-in epochs, a beta value of $10$, $10$ Monte Carlo samples at training time, and $500$ at test. The model is evaluated for $50$ training epochs.

\textbf{Laplace Approximation.} An MLP of $4$ hidden layers with $100$ neurons each is used to find the MAP estimate, together with a learning rate of $0.0009$. For the Laplace step, a prior precision of $1$ and a noise value of $0.1$ are used. The model is trained for $50$ epochs, and uses $100$ samples from the posterior distribution.

\textbf{Hamiltonian Monte Carlo.} An MLP of $10$ hidden layers with $300$ neurons each is used as the underlying model. The sampling step consists of $70$ samples with a sampling step of $0.00015$. The first $10$ samples are always burnt. A $\tau$ value of $1$ is used. At inference time, we sample $100$ times from the posterior distribution and drop the first $20$. The model is trained for $50$ epochs.

\textbf{Heteroscedastic Gaussian Process.} The implementation is the same as in the synthetic data case. The number of inducing points (see \cite{hastie2009elements} for an explanation) is set to $256$, the learning rate to $0.0009$, epochs to $50$.

\subsection{Further figures}


\begin{figure}[htbp]
    \centering
    \begin{subfigure}{0.23\textwidth}
        \includegraphics[width=\linewidth]{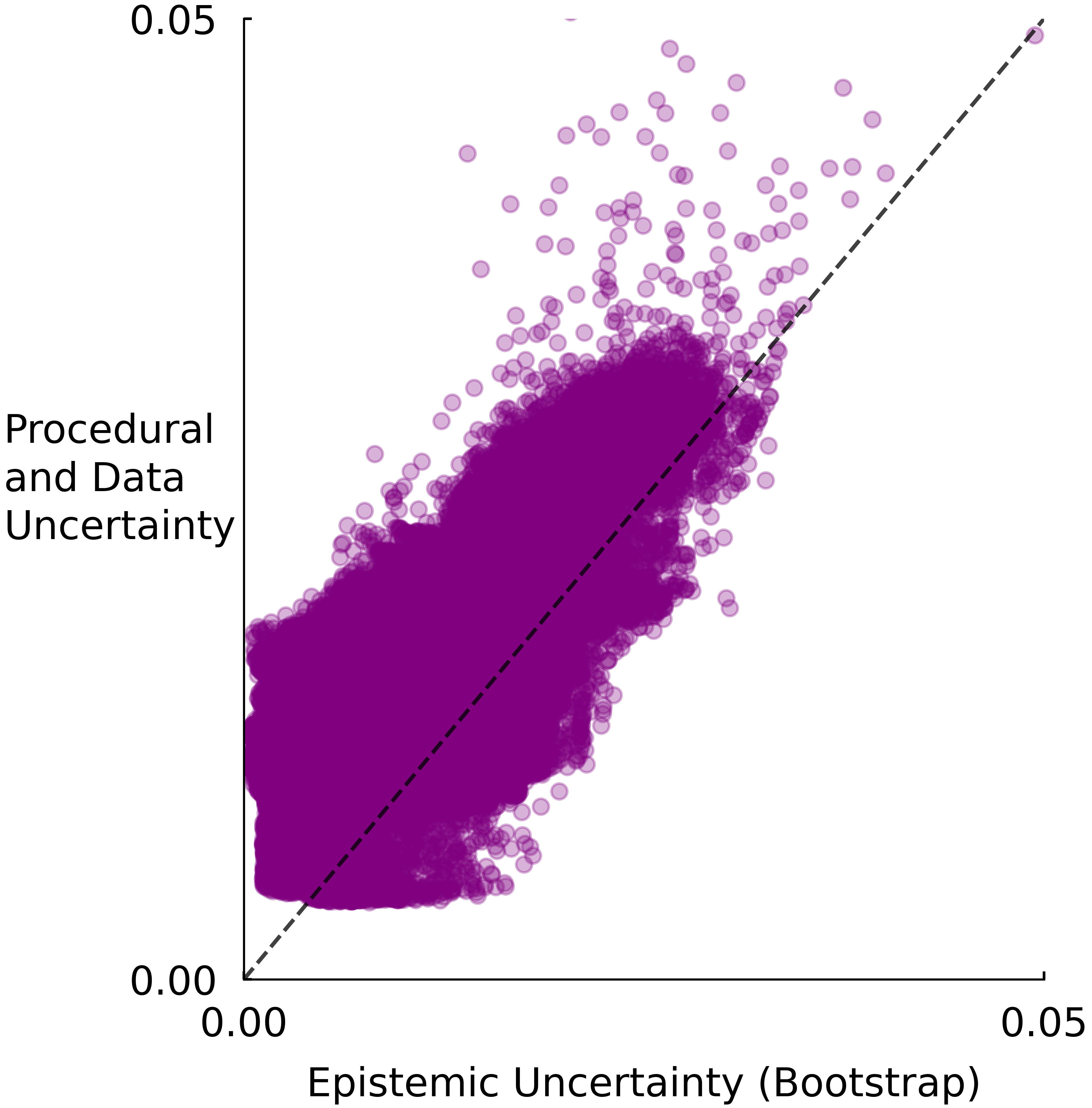}
        \caption{Bootstrap}
    \end{subfigure}
    \begin{subfigure}{0.23\textwidth}
        \includegraphics[width=\linewidth]{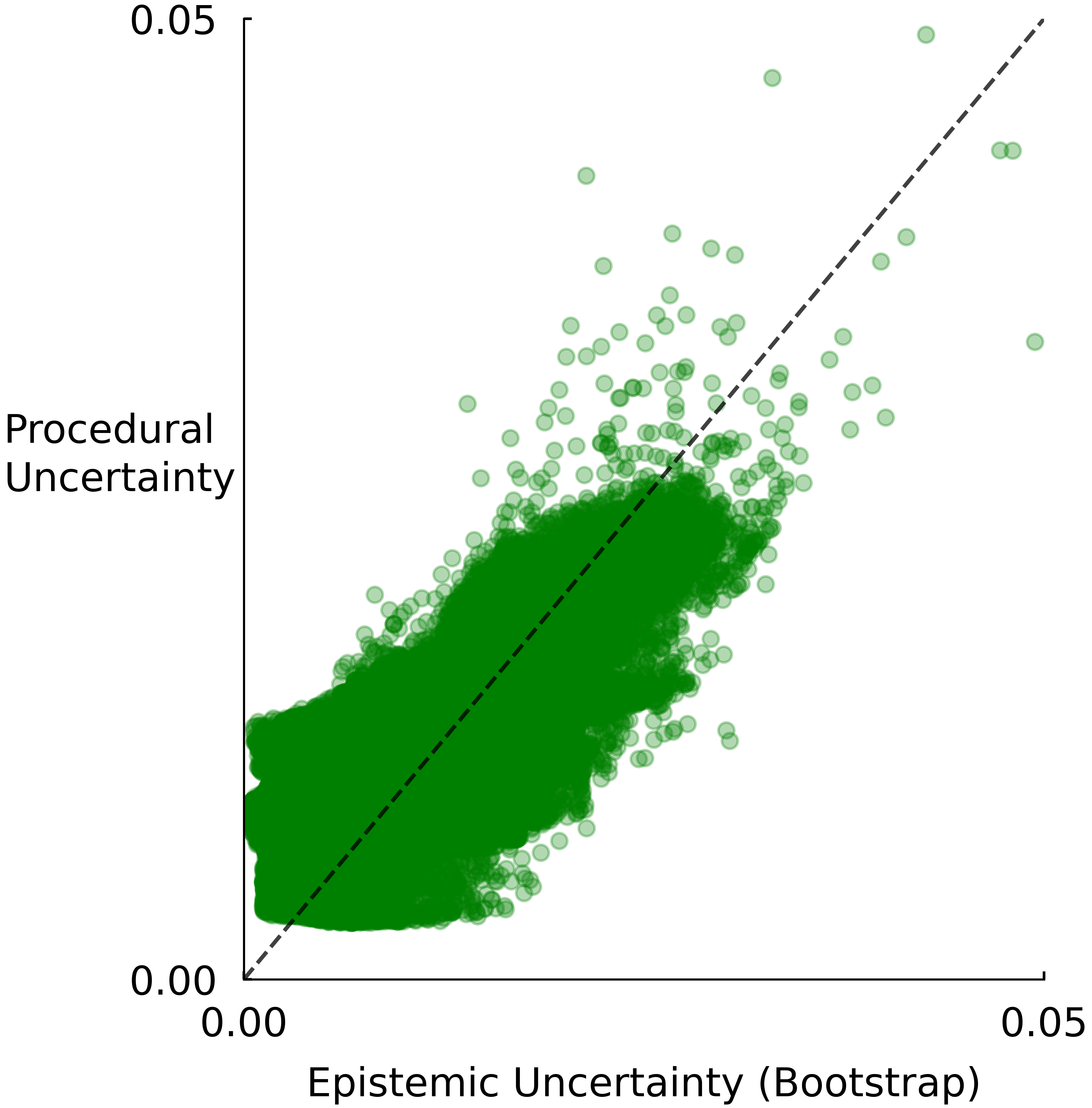}
        \caption{Bootstrap}
    \end{subfigure}
    \begin{subfigure}{0.23\textwidth}
        \includegraphics[width=\linewidth]{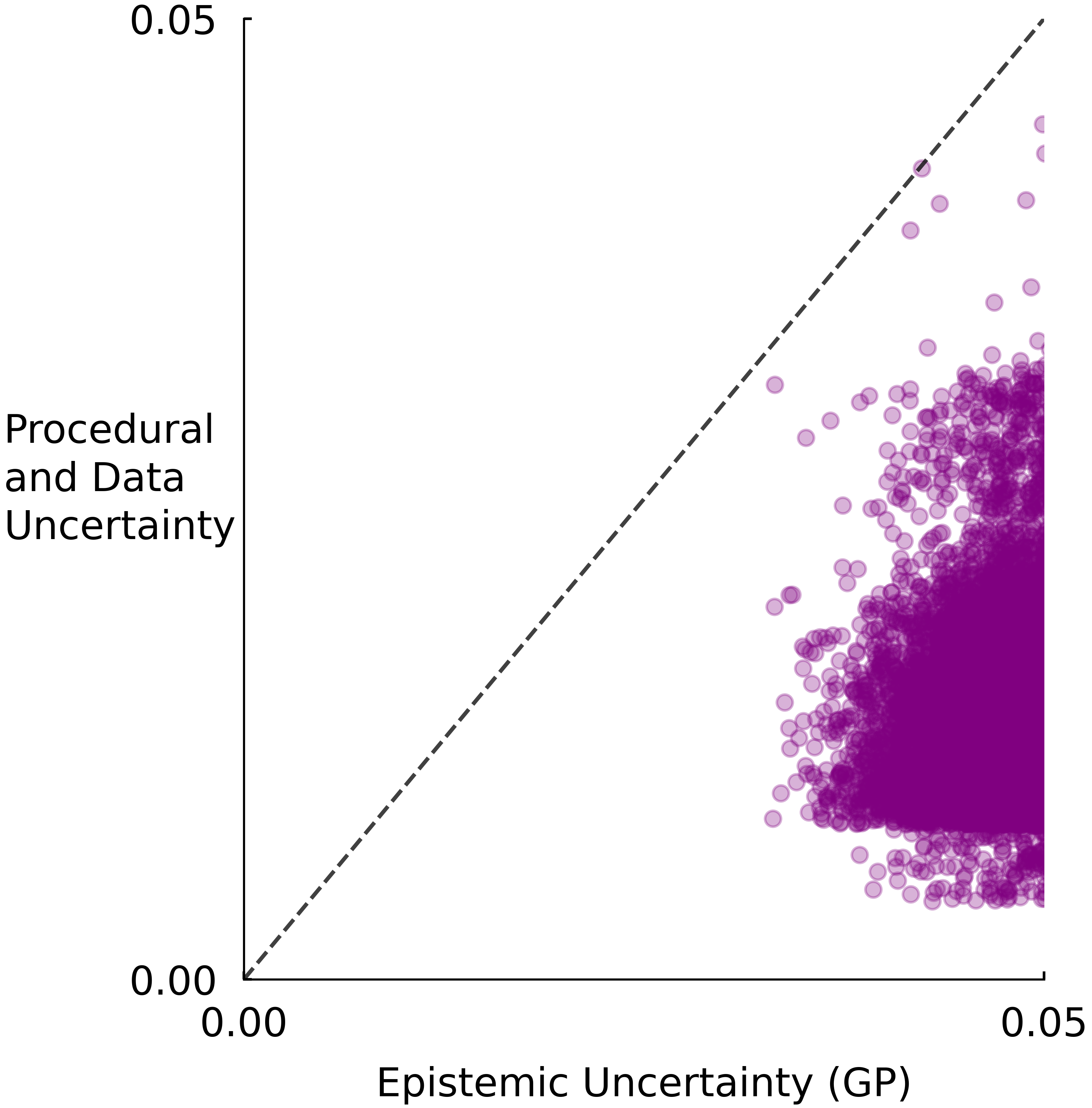}
        \caption{GP}
    \end{subfigure}
    \begin{subfigure}{0.23\textwidth}
        \includegraphics[width=\linewidth]{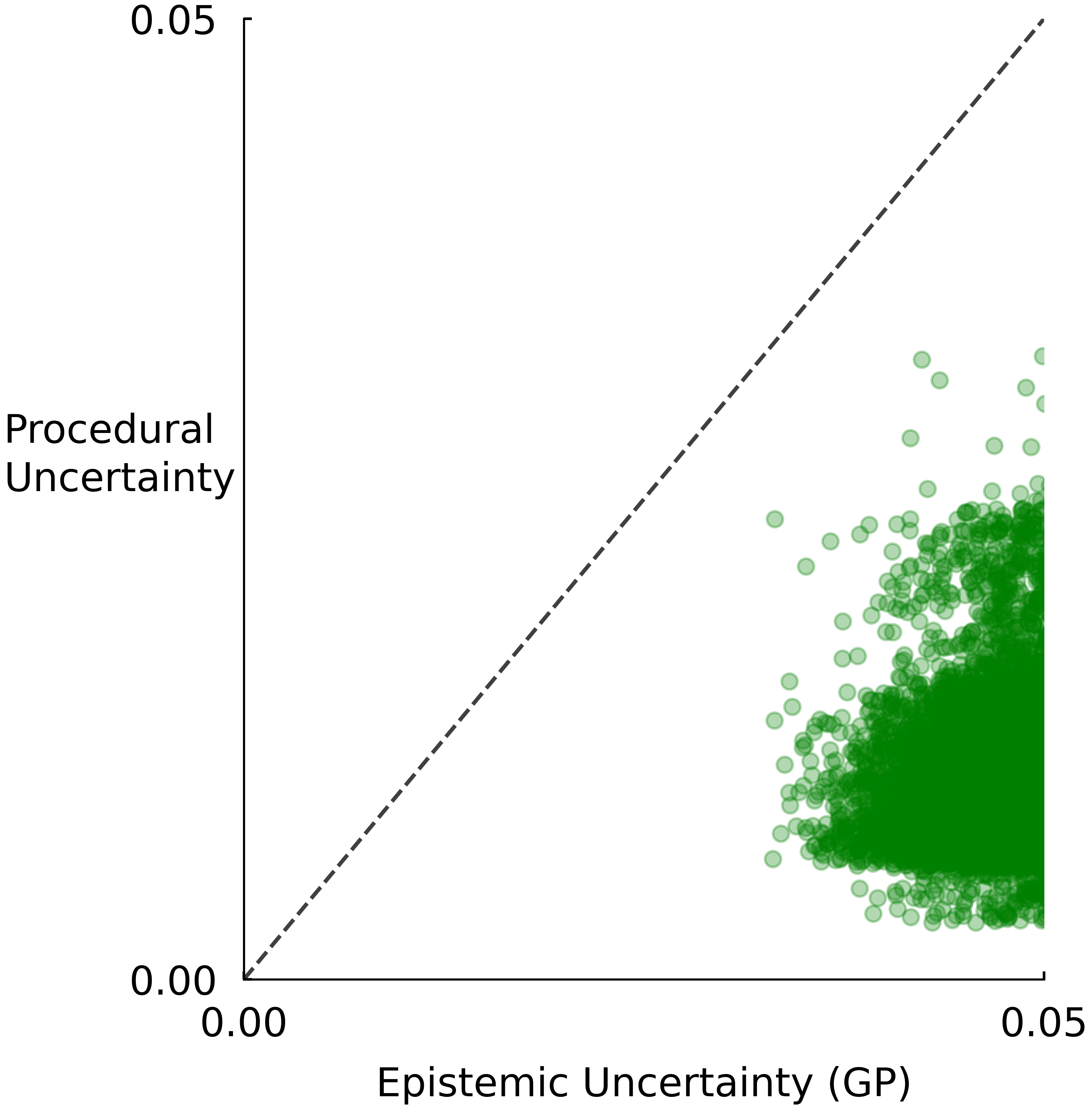}
        \caption{GP}
    \end{subfigure}

    \vspace{1em}

    \begin{subfigure}{0.23\textwidth}
        \includegraphics[width=\linewidth]{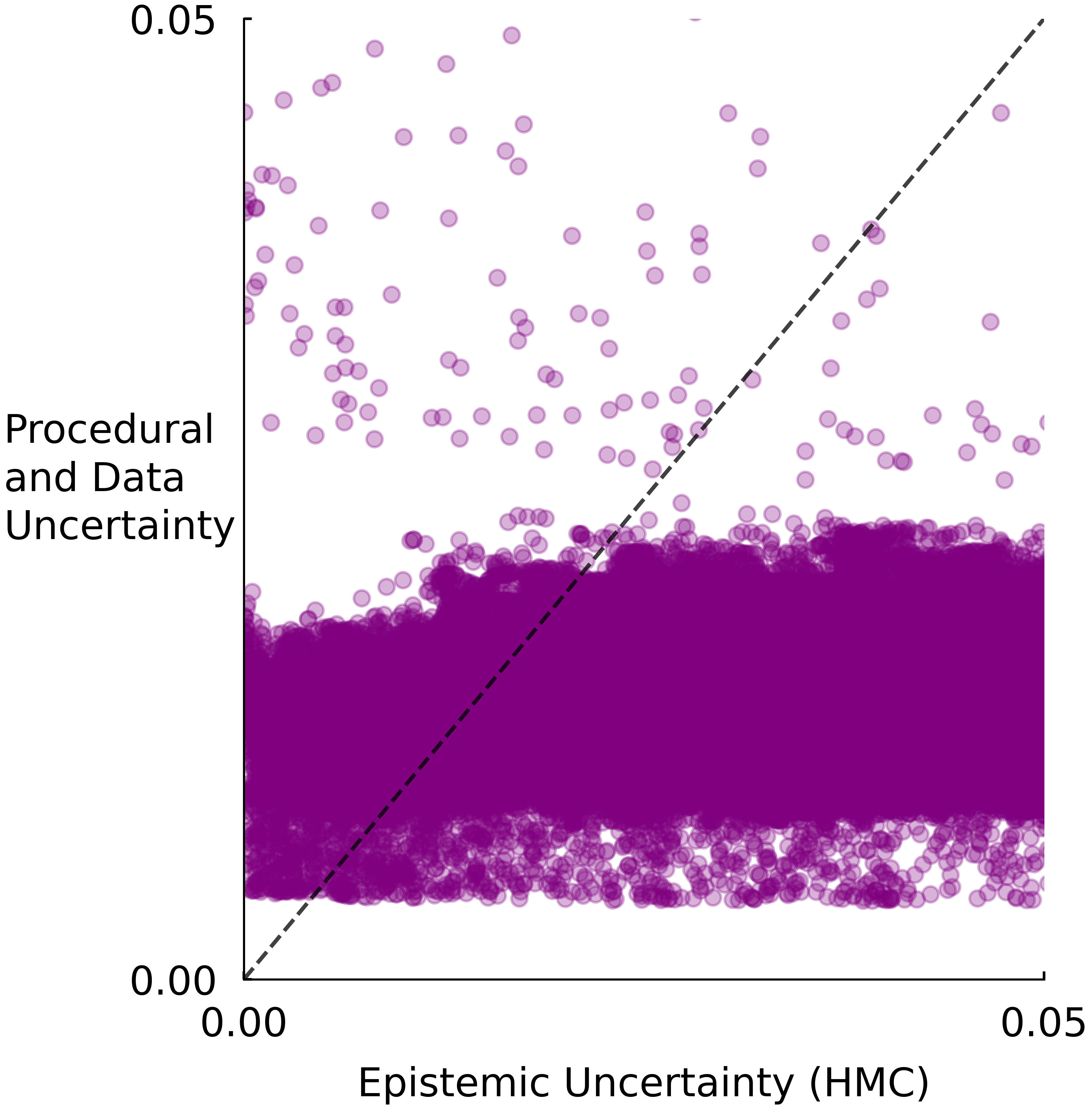}
        \caption{HMC}
    \end{subfigure}
    \begin{subfigure}{0.23\textwidth}
        \includegraphics[width=\linewidth]{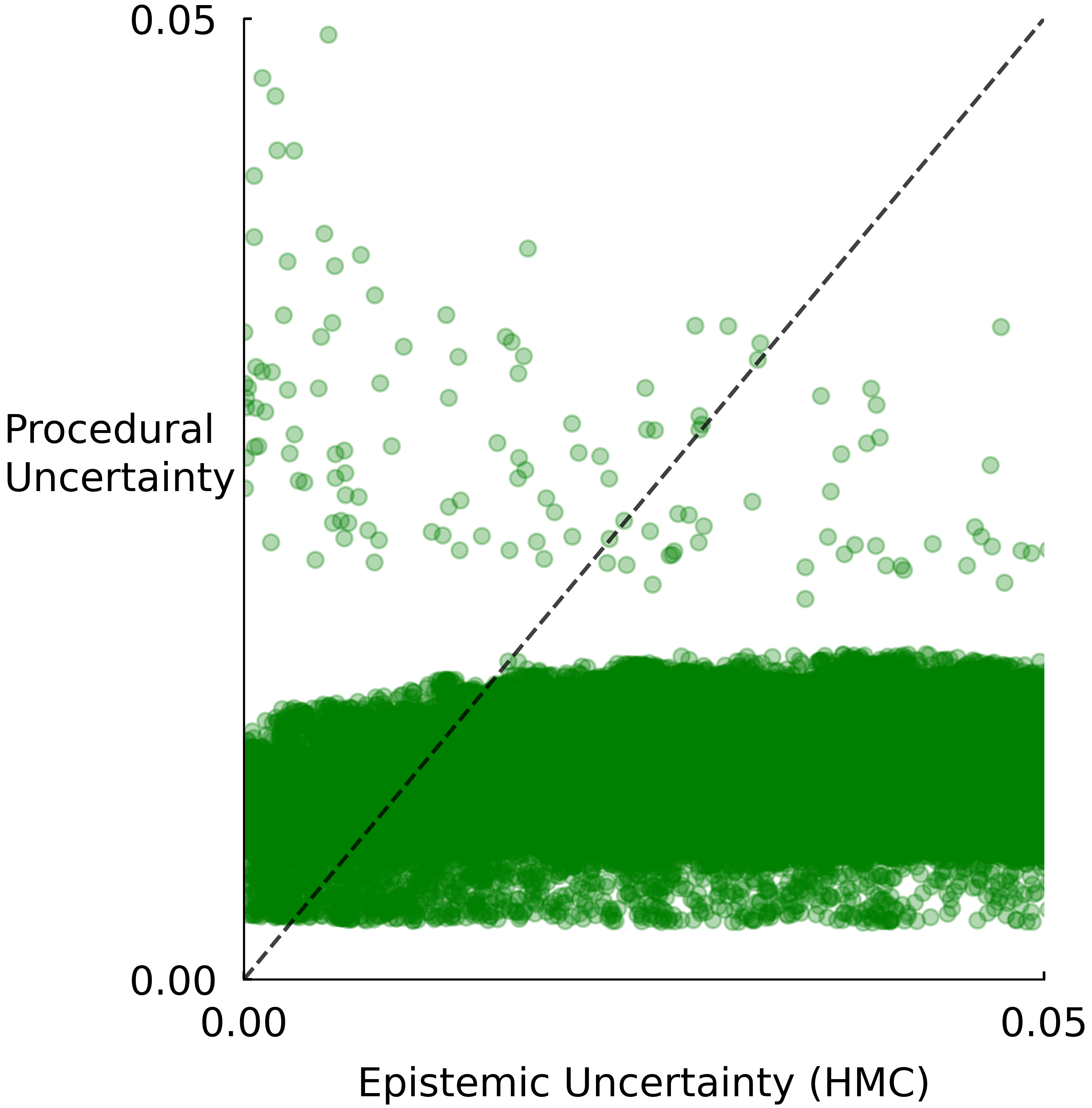}
        \caption{HMC}
    \end{subfigure}
    \begin{subfigure}{0.23\textwidth}
        \includegraphics[width=\linewidth]{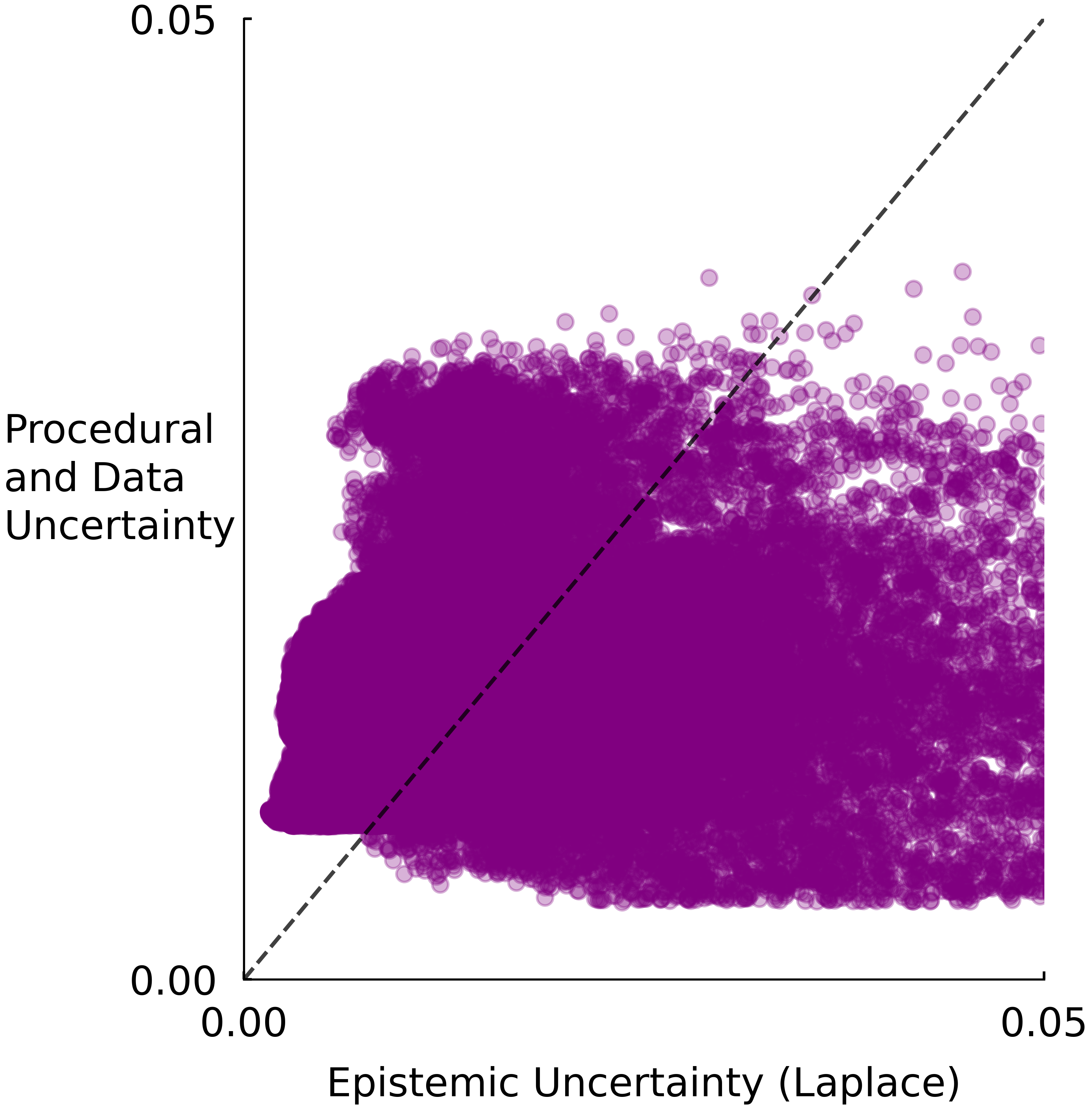}
        \caption{Laplace}
    \end{subfigure}
    \begin{subfigure}{0.23\textwidth}
        \includegraphics[width=\linewidth]{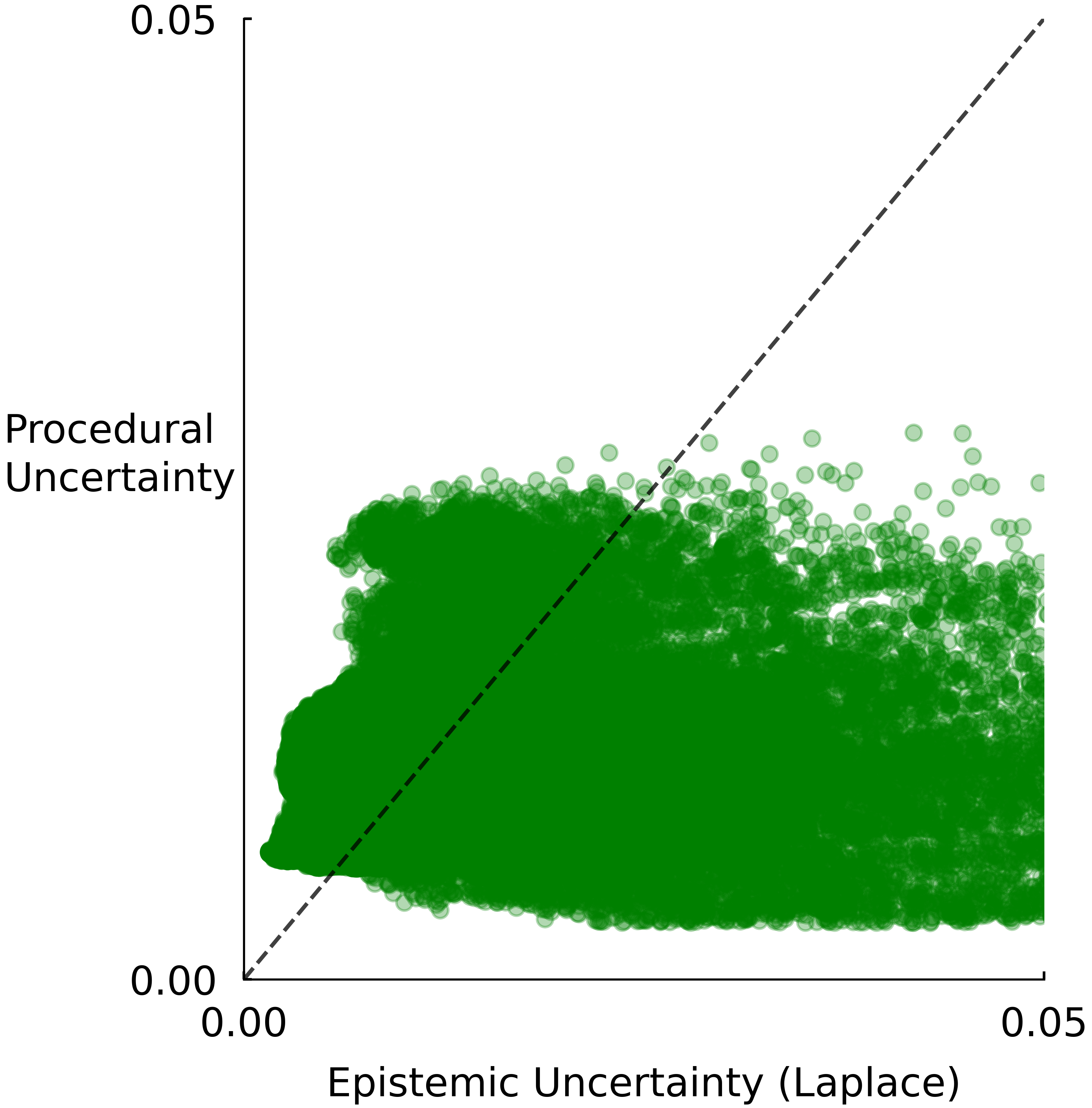}
        \caption{Laplace}
    \end{subfigure}

    \vspace{1em}

    \begin{subfigure}{0.23\textwidth}
        \includegraphics[width=\linewidth]{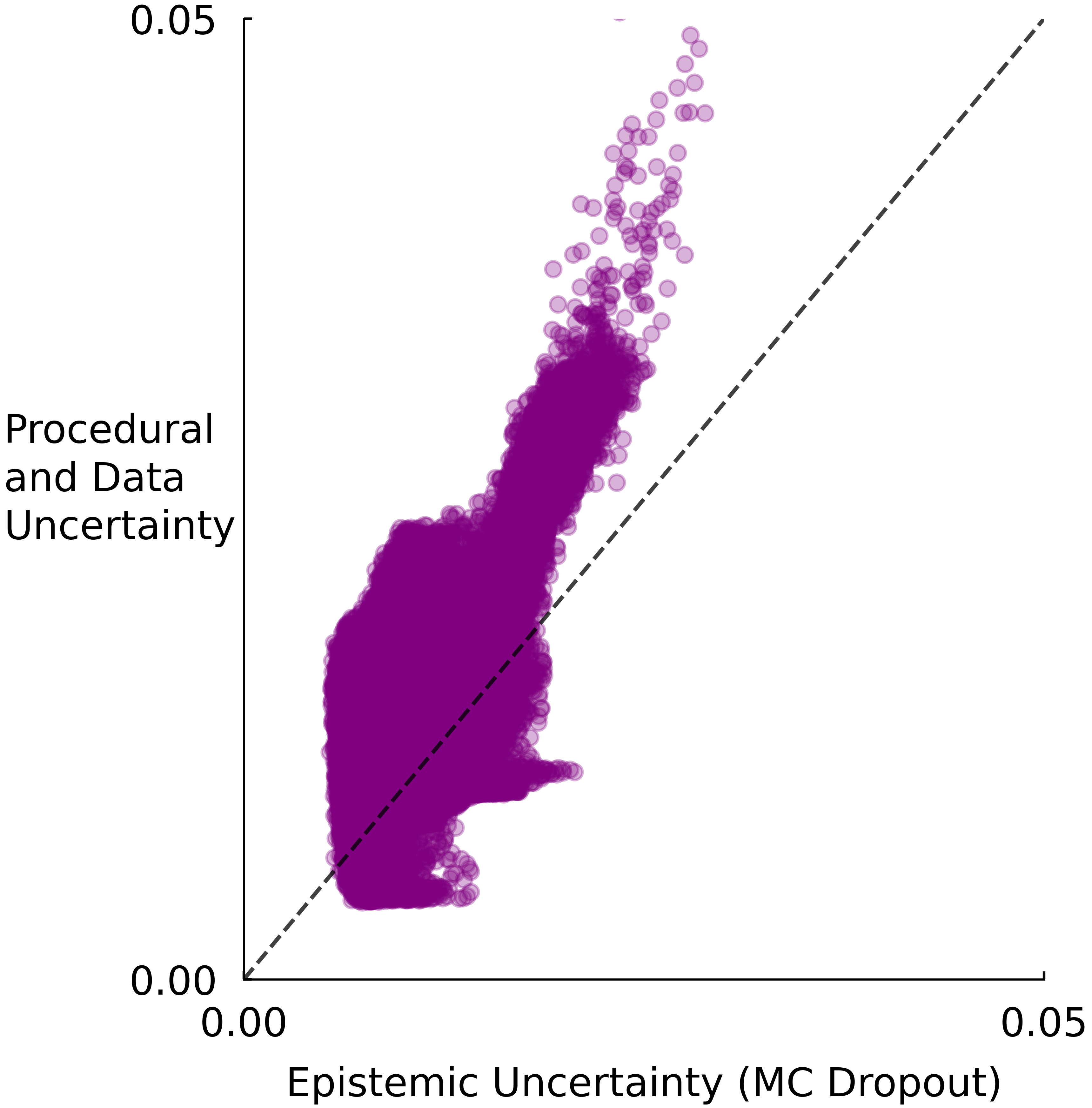}
        \caption{MC Dropout}
    \end{subfigure}
    \begin{subfigure}{0.23\textwidth}
        \includegraphics[width=\linewidth]{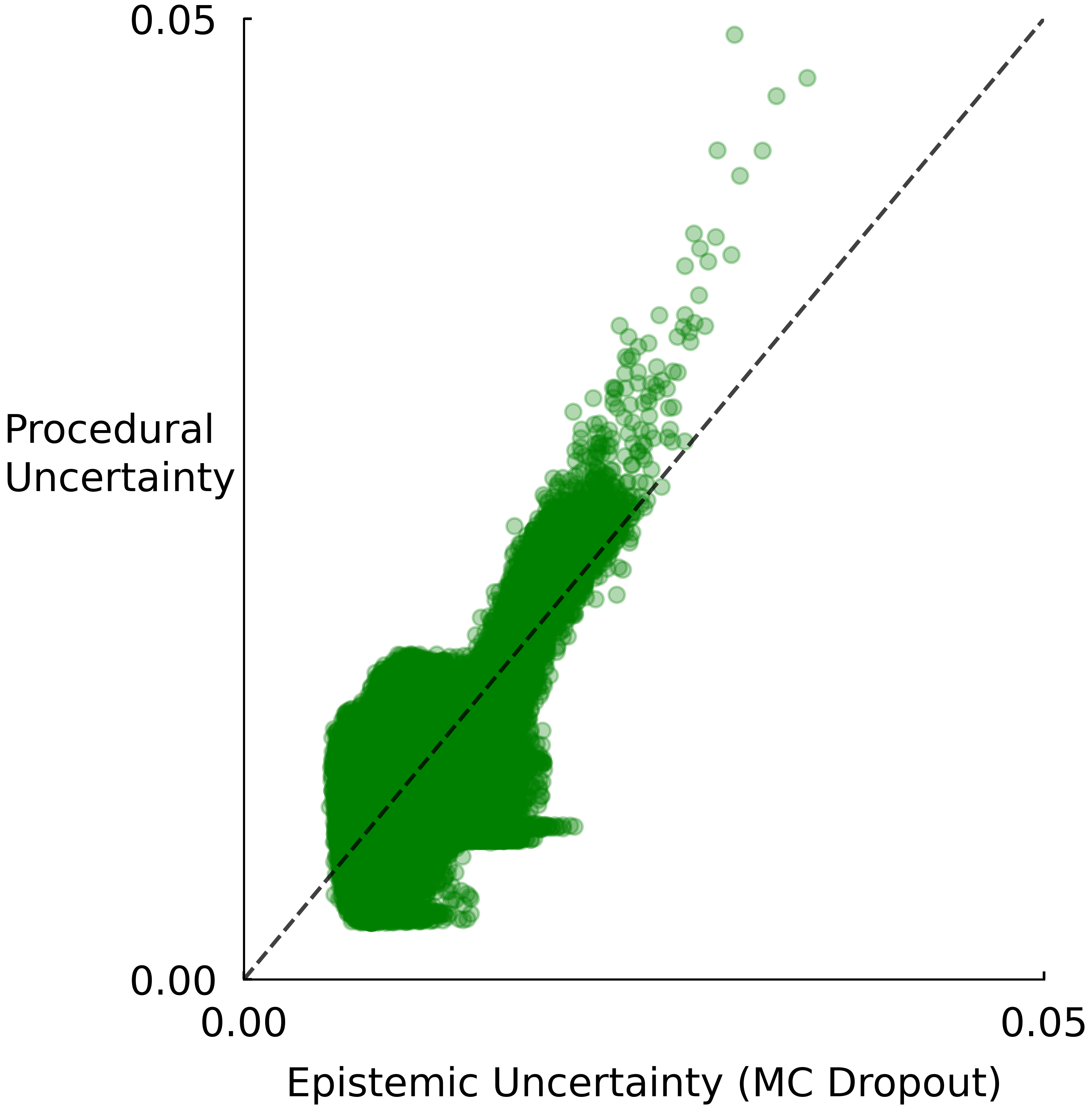}
        \caption{MC Dropout}
    \end{subfigure}
    \begin{subfigure}{0.23\textwidth}
        \includegraphics[width=\linewidth]{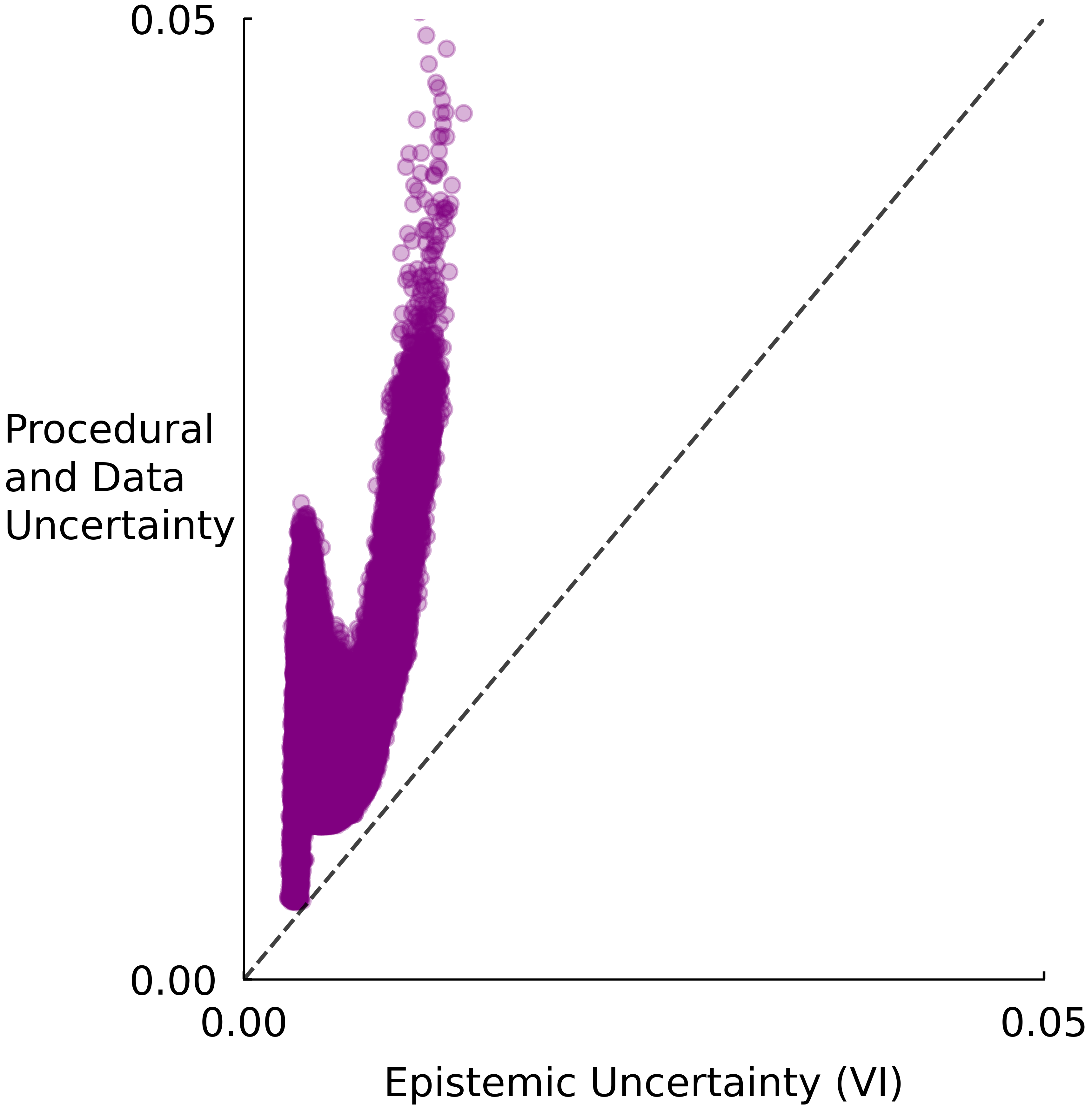}
        \caption{VI}
    \end{subfigure}
    \begin{subfigure}{0.23\textwidth}
        \includegraphics[width=\linewidth]{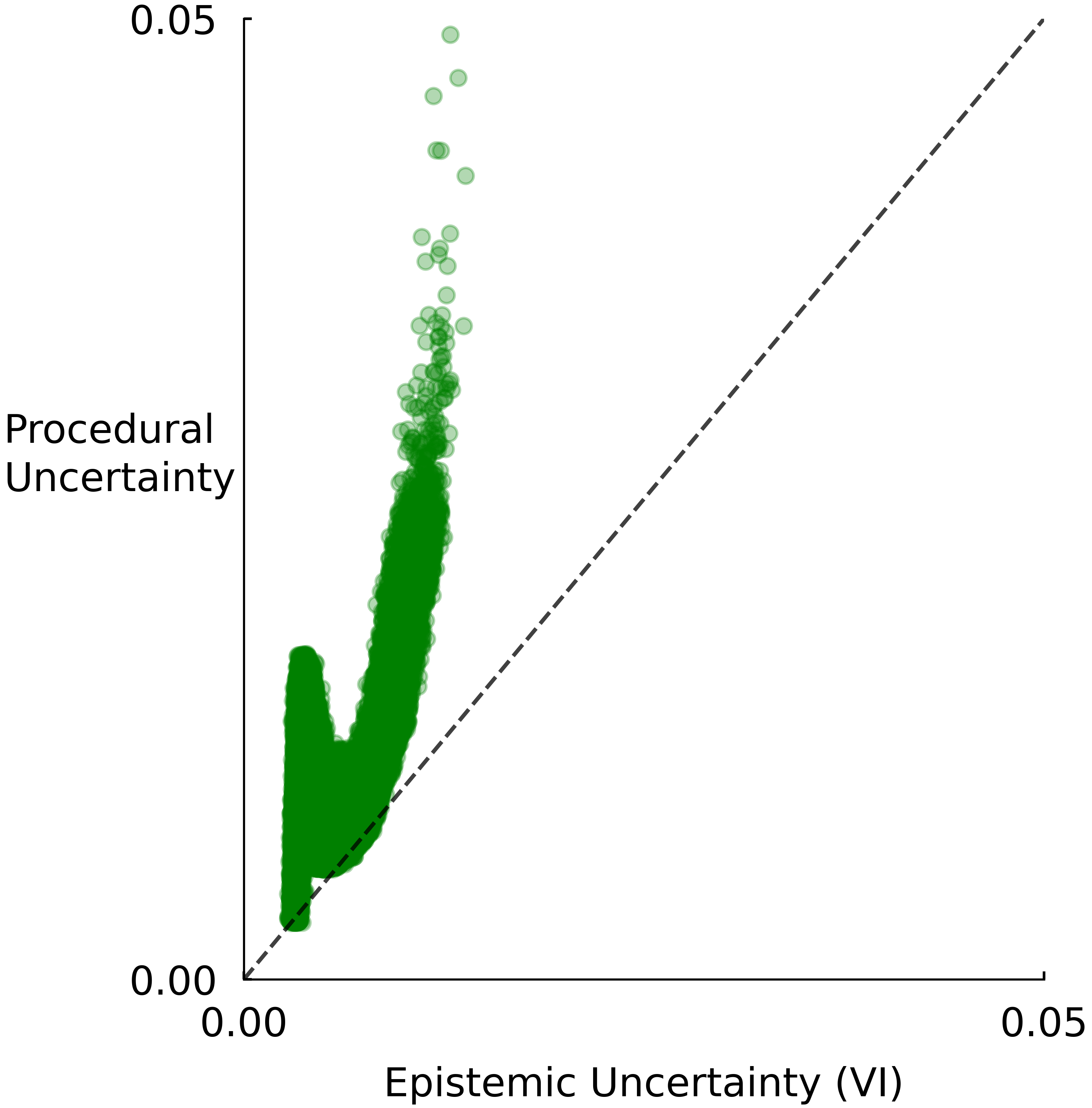}
        \caption{VI}
    \end{subfigure}

    \caption{Comparison of procedural and data uncertainties vs epistemic uncertainty across methods on the Taxi dataset. Figures in purple (first figure of every method) show the relation between the epistemic uncertainty from the reference distribution (including procedural and data uncertainties), and green pictures (second figure of every method) show the relationship between the procedural uncertainty from the reference distribution and the epistemic uncertainty from each method. In general, we see that adding the data uncertainty component shifts the relation between the two estimates.}
    \label{fig:taxi_uncertainty_grid}
\end{figure}

\end{document}